# Transductive Confidence Machine and its application to Medical Data Sets

David Lindsay
Department of Computer Science
Royal Holloway University of London
Full unit project

March 22, 2002

## Acknowledgements

I would like to acknowledge the guidance and support of Alex Gammerman throughout the course of this project. I would also like to thank Kostas Proedrou for explaining the TCMNN algorithm, and Leo Gordon and David Surkov for the use of their pCoMa system during the experimental process. Ian Jacobs and Philippe Van Trappen of St. Bartholomew's Hospital are acknowledged for their provision of the ovarian cancer data set. Finally, I would like to thank Siân Cox for her help in proofreading this report and for her general comments throughout.

**Abstract**

The Transductive Confidence Machine Nearest Neighbours (TCMNN) algorithm and a supporting, simple user interface was developed. Different settings of the TCMNN algorithms' parameters were tested on medical data sets, in addition to the use of different Minkowski metrics and polynomial kernels. The effect of increasing the number of nearest neighbours and marking results with significance was also investigated. SVM implementation of the Transductive Confidence Machine was compared with Nearest Neighbours implementation. The application of neural networks was investigated as a useful comparison to the transductive algorithms.

# Contents

# List of Figures

# List of Tables

# Chapter 1

# Introduction

## 1.1 Introduction to machine learning

Websters dictionary defines 'learning' as:

> "The act, process, or experience of gaining knowledge or skill through schooling or study".

Machine learning addresses the question of how to build computer programs that improve their performance at some task through experience [4]. Since the dawn of the computing age, people have wondered whether computers might be able to learn. In recent years many successful machine learning applications have been developed, tackling a wide range of problems such as creating DNA profiles, reading postcodes, diagnosing symptoms of illnesses, and analysing financial markets to mention but a few.

The field of machine learning draws on concepts and results from many disciplines including statistics, artificial intelligence, philosophy, information theory, biology, cognitive science and computational complexity.

The ease in which human beings perform everyday tasks such as recognising faces, understanding natural speech and reading handwritten digits, hides outstanding complexity, and are envious tasks to machine learning practitioners.

To comprehend this complexity consider the problem of trying to classify a 10 $\times$ 10 pixel image, with each pixel described by a colour ranging from 1 to 256. A naive solution would be to store a look-up table of all possible images, along with their classifications. Although this technique would ensure error-free classification (assuming that the problem could be formulated this way in the first place!), the time taken to search through all $256^{10\times10} \approx 10^{102}$ patterns of images would be infeasible. To carry out this kind of computation would be prohibitive in terms of both space and time complexity. Estimating that a standard desktop computer can check roughly 100,000 patterns a second, a complete search would take $10^{95}$ seconds (approx $10^{79}$ billion years!) to complete.

There are several major directions in machine learning. Among them are:

**Supervised Learning** This assumes the existence of a set of training examples, constructed by the *supervisor* of the task. Each training example contains a set of attributes, that are required to describe the problem, and a target category/value that is to be learned. Using these training examples the learning algorithms build up some form of general decision rules that can be used to classify new sets of test examples, which have unknown categories.

**Unsupervised Learning** This area of learning differs mainly in the fact that the classifications or structure of the problem to be learned are unknown. In *unsupervised learning* or *clustering* there is no explicit teacher/supervisor, the system must therefore form clusters of 'natural groupings' of the input patterns. The way in which 'natural' is defined, depends on the particular learning algorithm used.

**Reinforcement Learning** In *reinforcement learning* or *learning with a critic*, no target category is explicitly expressed, instead the only teaching feedback is whether the tentative prediction made by the algorithm is right or wrong. This is analogous to a critic that merely states if something is right or wrong, but does not specify *how* it is wrong.

### 1.1.1 Supervised learning techniques

This project will concentrate on the particular area of supervised learning. This is by far one of the most well developed areas of machine learning, and many different algorithms can be considered. This section will outline the usage of some of the most commonly used algorithms and their strengths and weaknesses. According to Vapnik there are three main learning problems to be considered, that encompass many specific problems [5]. These general formulations are as follows:

#### Problem 1 : Pattern Recognition

This is where each of the examples is labelled with a finite discrete value. This value is said to be the class that the particular example belongs to. For example, in the ovarian cancer data set, benign cases are classified with label '0' and malignant cases with label '1'. These labels for each of the examples in the training set are provided by the supervisor. The problem is to correctly predict the labels of new test examples, which the supervisor does not know the labels for.

**Problem 2 : Regression**

This is where each of the examples is labelled with a continuous real target value provided by the supervisor. The problem is to minimise the squared sum of all the differences between the target values and the values predicted the learning algorithm.

**Problem 3 : Density Estimation**

This is concerned with inferring the specific details of the probability distribution that underlies any of the previous two problems. The problem is to be able to produce some probabilistic measure with each prediction that the learning algorithm makes.

This project will mainly be concerned with the problems of pattern recognition and density estimation, as the medical databases that have been provided for analysis in this project can easily be formulated in this way.

### 1.1.2 Statistical methods vs. machine learning techniques

Up until the early 1980's, classical statistical methods had been predominantly used to extract useful information from data, and to improve decision making.

However the volume of data being generated by businesses, medical and scientific establishments became more and more substantial, which began to reveal the limitations to the use of classical statistical techniques. As the cost of computing decreased and processing power increased, attention turned to developing from some of the original machine learning techniques that had originated in the early 1960's.

Unlike classical parametric and non-parametric statistics, machine learning techniques have proved capable of dealing with very high dimensional hypothesis spaces. There has been a dramatic push from the various communities over recent years to develop ever more efficient and powerful algorithms.

### 1.1.3 The *i.i.d* assumption

Many machine learning methods are designed to work and their performance is analysed under the general *i.i.d* assumption. In contrast, many classical parametric statistics are based on more complex probability distributions such as normal distribution that requires knowledge of parameters such as variance, and the calculation of a mean. Non-parametric statistics offers solutions that are not based on assumptions about data, but are limited by their complexity to analysing relatively smaller dimensional problems.

'i.i.d' stands for 'independently and identically distributed'. The 'identical' part implies that the set of training examples are drawn randomly from the some fixed probability distribution. The 'independent' part assumes that each example is drawn independently of the other from this distribution. Using the example of the ovarian cancer data set, the assumption is made that the diagnosis or measurements of one patient from the set of examples does not affect that of another patient. Note: This can be a very strong assumption in certain applications!

### 1.1.4 Inductive vs. transductive approach to machine learning

The traditional *inductive* approach to machine learning involves making two distinct steps:

1. Processing the set of training examples to find a decision function.

2. Using the decision function to make predictions on new unseen test examples.

In contrast the *transductive* or *instance based* approach delays this processing to find a decision function until a new test example is presented. It therefore merges the two separate steps of the inductive approach into one single step, bypassing the need to find a general decision rule. For this reason, transductive methods are often referred to as 'lazy learning' techniques [4].

The majority of existing learning methods such as neural networks, Support Vector Machines (SVM) and genetic algorithms, fall under the category of inductive learning. There are relatively fewer transductive algorithms at present however the most commonly used are the $k$-Nearest Neighbour algorithm and radial basis function networks [4].

Inductive learning is often preferred as it is more efficient to process than transductive techniques. This is because the majority of calculations can be carried out off-line, and then a relatively smaller amount of calculations can be made when using the decision function to classify new examples.

Inductive techniques lend themselves perfectly to systems which require almost real time responses to queries submitted to it. Although the initial preprocessing to find the decision function may be intensive, the process of classifying new examples is relatively faster using inductive techniques.

The problem with the inductive process is that by the time the new test example $x_q$ is observed, the method has already chosen its global approximation to the decision function. This means that if a new example is presented that is from a different region that has not been expressed by the training examples, the function may fail to produce a valid prediction. This often causes the inductive approaches generalisation to suffer in comparison to solutions of the transductive approach.

The key advantage of the transductive approach is that instead of estimating the decision function for the entire input space, the decision function can be estimated locally and differently for each new example to be classified. This is a big advantage when the decision function to be modelled is very complex.

The main disadvantage with transductive algorithms is that the cost of classifying new test examples is high, because relatively little off-line preprocessing is possible.

Almost all of the calculations must be made at the point of querying a new example. Another disadvantage of many transductive approaches is that they consider all the attributes of the examples. If the decision function really only depends on a few of the many available attributes, then the examples that are really 'similar' may be calculated to be a large distance apart. Both the inductive and transductive approach can be applied to both part recognition and regression cases as discussed earlier.

### 1.1.5 Criticism of traditional machine learning algorithms

Traditional machine learning methods such as neural networks and support vector machines have been concerned in producing 'bare predictions', without any measure of confidence. These learning algorithms have been attributed to many successes in applications of pattern classification problems in recent years, yet none of them address the problem of density estimation.

Bayesian algorithms make explicit manipulations of raw underlying probabilities of problems, to create their predictions. The Bayesian approach to machine learning offers techniques of developing measures of confidence with predictions, however these are built on strong assumptions about the underlying probability distribution of problems. Trying to make assumptions about a problem may be difficult and open to interpretation.

The probably approximate correct (PAC) framework to machine learning could loosely be considered as solution to the problem, in that it attempts to provide bounds on the errors made with classification. This approach does not prove practical and produces ill posed confidence measures with probabilities greater than 1, as discussed in [19].

Ideally one would like to harness the power of existing learning algorithms, and develop techniques of extracting confidence measures with our predictions, using only the existing assumption that the training examples are i.i.d mentioned earlier.

## 1.2 Development of a learning system, and applications to medical data sets

The main focus of this project will be to apply transductive confidence techniques (discussed later) to important medical data sets. This will include the development of a reliable and accurate learning system. In particular the ovarian cancer diagnosis problem posed by the St. Bartholomews hospital data sets will be focussed on in greater detail.

Successful applications of machine learning to medical diagnosis can offer fast and inexpensive solutions to what can be very difficult complex problems. The need for such systems is highlighted by the problem of diagnosis and screening for ovarian cancer.

Approximately 60% of ovarian cancer cases are fatal, with women dying within five years [27]. This is one of the highest cancer death ratios for any malignancy. Ovarian cancer is often known as the "disease that whispers" - symptoms aren't normally noticed until the cancer has spread outside the ovary.

Early detection of ovarian cancer increases survival rate to 80% from 15%. If the cancer has not spread beyond the ovary, patients have a 85-95% chance of living five years or more with proper treatment.

Effective screening programs have yet to be developed for the disease. Tests currently available have limited effectiveness, and for widespread testing, are prohibitively expensive.

When using a learning system to tackle problems such as ovarian cancer screening is necessary to gather a corpus of data that encapsulates the problem. The role of machine learning systems to these problems would be to try to discover methods of inferring relationships to generate hypotheses about the data provided. Using these hypotheses, the learning systems would then attempt to act as a 'consultant' for new examples where the diagnoses are not known.

## 1.3 Formulation of objectives

The aim of this project will be to address the following issues:-

1. Study the background theory behind the construction of the transductive confidence machine.

2. Design and implement the TCM Nearest Neighbours algorithm and a simple front end user interface to the algorithm.

3. Apply the transductive confidence machine to various medical data sets.

4. Time permitting, further extensions of the theory will be investigated, such as:-

    (a) Investigate the use of alternative distance metrics with the TCMNN algorithm.

    (b) Investigate the use of polynomial kernels feature mappings with the dual version of the nearest neighbours algorithm.

    (c) Compare results with inductive algorithms such as neural networks.

    (d) Investigate increasing the number of nearest neighbours with the TCMNN algorithm.

# Chapter 2

# Background Theory

This chapter will give some detail about the underlying theory behind my project. Firstly it will briefly discuss Kolmogorov complexity and Algorithmic Randomness. The recent development of this theory as an application to machine learning will be illustrated, detailing the construction of the transductive confidence machine. Attention will be focused on the $k$-Nearest Neighbour classifier and how it is adapted for transductive confidence techniques. Throughout this chapter references to various medical data sets will be made in my examples.

## 2.1 Applications of algorithmic randomness theory

Randomness had traditionally been a 'grey' area of probability and statistics. However, mathematical computer science had a keen interest to the problem. The notion of randomness was proposed by Kolmogorov to provide a suitable basis for the *applications* of probability. Kolmogorov developed his axioms in 1933, which then became universally accepted as the basis for the *theory* of probability. Much of the definitions of the algorithmic theory of randomness concentrate on the randomness of finite and infinite binary sequences [6]. This section will show how computable approximations to algorithmic randomness have been developed by V. Vovk and A. Gammerman of Royal Holloway University of London as a new application to machine learning for extracting probabilities.

### 2.1.1 Definition of Kolmogorov complexity

Two main contributors to the theory of Kolmogorov complexity were Ray Solomonoff and Andrei Nikolaevich Kolmogorov. Alan Turing's original proposition of the Turing machine as a precise notion of a computing machine was further developed in 1936, to that of a universal Turing machine. It was found that this automaton could be used to simulate any other Turing machine. Complexity of an observed sequence of data, according to Kolmogorov, can be measured by length of the shortest program for a universal Turing machine that correctly reproduces the observed data.

### 2.1.2 Martin Lőf randomness test

One particular result of the study of algorithmic randomness that we are particularly interested in is Martin Lőf's definition of a randomness test developed in 1966. Martin Lőf extended the definition of sets to probability distributions, introducing so-called randomness tests.

**Definition 1** ***Martin-Lőf randomness test:*** *Let* $\mathbf{Z}$ *be the set of all possible classified examples of size n. A function* $t : Z^* \rightarrow [0, 1]$ *is a test for randomness if*

1. *For all* $n \in \mathbf{N}$*, for all* $r \in [0, 1]$ *and for all probability distributions* $P$ *in* $Z$*,*

$$P^n\{z \in Z^n : t(z) \leq r\} \leq r \tag{2.1}$$

2. *t is 'upper semi-computable'.*

Notice that this definition is intuitively the notion of a statistical $p$-value, in that given a chosen 'significance level' of $r\%$, we expect $r\%$ of examples to have randomness level of $r\%$. We can interpret this as expecting random sequences $z$ with low values $t(z)$ to be rare.

### 2.1.3 Creating a computable approximation of a randomness test

Based on the Martin-Lőf definition, the work of Vovk and Gammerman 1999 [20], developed applications of algorithmic randomness to the field of machine learning. The key thinking was that by developing a technique of analysing the randomness of a set of training examples, one could construct the confidence measure that was missing from many existing learning algorithms.

As much of theory had been mainly concerned with the randomness of binary strings, Vovk and Gammerman firstly expanded the concept by thinking of the set of training and test examples used by learning algorithms as a sequence of objects (encapsulating their attributes which described them), with labels (namely their respective classification).

Another problem with the original theory was that the second part of the Martin-Lőf definition of a randomness test explains that the true value of this measure of randomness is not feasibly computable. By being 'upper semi computable', the value can only be approximated from above, implying that the calculation may need an infinite amount of time to converge to the true value.

### 2.1.4 Formal setting of the problem

To formulate the problem in terms of a machine learning context, consider a set of $l$ training examples $\{(x_1, y_1), \ldots, (x_l, y_l)\}$. We consider our new examples in turn, as the $l+1$ example in the sequence.

We now require to be able to develop a function $t(z)$ that behaves as defined in Martin-Lŏf definition of a randomness test, so that we can measure the randomness of the sequence of training examples (as opposed to a binary sequence).

### 2.1.5 Creating a measure of strangeness

The first crucial step in creating a computable approximation of randomness was to create a level of 'strangeness' for each example in the set of training examples, and the new test examples.

The intuitive notion behind this measure is of how surprising an example with those particular attributes and that classification are in relation to the other examples that are presented.

How this measure of 'strangeness' for each individual example is calculated depends on which machine learning algorithm is being used underneath the transductive confidence machine (TCM) framework. These measures of strangeness are often referred to as $\alpha$ values.

### 2.1.6 Creating a randomness test

Once these measures of strangeness $\alpha_i$ have been created using the respective learning algorithm, they are inserted into the following function to calculate the $p(\alpha_{l+1})$.

$$p(\alpha_{l+1}) = \frac{\#\{i : \alpha_i, 1 \leq i \leq l+1, \geq \alpha_{l+1}\}}{l+1} \tag{2.2}$$

The value returned from the function is called a $p$-value. Therefore the $p$-value for the sequence $\{\alpha_i, \ldots, \alpha_l, \alpha_{l+1}\}$ is $p(\alpha_{l+1})$, where $\{\alpha_i, \ldots, \alpha_l\}$ are the strangeness measures for the training examples, and $\alpha_{l+1}$ is the strangeness measure for the new test example with a possible classification assigned to it.

The intuition behind this function is that it returns the proportion of examples that are at least as strange as the new example $z_{l+1}$ with a possible classification assigned to it. So if a new example is very strange compared to the training set, the function will return a low value (and vice versa). A $p$-value close to 0 is viewed as a totally random (or untypical) completion, and a $p$-value close to 1 as not random (a typical completion).

### 2.1.7 Validity of the randomness test

The function $p(\alpha_{l+1})$ in equation 2.2 has been designed to be 'valid' with respect to Martin-Löf's definition of a randomness test (see Equation 1) under a slightly weaker condition of exchangeability $\mathcal{P}_{exch}$.

We define our condition of exchangeability $\mathcal{P}_{exch}$ as the assumption that given any sequence of objects, all permutations of the sequence are equally as likely, and will consequently mean that each example's strangeness measure defined earlier in equation 2.7 will be the same, independent of its placement in all possible permutations of the sequence [1].

**Theorem 1** *(Vovk and Gammerman, 1999) The function* $p(\alpha_{l+1})$ *satisfies the definition of Martin-Löf randomness test w.r.t to* $\mathcal{P}_{exch}$.

**Proof** Given a sequence $\mathbf{z} = \{z_1, \ldots, z_l, z_{l+1}\}$ where $\{z_1, \ldots, z_l\}$ is the set of training examples and a new test example $z_{l+1}$.

Let $\alpha_{l+1}$ be the strangeness of the new test example $z_{l+1}$ as some possible classification $c \in C$, and

Let $\{\alpha_1, \ldots, \alpha_l\}$ denote the strangeness measures of the training examples.

Let $S_j(\mathbf{z})$ define the $j$ smallest corresponding strangeness values $\alpha_i$ of the $l+1$ examples in sequence $\mathbf{z}$.

We aim to show that for all sequences $\mathbf{z}$ of length $l+1$,

$Prob\{\mathbf{z} \in Z^{l+1} : p(\alpha_{l+1}) \leq \frac{j}{l+1}\} \leq \frac{j}{l+1}$

By the definition of the function $p(\alpha_{l+1})$,

$p(\alpha_{l+1}) = \frac{j}{l+1} \Rightarrow$ the strangeness value $\alpha_{l+1}$ of the new test example $z_{l+1}$ belongs to the ordered set of the $j$th strangest example $S_j(\mathbf{z})$.

The $\mathcal{P}_{exch}$ assumption $\Rightarrow$ every example is equally as probable in any position in sequence, therefore each example has probability $\frac{j}{l+1}$ of occurring.

There are $j$ different ways of $\alpha_{l+1}$ belonging to $S_j(\mathbf{z})$.

Therefore the probability of $p(\alpha_{l+1}) = \frac{j}{l+1}$ is $\frac{j}{l+1}$.

[1]This causes problems with SVM implementation of TCM as the order in which the training examples are presented with SVM will affect the alpha values returned by the algorithm

This completes the proof, as indeed the function $p(\alpha_{l+1})$ satisfies $Prob\{\mathbf{z} \in Z^{l+1} : p(\alpha_{l+1}) \leq r\} \leq r$ for some constant $r$.

### 2.1.8 How to use the $p$-values

The $p$-values that are created give us an intuitive idea of confidence. They reflect how random (typical) a particular completion is in relation to a set of training examples. An example of TCM in action can be seen in Figure 2.1.9.

As seen in the TCM pseudo code later, the following steps are made in constructing confidence measures.

1. **Predict the class with the largest $p$-value** - so as to predict the most typical classification.

2. **Output confidence as $1 - 2^{nd}$ largest $p$-value** - this explains how confidently other possible classifications can be ignored; lower $p$-values for other classifications indicate that they are less likely. If the first and second largest completions have equal $p$-values, then it is not possible to easily distinguish which is the most typical completion, therefore confidence in the prediction would be low.

3. **Output credibility as largest $p$-value** - this is a measure of how much credibility can be placed in a prediction and its respective confidence reading. If the largest p-value is very low then the completion is not very typical, therefore indicating that although other predictions with a high confidence reading may be discounted, the actual predication is still not to be trusted.

### 2.1.9 The Transductive Confidence Machine (TCM) algorithm

Assembling all the previous parts together we can now specify the general transductive confidence machine algorithm as follows.

**TCM Algorithm**

Given a set of $l$ training examples, and $r$ new test examples, we consider each new test example in turn as the $l + 1$ example in the new sequence to be analysed.

for each new test example $x_{l+1}$
  for all possible classifications $c$
    ♢ calculate the strangeness values $\alpha_i$ of each of the $l$ training examples in the set $\{(x_1, y_1), \ldots, (x_l, y_l), (x_{l+1}, y_{l+1})\}$
    ♢ calculate the strangeness value $\alpha_{l+1}$ of the new example $(x_{l+1}, y_{l+1})$ classified as $c$
    calculate p-value of new example $x_{l+1}$ classified as $c$

end for
predict the class with the largest $p$-value
output as confidence 1-2nd largest $p$-value
output as confidence the largest $p$-value
end for

Note that the only part which needs to be adapted for the TCM to work with any learning algorithm is the lines of the code that deal with the calculation of the strangeness alpha values (marked with $\diamondsuit$).

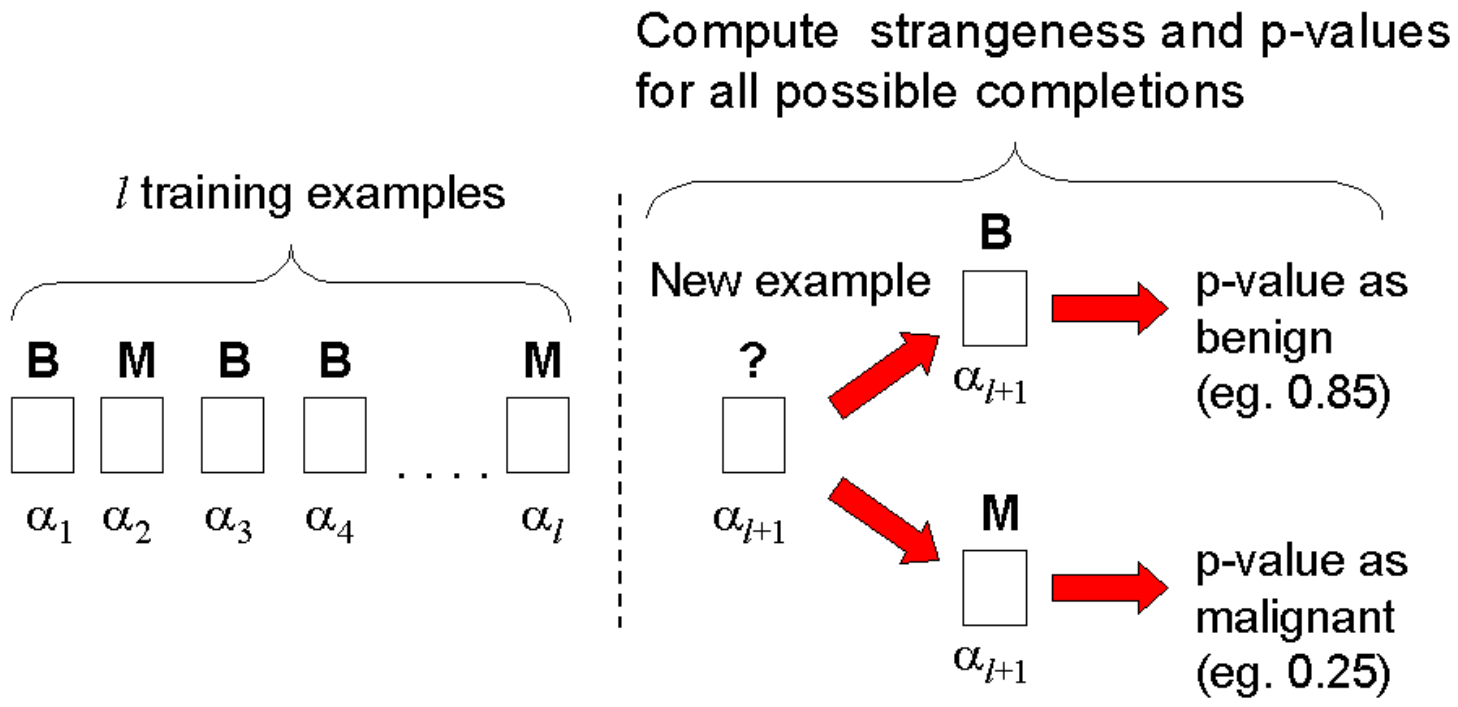

Figure 2.1: **Example of the TCM in action:** Continuing with the example of St. Barts data set (B for Benign, M for Malignant), the diagram above outlines how the TCM will be used to make predictions. In this instance the new example would be predicted as Benign as 0.85 > 0.25. Confidence as 75% (1 - 0.25 = 0.75), and credibility would be 85%.

## 2.2 The $k$-Nearest Neighbours classifier

This section will discuss the workings of one particular transductive machine learning algorithm, namely the $k$-Nearest Neighbours classifier.

### 2.2.1 The $k$-nearest neighbours algorithm (KNN)

The $k$-Nearest Neighbour classifier is the most basic of the available transductive learning algorithms. This algorithm assumes that all examples correspond to points in $n$-dimensional space $\mathbf{R}^n$. The Euclidean distance between training examples determine which example are so called 'nearest neighbours'. The distance between two examples $x_i$ and $x_j$ is defined to be $d(x_i, x_j)$, where

$$d(x_i, x_j) = \sqrt{\sum_{k=1}^{n} (x_i^k - x_j^k)^2} \quad (2.3)$$

## $k$-NN Algorithm (KNN)

Given a new training example $x_{l+1}$, and let $x_1 \dots x_k$ be the $k$ training examples nearest to $x_{l+1}$.

Return
$\hat{f}(x_{l+1}) \leftarrow argmax(of\ \ classes\ \ c \in C) \sum_{i=1}^{k} \delta(c, y_i)$
where $\delta(a, b) = 1$ if $a = b$ and where $\delta(a, b) = 0$ otherwise.

Figure 2.2 demonstrates the mapping of the ovarian cancer problem in Euclidean space. For problems with more than 3 attributes it is obviously impossible to visualise graphically in this way.

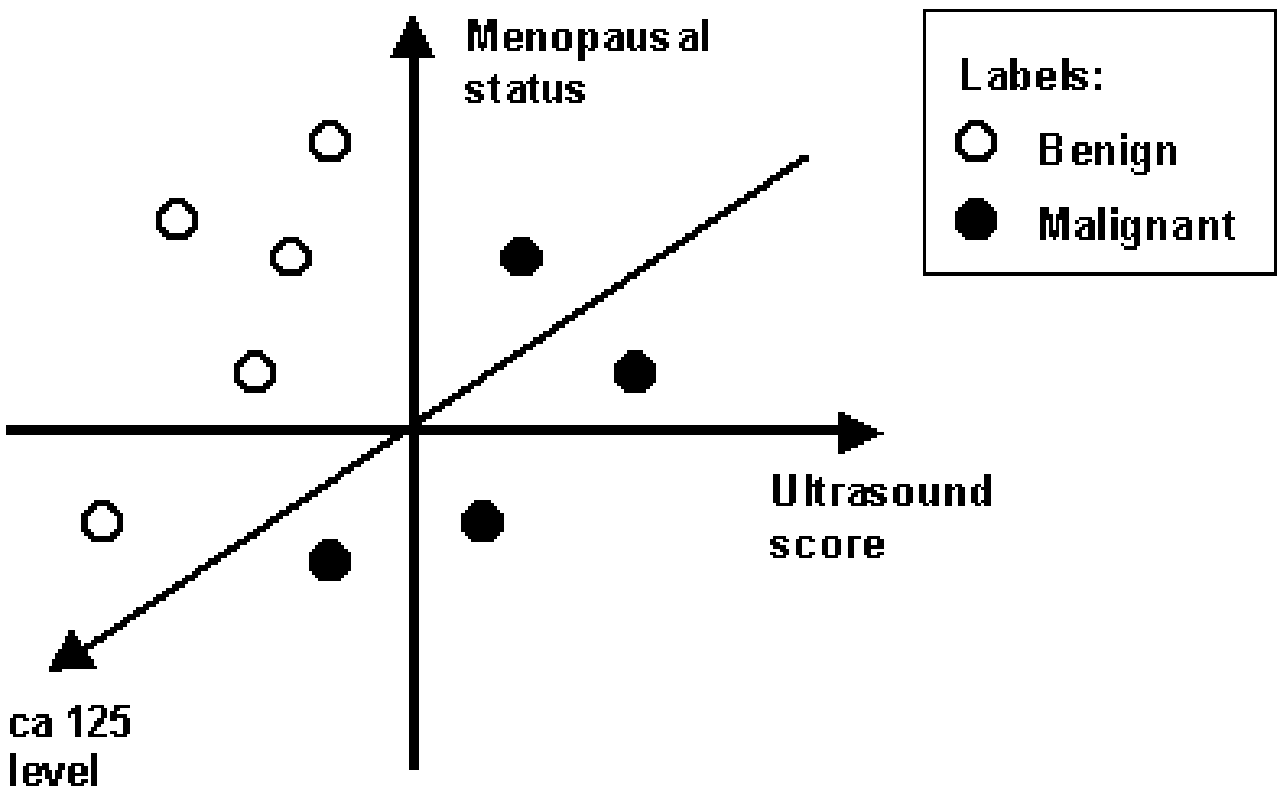

Figure 2.2: **Example of mapping a problem into Euclidean space:** Referring to the example of the ovarian cancer data set, the training examples can be viewed as being mapped into 3-dimensional Euclidean space. Each dimension represents the value of a particular attribute pertaining to each example.

### 2.2.2 Distance-weighted $k$-nearest neighbours (DWKNN)

An obvious extension to $k$-Nearest Neighbour algorithm is to weight the contribution of each of the $k$ nearest neighbours according to their distance to the new test example $x_{l+1}$ being queried, giving greater weight to the closest neighbours.

The decision function can be modified by considering new weighting $w_i$ to the equation.

$$w_i = \frac{1}{d(x_{l+1}, x_i)} \tag{2.4}$$

The function $\hat{f}(x_q)$ can be modified like so

$$\hat{f}(x_{l+1}) \leftarrow argmax(of\ \ classes\ \ c \in C) \sum_{i=1}^{k} w_i \delta(c, y_i) \tag{2.5}$$

### 2.2.3 Computational complexity of the $k$-nearest neighbour rule

The computational complexity of the nearest neighbour algorithm has received a great deal of analysis in the past. Its spacial complexity mainly revolves around the pivotal choices of how to store the training data and necessary calculations efficiently. Its time complexity is dependent on how fast it takes to compute the distances which are needed to search for the closest neighbouring point.

Looking at $k = 1$ nearest neighbours and supposing we have $l$ labelled training examples in $n$ dimensions and we seek the single nearest training example to a test point $x_q$. In the most naive approach we inspect each stored point in turn, calculate its Euclidean distance to $x_q$, retaining only the identity of the current closest one. Each distance calculation is $O(n)$, and thus the search is $O(nl^2)$.

There are numerous techniques that may be implemented to increase efficiency of the algorithm. One such method is by using partial distances, calculating distances using only a subset of the available attributes. Another successful method is to store the examples in the form of a search tree to speed up calculations [2]. This report will not be concentrating on the complexity aspect of the algorithm.

### 2.2.4 Metrics and nearest-neighbour classification

As mentioned earlier, the nearest-neighbour classifier relies on a metric to define the distance between patterns. Although standard Euclidean distance in $n$ dimensions is commonly used, the metric can be far more general, and can use alternative measures of distance to address key problems in classification.

#### Properties of metrics

A metric must have the following four properties. For all vectors $\mathbf{a}$, $\mathbf{b}$, $\mathbf{c}$.
**nonnegativity** : $d(\mathbf{a}, \mathbf{b}) \geq 0$

**reflexivity** : $d(\mathbf{a}, \mathbf{b}) = 0$ iff $\mathbf{a} = \mathbf{b}$
**symmetry** : $d(\mathbf{a}, \mathbf{b}) = d(\mathbf{b}, \mathbf{a})$
**triangle inequality** : $d(\mathbf{a}, \mathbf{b}) + d(\mathbf{b}, \mathbf{c}) \geq d(\mathbf{a}, \mathbf{c})$

It is easy to verify that Euclidean distance formula (Equation 2.3) possesses the properties of a valid metric, however the results of using this particular metric may or may not be meaningful.

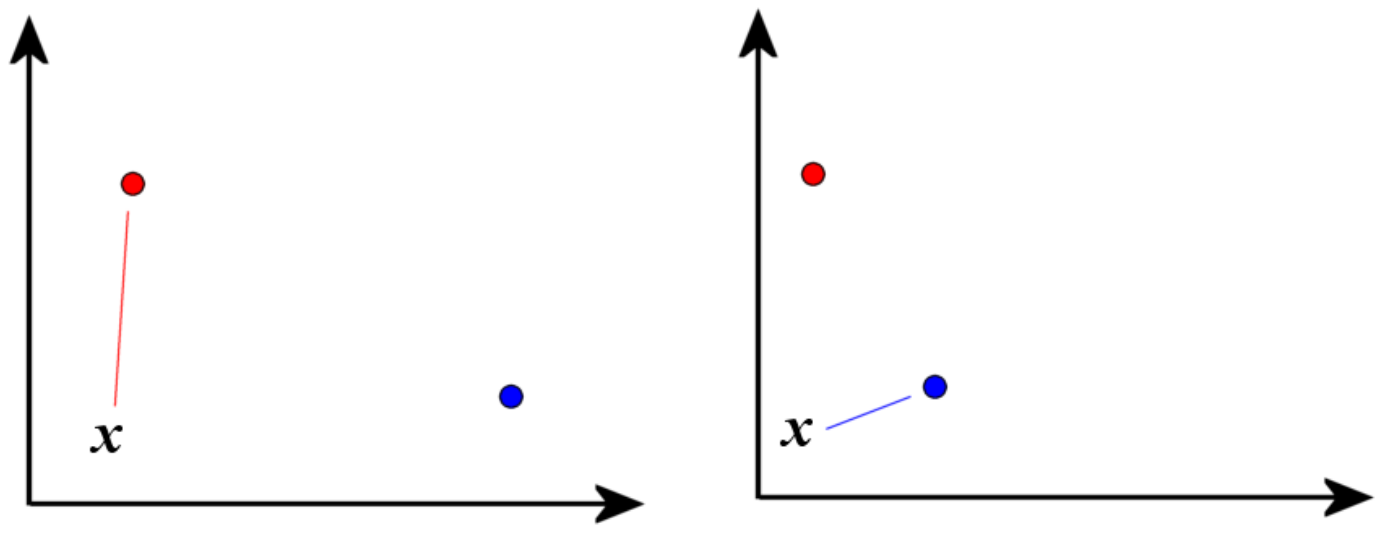

Figure 2.3: **Effect of scaling attributes:** Scaling the coordinates of a feature space can change the distance relationships computed by Euclidean distance metric. Above is an example such scaling can change the behaviour of a nearest neighbour classifier. Originally the nearest neighbour to the query point $x$ would be the blue point, however by scaling the $x_1$ axis by a factor of 3, the red point becomes closest.

If the space is transformed by multiplying each attribute by an arbitrary constant, the Euclidean space relationships in the transformed space may be very different from the original distance relationships. Such scale changes can have a major impact on nearest neighbour classifiers. If there is a large disparity in the ranges of the full data in each dimension, a common procedure is to re-scale all the data to equalize such ranges, and this is equivalent to changing the distance metric in the original space.

### The Minkowski Metric

One general class of metrics for $n$ dimensional patterns is the Minkowski metric also referred to as the $L_k$ norm.

$$L_k(\mathbf{a}, \mathbf{b}) = \left( \sum_{i=1}^{n} |a_i - b_i|^k \right)^{\frac{1}{k}} \tag{2.6}$$

The $L_1$ norm is often referred to as the *Manhattan* or *city block* distance, which calculates the shortest path between $a$ and $b$, with each segment of the path running parallel with each of the $n$ coordinate axis[2].

### 2.2.5 Kernel induced feature spaces

To possibly extend the computational power of the $k$-Nearest Neighbour algorithm, kernel techniques could be used. Kernel representations offer an alternative solution by projecting the data into a higher dimensional feature space for the machine learning algorithms to work in.

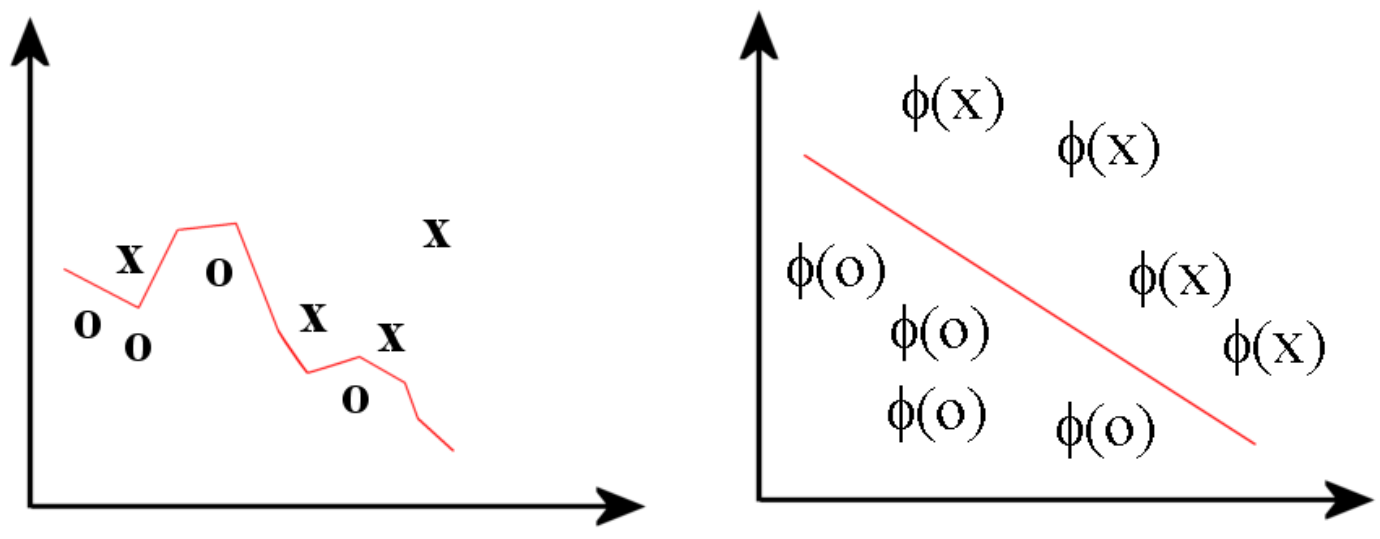

Figure 2.4: **An example of a feature mapping:** A feature map can simplify the classification task

**Definition 1** *A* Kernel *is a function $K$, such that for all $x$, $z \in X$, $K(\boldsymbol{x}, \boldsymbol{y}) = \langle \Phi(\boldsymbol{x}), \Phi(\boldsymbol{x}) \rangle$ where $\Phi$ is a mapping from $X$ to an (inner product) feature space $F$.*

Of course it would not make sense to explicitly create a new higher dimensional feature space, as this would mean that the inherent computational complexity would make the problem unfeasible. We therefore implicitly map into feature space, returning just the value of the calculation of the inner product.

An example of a kernel commonly used is the class of polynomial kernels of the form $(\langle \mathbf{x} \cdot \mathbf{z} \rangle + c)^d$. For example expanding the equation for polynomial degree 2, the following results:-

$$\begin{aligned}
(\langle \mathbf{x} \cdot \mathbf{z} \rangle + c)^2 &= \left( \sum_{i=1}^{n} x_i z_i + c \right) \left( \sum_{j=1}^{n} x_j z_j + c \right) \\
&= \sum_{i=1}^{n} \sum_{j=1}^{n} x_i x_j z_i z_j + 2c \sum_{i=1}^{n} x_i z_i + c^2 \\
&= \sum_{(i,j)=(1,1)}^{(n,n)} \sum_{j=1}^{n} (x_i x_j)(z_i z_j) + \sum_{i=1}^{n} \left( \sqrt{2c} x_i \right) \left( \sqrt{2c} z_i \right) + c^2
\end{aligned}$$

This shows that by adding a constant $c$ to an inner product and squaring the value, we will in fact be returning the inner product of a higher dimensional feature space which contains all the monomials up to degree 2, as well as scaled (by $\sqrt{2c}$) versions of the original features. By setting the constant $c = 0$ the number of features decreases and does not include the original features.

In general polynomial kernels with constants using $n$ features and of degree $d$ map to a space with $\binom{n+d}{d}$ distinct features, and without constant $(c = 0)$ the number of features is $\binom{n+d-1}{d}$.

As discussed throughout this section, the central dogma to the $k$-Nearest Neighbour algorithm is the use of Euclidean distances between examples. To enable us to map implicitly and return the Euclidean distance between vectors in a higher dimensional feature space we must consider the $k$-Nearest Neighbour algorithm in its *dual* form. This is achieved by considering equation (Equation 2.3) for calculating the Euclidean distance in terms of inner products between vectors.

The Euclidean distance between vectors $x$ and $z$ is $d(x,z) = \sqrt{\sum_{i=1}^{n} (x_i - z_i)^2}$

The inner product between vectors $x$ and $z$ is $\langle x \cdot z \rangle = \sum_{i=1}^{n} (x_i z_i)$

$$\begin{aligned} d(x,z) &= \sqrt{\sum_{i=1}^{n} (x_i x_i - 2x_i z_i + z_i z_i)} \\ &= \sqrt{\sum_{i=1}^{n}(x_i x_i) - 2\sum_{i=1}^{n}(x_i z_i) + \sum_{i=1}^{n}(z_i z_i))} \\ &= \sqrt{\langle x \cdot x \rangle - 2\langle x \cdot z \rangle + \langle z \cdot z \rangle} \\ &= \sqrt{K(x,x) - 2K(x,z) + K(z,z)} \end{aligned}$$

We could therefore replace $K(x,z)$ with the respective kernel between these vectors $x$ and $z$, to calculate the Euclidean distance of the vectors in a higher dimensional feature space, without explicitly creating the vectors. Other more complex kernels could be used such as the radial basis, and sigmoidal kernels[3].

### 2.2.6 Remarks on $k$-Nearest Neighbours algorithm

One practical issue in using the $k$-Nearest Neighbour algorithm is that the distances calculated are based on *all* the attributes of the examples (see Equation 2.3). For example, consider a simple problem described by 20 attributes, but where only 2 of the attributes are relevant to determining the classification of the problem. By considering the problem in a full 20-dimensional input space, examples which have identical values for these crucial 2 attributes may still be considered distant because of the other less important attributes values. This is especially important for highly dimensional problems like the abdominal pain database used by Computer Learning Research Centre (Royal Holloway University of London) , which has 135 attributes.

One solution to this problem is to investigate which of the attributes have more influence to the classification, and then calculate a numerical weighting of the attributes importance. This weighting could then be used to put less emphasis on less important attributes by scaling each attribute's dimension in input space. One disadvantage is that this could over-fit the data and force the algorithm to create a decision function that over-accommodates the extreme training examples that are not best representative of the problem.

The number of $k$ nearest neighbours is a parameter to the algorithm which can be adjusted to cope with different noise levels in the training data. For example, if there is noise present in the training data, then analysing to 1 nearest neighbour may cause the algorithm to give erroneous results where the new examples are closest to a training example which itself has the wrong classification. Consequently new examples would be incorrectly classified. If the number of $k$ nearest neighbours is increased, the algorithm will put less emphasis on just the nearest example, and consider a larger sample.

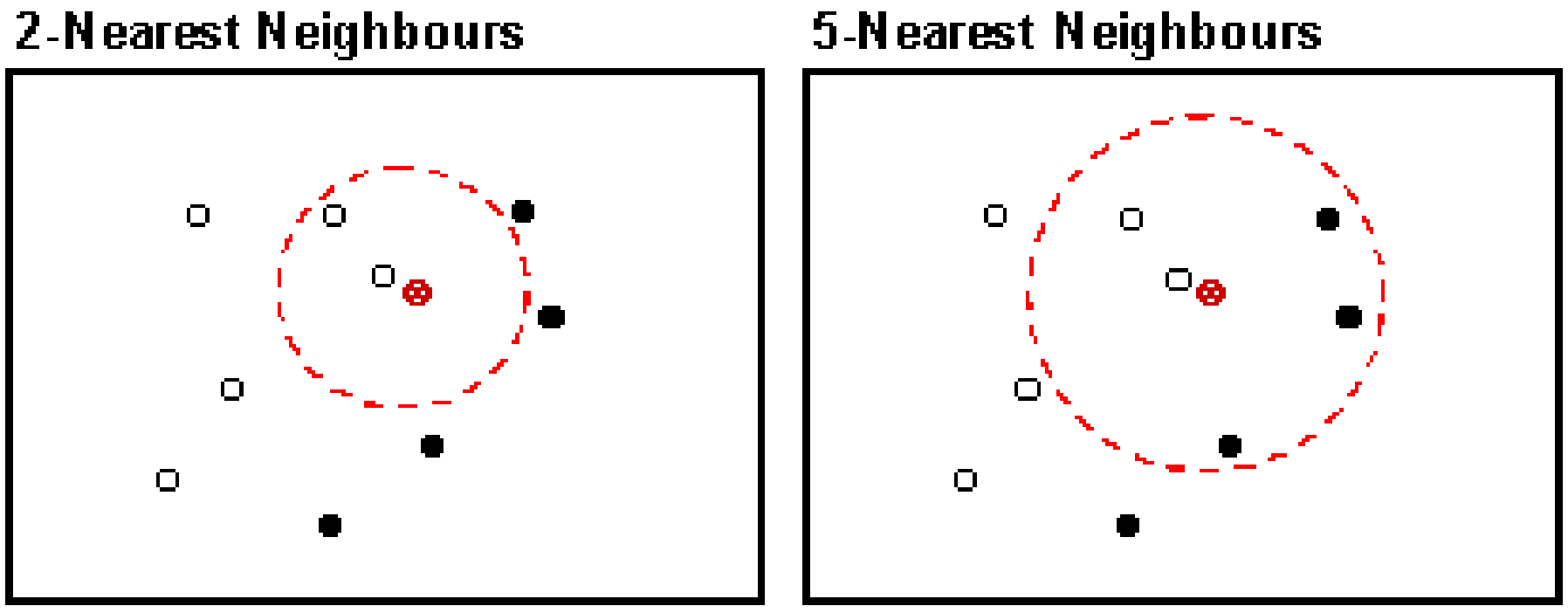

Figure 2.5: **An example of $k$-Nearest Neighbours algorithm in use:** By simplifying the diagram in Figure 2.2 into 2 dimensions and removing the axis (for demonstration purposes only), one can easily how the algorithm works. With 2-Nearest Neighbours the algorithm considers only the 2 closest example to the new query example $x_q$ (marked in red), these are both benign cases, therefore we would classify the new example as benign. However if we analyse to 5-Nearest Neighbours, there are 2 benign and 3 malignant cases, therefore we would classify the point as malignant.

This is demonstrated in Figure 2.2.6 which illustrates how the $k$-Nearest Neighbour algorithm will give different results depending on how many nearest neighbours are being analysed to.

## 2.3 Transductive Confidence Machine - Nearest Neighbours (TCMNN)

This section will gather the theory from previous sections to explain a particular implementation of the TCM using the $k$-Nearest Neighbour algorithm underneath.

### 2.3.1 Formal setting of the problem

We have a training set $\{(x_1, y_1), \ldots, (x_l, y_l)\}$, of $l$ examples, where $x_i = (x_i^1, \ldots, x_i^n)$ is the set of $n$ attributes for example $i$ and $y_i$ is the classification for examples $i$. We consider our new examples in turn, as the $l + 1$ example in the sequence.

As noted earlier, we aim to move from giving 'bare' predictions, to extracting a confidence measure for any given prediction. This measure is calculated by assessing how random a given sequence of training examples are. To calculate the randomness of a given sequence of examples, we extend from the ideas from Algorithmic randomness that were discussed in Section 2.

For our problem we extend from the idea of measuring the randomness of a sequence of binary characters to a sequence of identically structured objects (the training test examples which encapsulate their attributes) each with a label (the classifications) belonging to a finite set of labels.

Using the example of St. Bart's medical data set, each example has a set of attributes describing it, and a possible classification of either benign or malignant. The concept behind our solution to the problem is that we want to predict the most 'typical' (least random) possible completion of the sequence of training examples.

The idea is that if a completion causes the overall sequence to be measured as random, then it will be an untypical completion, therefore unlikely to happen.

### 2.3.2 Creating a strangeness measure using distances

As mentioned earlier, the crucial step when applying transductive confidence techniques to a machine learning algorithm is to develop from it a method to calculate a strangeness value for each example. For the $k$-Nearest Neighbour algorithm the strangeness measure is calculated using the Euclidean distances between examples in the following way.

Let us define the sorted sequence (in ascending order) of the distances of example $i$ from the other examples with the same classification $y$ as $D_i^y$, the $j$th shortest distance in the sequence as $D_{ij}^y$. Similarly let $D_i^{-y}$ define the distances of example $i$ from the other examples with a different classification, $D_{ij}^{-y}$ as the $j$th shortest distance in the sequence.

We assign every example with an individual strangeness measure $\alpha_i$, in the $k$-

Nearest Neighbour case for each example $i$ with classification $j$

$$\alpha_i = \frac{\sum_{j=1}^{k} D_{ij}^{y}}{\sum_{j=1}^{k} D_{ij}^{-y}} \tag{2.7}$$

This measure of strangeness is the ratio of the sum of the $k$ nearest distances from all other classes. This is a natural measure to use, as the strangeness of an example increases when the examples of the same class become bigger or when the distance from the other classes become smaller. Note: because the classification $y_{l+1}$ of the new test example $x_{l+1}$ is withheld from us, the strangeness $\alpha_{l+1}$ has to be computed for each possible classification $c \in C$.

### 2.3.3 Exchangeability conditions

As mentioned earlier, for the results of the computable approximation of the randomness test to be valid, the setting must be valid under the condition of exchangeability $\mathcal{P}_{exch}$. This condition implies that the strangeness values computed for each example must be the same no matter for all possible permutations of the training set.

The beauty of using the $k$-Nearest Neighbour algorithm is that it automatically satisfies these conditions due to the symmetrical property of the distance metric used to create the strangeness values. In contrast when implementing TCM using Support Vector Machine learning algorithm underneath, one must be careful as the strangeness values are dependent on the order in which the training examples are presented, thus failing under exchangeability. This problem can be solved by performing an arbitrary sorting of the data before it is presented to the SVM.

### 2.3.4 The TCMNN algorithm

Using this method of calculating the strangeness values, these values can be fed into the TCM framework, using the function $p(\alpha_{l+1})$ in Equation 2.2 mentioned earlier. The TCMNN Algorithm is given below.

**TCMNN Algorithm**

for $i = 1$ to $l$ do
  Find and store $D_i^y$ and $D_i^{-y}$
end for Calculate alpha values for all training examples
Calculate the dist vector as the distances of the new example, from all training examples
for $i = 1$ to $r$ do
  for $j = 1$ to $c$ do
    for every training example $t$ classified as $j$ do

if $D_{tk}^{j}$ > dist(t)
recalculate the alpha value of example $t$
end for
for every training example $t$ classified as $non - j$ do
if $D_{tk}^{-j}$ > dist(t)
recalculate the alpha value of example $t$
end for
calculate alpha value for the new example classified as $j$
calculate $p$-value for the new example classified as $j$
end for
predict the class with the largest $p$-value
output as confidence 1-2nd largest $p$-value
output as confidence the largest $p$-value
end for

---

### 2.3.5 Optimizing off-line calculations

As outlined earlier on page 2.1.9 in the pseudo code for the TCM, the technique requires that the strangeness values be calculated for all training examples with respect to a new test example for each possible classification. These strangeness values are then used to create the respective $p$-values needed for subsequent predictions and probabilities

This is done to simulate the effect of creating a sequence of training examples adding the new example (with an assigned label/classification) at the end of the sequence, and then computing the randomness of the sequence of examples.

With the $k$-Nearest Neighbour implementation of TCM, it is possible to make a slight computational shortcut in this process by not having to explicitly recalculate the strangeness values for the training examples each time a new example is classified.

The first few lines of the TCMNN pseudo code above state that the values $D_i^y$ and $D_i^{-y}$ must be initially calculated and stored. These values can be seen as the 'partial' strangeness values calculated for each training example. Once these values are calculated, the rest of the TCM process only needs to update these values if the new test example is closer than the existing $k$ nearest distance distances stored in $D_i^y$ and $D_i^{-y}$. For the majority of cases this will mean that the strangeness does not need to be recalculated at all.

These partial strangeness values only need to be calculated and stored once at the beginning of the TCM process and can be cached for later reuse with any other test data set of the same dimensions. This means that when reusing the training data set in a new TCM process, it is possible to load the cached values back into memory to start immediate classification. With the use of larger training data sets this can accelerate calculations required to make the predictions.

This is a significant advantage over implementations of TCM using SVM as these strangeness values must be explicitly recalculated each time for the training examples.

### 2.3.6 Marking results with significance

One advantage of giving confidence and credibility values to predictions is that it is possible to place significance on the results. This is especially useful in applications to medical problems where the accuracy of prediction can be vital.

Traditionally, learning algorithms are marked according to the fraction of test examples that have been correctly classified. With the use of the credibility values computed by the TCM, one can mark the results to a specific significance level $r\%$. This is achieved by considering only examples with credibility values above the set threshold $r\%$, and discounting the predictions of the remaining examples.

Obviously, a certain fraction of examples/predictions will not be considered as significant. Initially this may seem to be a useless process as effectively only a smaller subset of the test examples are given predictions, but on closer analysis this can be very practical in real life applications.

Consider the problem faced by the U.S. Postal service, where thousands of postcodes have to be manually sorted each day. The sorting process is time consuming, expensive in terms of manual labour and may be prone to human error. A machine learning solution to this problem could be developed in order to automate the process of sorting letters according to their postcode by being able to read the noisy inputs of handwritten digits. This potential system would then replace the need for mail to be sorted by hand, saving time and money on labour. The USPS data set consists of image data of roughly 30,000 grey scale images of handwritten digits taken from U.S. postcodes.

This data set has been thoroughly tested using many different algorithms in the past, achieving an overall accuracy of prediction of 96.0% in results by Vapnik [11]. This performance might be seen as a triumph of machine learning, however if we consider the practical usage of this result to the original problem we find a fundamental weakness to the solution.

Assume that the U.S. postal service requires the learning system to sort through roughly $10^6$ letters in one day, and that each letter has 6 characters written in a postcode. The probability of correctly classifying each character correctly is $P(\bigoplus) = 0.96$, therefore the probability of correctly classifying the whole postcode is $(P(\bigoplus))^6 \approx 0.783$. Therefore we would expect that roughly 20% (200,000 letters!) of the mail to be incorrectly sorted by the learning system each day. These errors could then cause further complications that could cost the postal service money to rectify. This is indeed the reason why none of the machine learning solutions are actively employed by the U.S. postal service, and that the mail is still sorted by hand.

A better solution to the problem would be to develop a learning system using the TCM framework and set a significance level that would cause the accuracy of prediction to 100%. This would of course mean that a separate pile of mail would be left for manual sorting, but the postal service could be confident that the mail actually sorted by the learning system would always be done correctly. This partial breakdown of the problem could then mean that less labour is needed to sort the mail, potentially saving the postal money in terms of salaries.

# Chapter 3

# Design & Implementation

The original goal for the project was to develop an ovarian cancer screening system to be used by St. Bartholomews Hospital in London that used transductive confidence machine learning techniques to make predictions with confidence. The scope of the system was then extended to be applied to any given data set, including the analysis of image data.

After implementing the TCMNN (Nearest Neighbours) algorithm, I developed many front end user interfaces to allow ease of use. The output of the programs presented the results of testing the TCMNN algorithm, and statistical performances on different data sets in HTML files.

While testing various data sets, I started to develop very interesting results, and concentrated on exploring different aspects that affected the performance of the TCMNN algorithm. Although my programs allow configuring of different parameters for the TCMNN algorithm, it became tedious to keep building in specialised testing functionality into the system.

Unfortunately the system could not enable the user to run the systems programs directly through their web browser. The user must therefore download the systems programs to their machine along with a distribution of the javac compiler.

On this basis, I decided to concentrate less on the TCMNN as the main focus of the project, and concentrated more on performing experiments with different settings and learning algorithms. When running the tests I developed simple test harness programs that were able to output the results and statistical performances to basic text files.

This chapter will provide an overview of the design and implementation choices that were made during the course of the project.

## 3.1 Overview of design and implementation

### 3.1.1 Hardware and software choices

The programs were intended to run on standard desktop PC architecture. When testing the system with large data sets (of around 23,000 examples), it appeared to make more sense to concentrate on the amount of space required by the programs to process the data, rather than trying to optimise the calculation time.

All programs were implemented using the current java distribution (Java Development Kit 1.3.1). This benefits of which were as follows:-

1. Cross platform compatibility - The programs are compiled into byte code (class files). These files are then interpreted by the java virtual machine (JVM) that runs on the users native architecture. JVM software is freely available for all the major platforms, such as Windows PC's, Linux and Apple OS.

2. Object orientated software design - The java language is a true object orientated programming language. This enables the problem to be easily broken down in smaller component objects and classes.

3. Packages - Can easily set up libraries of common used classes and procedures into packages. These classes can be structured into a hierarchy of components.

4. User interface - Swing interface components allow professional looking user interfaces to be built up. This can make the system easy to use by using a familiar windows style interface.

5. Readily available extensive library of core utility procedures, such as easy reading and writing to files, string handling, abstract data structures and mathematical functions.

### 3.1.2 Applet vs. application

I had hoped to design the system as an applet that could run in the users web browser. From my research into the problem the only restrictions on this method was that applets were not able to have full read write access to the users machine due to security restrictions [24]. The following lists suggestions to solve this problem:-

1. Upload the data set files to the server, and use server side technologies such as Java servelets or CGI/Perl scripts to read the file and communicate with the applet. This solution is the most complex to implement.

2. Change the security settings on the users machine so that all applets have access to the users machine [29]. This would make the use of the system far more complicated for the user. Another problem is that users on a network may not have the permission to alter these settings.

3. Use digital certificates to enable the web browser to grant access to users machine. This would be the simplest way for the user to access the system. One complication is that each web browser authority requires different certificates to enable access to each web browser. These certificates must be updated annually [30].

Each of these solutions were far too complicated and expensive to implement so the decision was made to keep the system running as an application.

### 3.1.3 Structuring of classes

To make the development of the system easier, I structured some of the more useful components and algorithms used in the system into 'packages'. This made the larger programs easier to develop, as once the structure was set up, any updates made to the core classes would also be updated for all the other systems that were dependent of the same component.

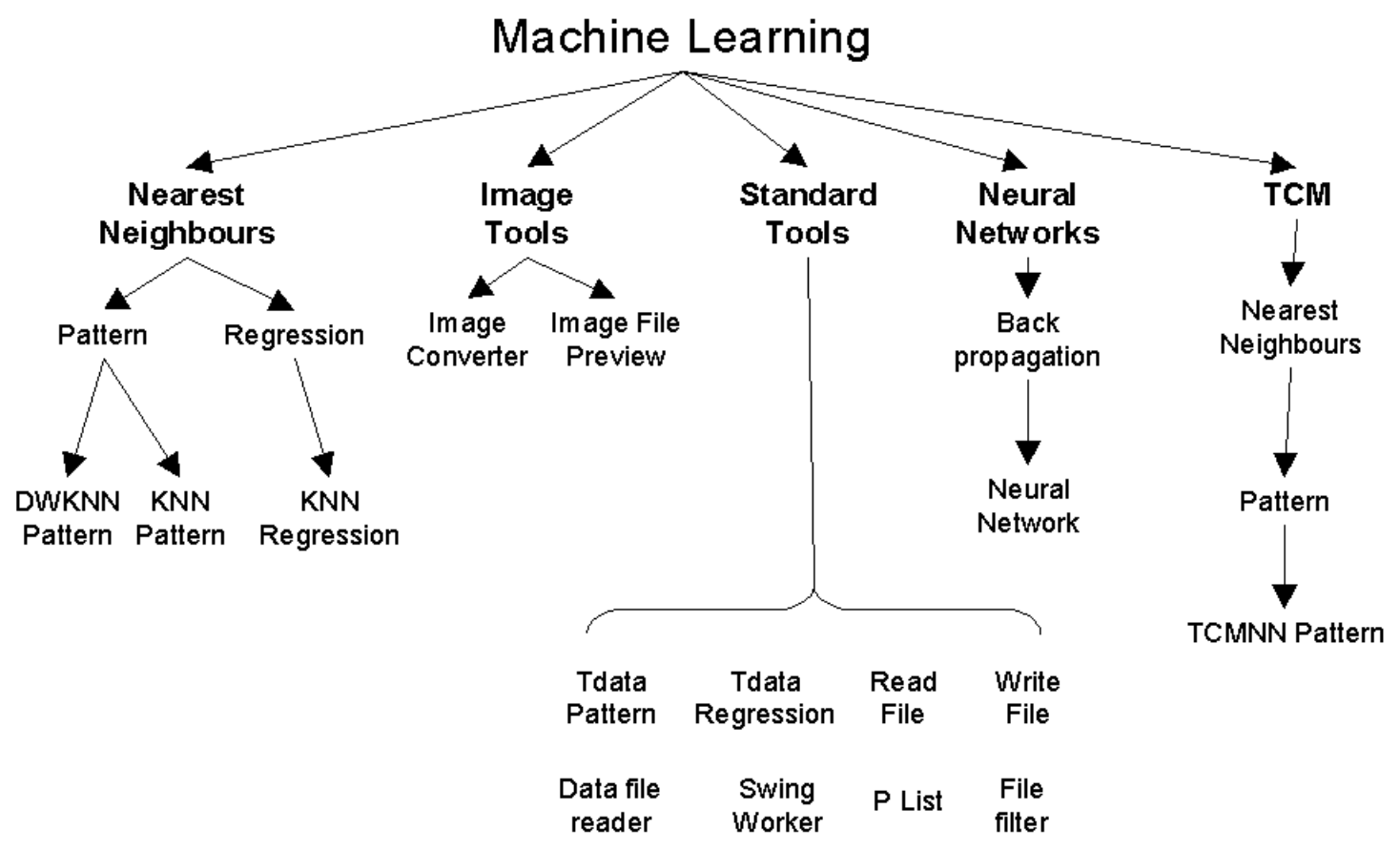

Figure 3.1: **Diagram of packages of class files:** This is a simplified diagram of the package hierarchy of the classes used in my system.

A simplified diagram of how the classes were structured into packages is given in Figure 3.1. The hierarchical structure meant that new branches could be added easily if new learning algorithms were to be implemented. The full library of programs developed in this project can be downloaded from my web site:
`http://www.rhul.ac.uk/users/zjac048/3rdyearproject/`

## 3.2 TCMNN Algorithm implementation

Prior to development of the TCMNN system it was necessary to implement the TCMNN algorithm. This section gives an outline of the major steps taken during the implementation process.

### 3.2.1 Construction of the TCMNN algorithm

The TCMNN algorithm was implemented as outlined in the pseudo code earlier on page 33. Implementing in Java made it easy to develop the algorithm as an object that could be used by many programs.

During system development other classes were developed to handle low level functionality of the algorithm such as methods for storing and sorting the distance calculations (KDoubleBuffer), keeping track of the $p$-values (PList) and so on. Simple classes were developed to represent training and testing data in the system (TDataPattern, TDataNNet, TDataRegression).

Finally general classes were developed to handle the basic functionality for reading (ReadFile) and writing (WriteFile) to files. These classes were crucial for the ability to read the information from data set files, and to output results and statistics of the TCMNN algorithms performance.

### 3.2.2 Testing modes available

The TCMNN algorithm can be tested in the following ways:-

1. Leave one out test - This test requires that a single combined training and test data set to be provided to the TCMNN. This test selects an example from the combined data set and tests it using the remaining examples, repeating the process for all examples in the data set.

2. Random split test - This test requires a single combined data set. It randomly extracts a specified number of examples from the combined data set and constructs a new test data set. The remaining examples are then used as a training data set. Both randomly split data sets are then tested using the normal separate training/testing method.

3. Separate test - This mode of use requires two data sets, one for training and the other for testing. The program selects one example at a time from the test set and uses the complete training set to make the predictions with the TCMNN. This process is repeated for all the examples in the test set.

### 3.2.3 Distance measures available

The TCMNN algorithm can be run using the following distance measures between vectors:-

1. Euclidean distance - This is the distance metric commonly used with the TCMNN algorithm.

2. Minkowski metric - This metric can be computed to any positive real power $d$. Setting $d = 2$ is equivalent to the Euclidean distance.

3. Polynomial kernels - The standard Euclidean distance can be calculated between higher dimensional feature vectors using an implicit polynomial kernels mapping. These polynomial kernel mappings can be used with various degrees and constants.

### 3.2.4 Outputting of performance to file

The TCMNN algorithms' performance on a data set can be assessed in the following ways:-

1. Normal assessment - The predicted class and real class are compared for each test example. From this an accuracy of prediction (percentage %) is calculated for overall (all classes combined) and for individual classes. Basic averages of the confidence and credibility are calculated. These values are then structured in a basic output to a text file.

2. Marking with significance - This assessment involves specifying a set threshold significance value. Test examples which have predicted credibility values less that the significance threshold are ignored and termed 'not classified'. The normal assessment is then applied to the remaining examples. The values for overall accuracy, individual class accuracy and percentage of examples not classified are output to a text file.

These simple assessment routines are very useful when creating the basic test harness programs used to the run the experiments as seen in the next chapter.

### 3.2.5 Additional TCMNN functionality

The TCMNN algorithm has additional functionality outlined below:-

1. Save cached TCMNN calculations for a training set - As mentioned previously page 33, the calculations of the distances and partial strangeness values can be used to accelerate computation in situations where the TCMNN algorithm is repeatedly used with the same training set. After performing the calculations using a training set these values can be saved to a cache file for later re-use.

2. Load cached TCMNN calculations for a training set - When the TCMNN algorithm is performed on a data set which has already had its calculations cached, the cached values can be reloaded back into the algorithm. Once cached values are loaded, the TCMNN algorithm can skip repeating calculations of distances between the training set and start classifying the new examples directly.

3. $K$ nearest setting - The TCMNN algorithm can be performed to different numbers of $k$ nearest neighbours. The size of the smallest class in the training set is the largest number which $k$ can be set to.

## 3.3 2D TCMNN example application

### 3.3.1 Purpose

This was one of the first programs developed. This program developed my skills in using java swing components to create interactive interfaces. It gave me experience of handling basic windows and mouse handling events. In addition, I was able to further my understanding of the workings behind the TCMNN algorithm.

### 3.3.2 Screen shots and description of interface

Less emphasis was placed on the interface. The key issue was to ensure that the system was functional, giving easy to interpret results. Figure 3.2 gives a brief explanation of the main components that make up the TCMNN 2d example application.

The options menu is a simple drop down menu that allows the user to:

1. Set the number of $k$ nearest neighbours to analyse to.

2. Set the distance metric as either using the standard Euclidean distance, polynomial kernel mapping, or Minkowski distance. The polynomial kernel option allows the user to set both the degree of the kernel $d$ and a constant $c$. The Minkowski metric has the option of what power $d$ it is set to.

The active training area is where the training and test points are added and calculated. The user clicks within this area with the left and right mouse buttons to add training vectors of each class. Each training point is identified by a red and green circle.

The progress bar indicates to the user how long they must wait for the calculations of the TCMNN process to complete. Once finished, the active training area will update by colouring each unspecified point respective to its predicted class. The intensity of the colour reflects how confident or credible the prediction is. Darker black indicates low confidence or credibility values close to zero, whereas bright green or red colours indicate high values.

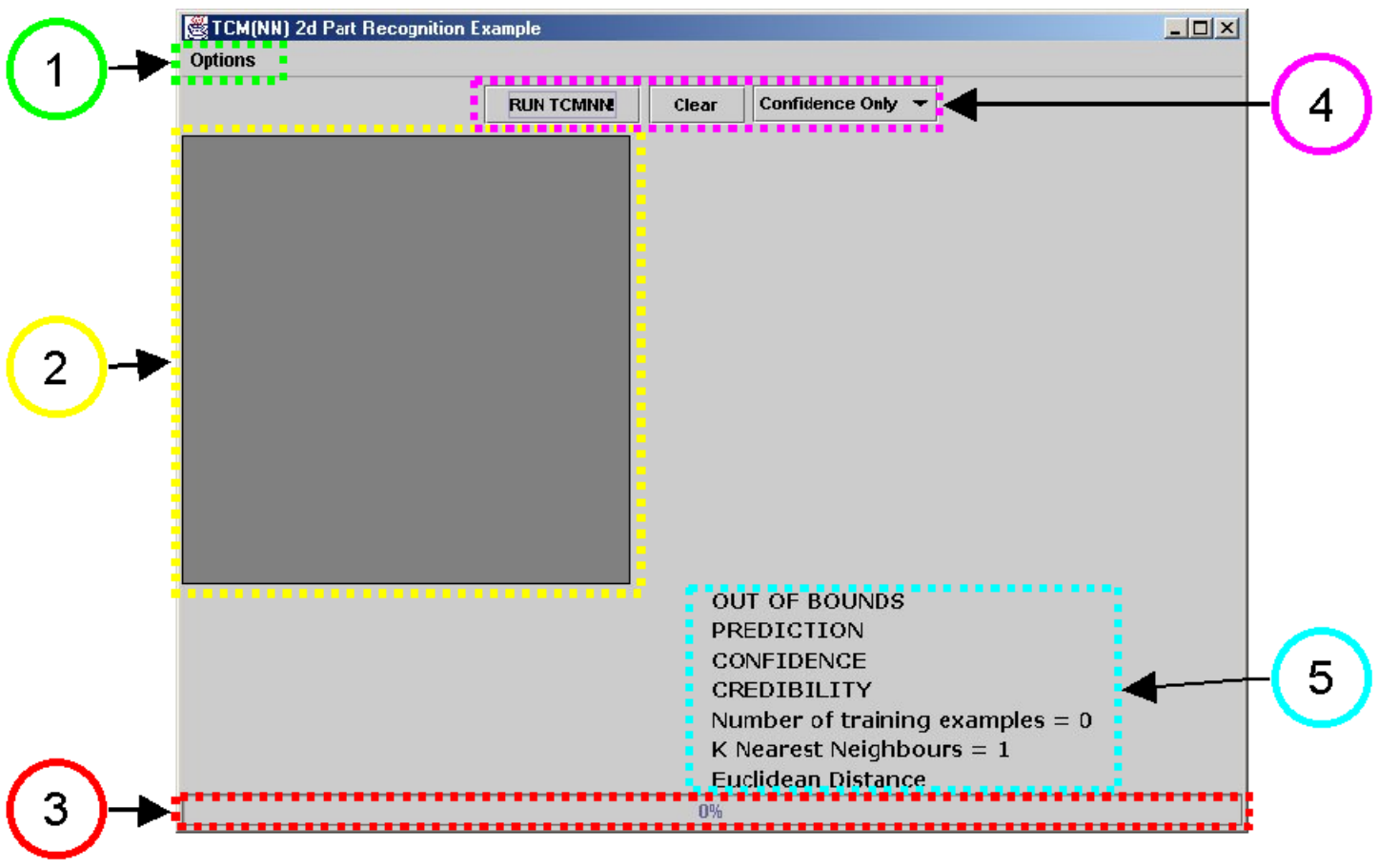

Figure 3.2: **Explanation of TCMNN 2d example application:** This application has 5 main components as labelled above :
1) Options menu
2) Active training area
3) Calculations progress bar
4) Main controls
5) Setting details

The main controls of the system offer the core functionality of the system. New TCMNN calculations can be initiated by clicking the 'RUN TCMNN' button. The clear button removes all the training and test points, so that new points can be specified. The confidence and credibility combo box allows the user to toggle between viewing the confidence and credibility regions predicted by the TCMNN.

The settings area seen in 5) of Figure 3.2 shows the currently selected options for the TCMNN. When the user hovers the mouse over the active training area, the current pixel coordinates and the confidence/credibility values are displayed in the settings area.

### 3.3.3 How to use the application

The system demonstrates how the TCMNN algorithm works on a simple 2 dimensional, binary classification problem. The user clicks the left and right mouse button within the testing area to add training points for each class, displayed as green and red circles.

The user can also specify additional parameters of which particular distance metric or the number of K nearest neighbours that they wish to analyse to.

Once the user has finished adding training points, they can select the 'RUN TCMNN' button to start the TCMNN algorithm training with the points clicked by the user, and testing with the remaining unspecified points. When these calculations are finished, the results are displayed graphically.

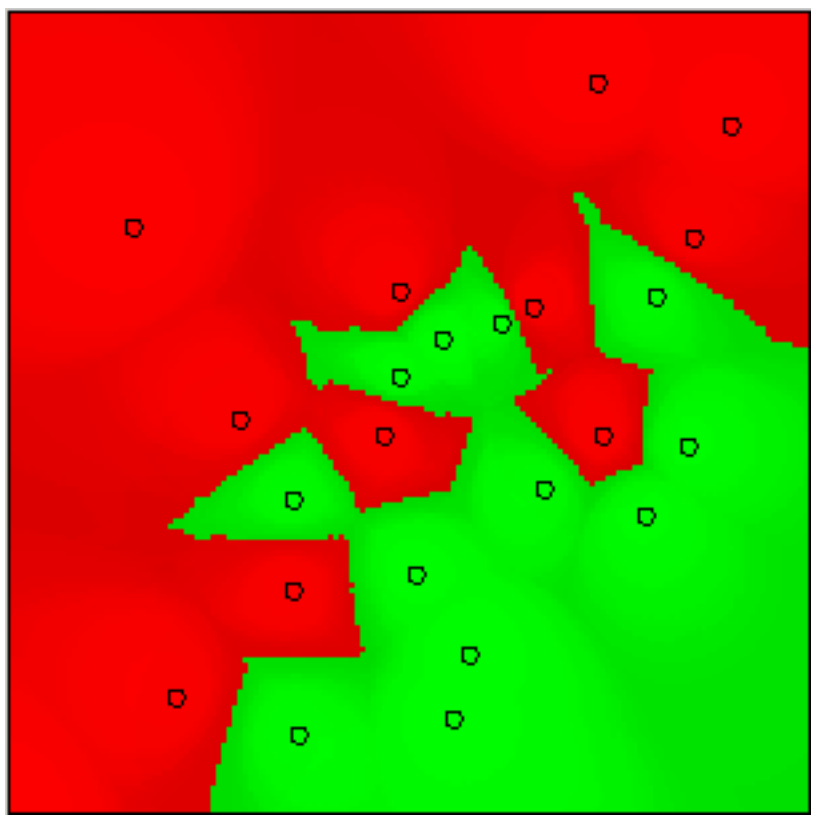

Figure 3.3: **Example of output results of the TCMNN 2d example application:** The training points are the circles outlined in black. This particular plot is taken from a screen grab of the confidence regions plot, using standard Euclidean distance with 1 nearest neighbour. Results of further experiments with artificial data sets using this application can be found in appendix B.

Once calculations are complete the user can either view the prediction in terms of confidence or credibility regions. An example of the output of the program can be seen in Figure 3.3. The user can also keep the existing training data points and rerun the calculations with different parameter settings. New training data points can be added to the problem, then rerun to see how the decision boundaries are affected.

Although the functionality of the system is very simple, the different features can be combined to run more complex tests with the data as shown in appendix B.

This application is very simple and does not require the ability to read and write to files. Because of this, the application could easily be converted at a later date into an applet that would run in a web browser.

## 3.4 TCMNN pattern classification system

This was the main system developed for my project. The system comprises of several applications offering different types of tests with the TCMNN algorithm. The system also contains a supporting application that structures data sets into predefined file formats. A more in depth explanation of how to use these applications can be found in the user guide appendix G.

The program was developed as an application that could be downloaded and run on the users machine. The key aim in designing the system was to make it as user friendly as possible for use by individuals not familiar with machine learning. This aim was achieved using the Java swing GUI components to create a simplistic windows interface.

### 3.4.1 Data set file structure

To use the TCMNN testing applications it is necessary that the data sets are formatted correctly so that the data can be read successfully.

#### Text file format

The data set must be initially formatted as a tab delimited (or other fixed delimiter) text file. This text file must be structured in the following format.
**First line of text file:**
$1^{st}$ Attribute Name $< TAB >$ $2^{nd}$ Attribute Name $< TAB >$ ... $n^{th}$ Attribute Name $< TAB >$ Output Name $< LINEBREAK >$
**Subsequent lines of text file:** $1^{st}$ Attribute Value $< TAB >$ $2^{nd}$ Attribute Value $< TAB >$ ... $n^{th}$ Attribute Value $< TAB >$ Real Class (If known) $< LINEBREAK >$

This first line provides the names for the respective attributes. The subsequent lines provide the attribute and class details for each example. Each line of the file represents the details of an example. The real class can be left out if it is not known, as used in the batch query application. Each line must be terminated by a line break. Every attribute value must be numeric, and the class must be represented as an integer from 0 to a finite number $n$.

Adding relevant attribute names will make the results from testing on the data set more easily readable to the user. Figure 3.4 shows an example of the text file for the ovarian cancer data set. The first line of the file shows the attribute names that have been provided. In this example the relevant attribute values are aligned under the relevant attribute names.

The benign cases are represented as class 0, and malignant cases as class 1. Also a relevant output name has been added to describe what the classifications are, in this case the output/class is best described as the *diagnosis* for a particular patient.

```
Menopausal status      Ultrasound Score      Pre-op CA125    Diagnosis
3          0           10.9       0
1          1           25.1       0
1          1           27.2       0
```

Figure 3.4: **Example layout of a data set text file**

For details on how to create a file in this format using commercial software packages such as Microsoft Excel, please refer to Figure G.2 in the user guide.

**Data file format**

The first few lines of the data file format consist of a header which presents details about the data set with pairs of consecutive lines in the following format:

1. a logical markup tag encapsulated in [...] brackets identifying a particular detail about the data set.

2. the respective values of the detail on the next line.

This is done to make the file easy to read by the user and by the TCMNN testing applications. A detailed description of the tags and the structure of the file is given below.

**Header of data file:**
[*NUMBER_OF_EXAMPLES*]
Number of examples present in this data file.
[*NUMBER_OF_ATTRIBUTES*]
Number of attributes that are used in this data file.
[*NUMBER_OF_CLASSES*]
Number of different classifications used in this data file.
[*PRESENCE_OF_CLASSES*]
Whether the classes are known for this data file.
[*CLASS_NAMES*]
Details of the respective names for each class.
[*IMAGE_FILE*]
Whether this file contains image data.
[*PRESENCE_OF_ATTRIBUTE_NAMES*]
Whether the attribute names are present for this data file.

The rest of the data file is exactly the same as the original text file that was used to create the data file.

**Subsequent lines of data file:** $1^{st}$ Attribute Name $< TAB >$ $2^{nd}$ Attribute Name $< TAB >$ $\ldots$ $n^{th}$ Attribute Name $< TAB >$ Output Name $< LINEBREAK >$ $1^{st}$ Attribute Value $< TAB >$ $2^{nd}$ Attribute Value $< TAB >$ $\ldots$ $n^{th}$ Attribute Value $< TAB >$ Real Class (If known) $< LINEBREAK >$

```
[NUMBER_OF_EXAMPLES]
259
[NUMBER_OF_ATTRIBUTES]
3
[NUMBER_OF_CLASSES]
2
[PRESENCE_OF_CLASSES]
true
[CLASS_NAMES]
Benign	Malignant
[IMAGE_FILE]
false
[PRESENCE_OF_ATTRIBUTE_NAMES]
true
Menopausal Status	Ultrasound Score	Pre-op CA125	Diagnosis
3.0	0.0	10.9	0
1.0	1.0	25.1	0
```

Figure 3.5: **Example layout of a data file**

Comparing Figure 3.4 to Figure 3.5 one can see that the data file format only differs from the raw text file by the inclusion of the tags that encapsulate the data sets' details. Having these details encapsulated in the file means that the user does not have to re-enter them every time that they use the data set.

### 3.4.2 Data file creator application

This application was designed to ensure that the data set files were structured in the correct format to be used by all the subsequent TCMNN testing applications. The only initial requirement for this system is that the data set must be initially presented as a tab delimited text file in the format described earlier.

As with all the applications developed for the system, a simple tabular paned interface was implemented. Each tab in the application groups together relevant questions to the user. A screen shot of the Data file creator application in use is shown in Figure 3.6.

To use the program the user can simply click through each tab from left to right filling in the relevant details required by the system. By designing the system in this way the user is logically guided through each step with plenty of on screen assistance.

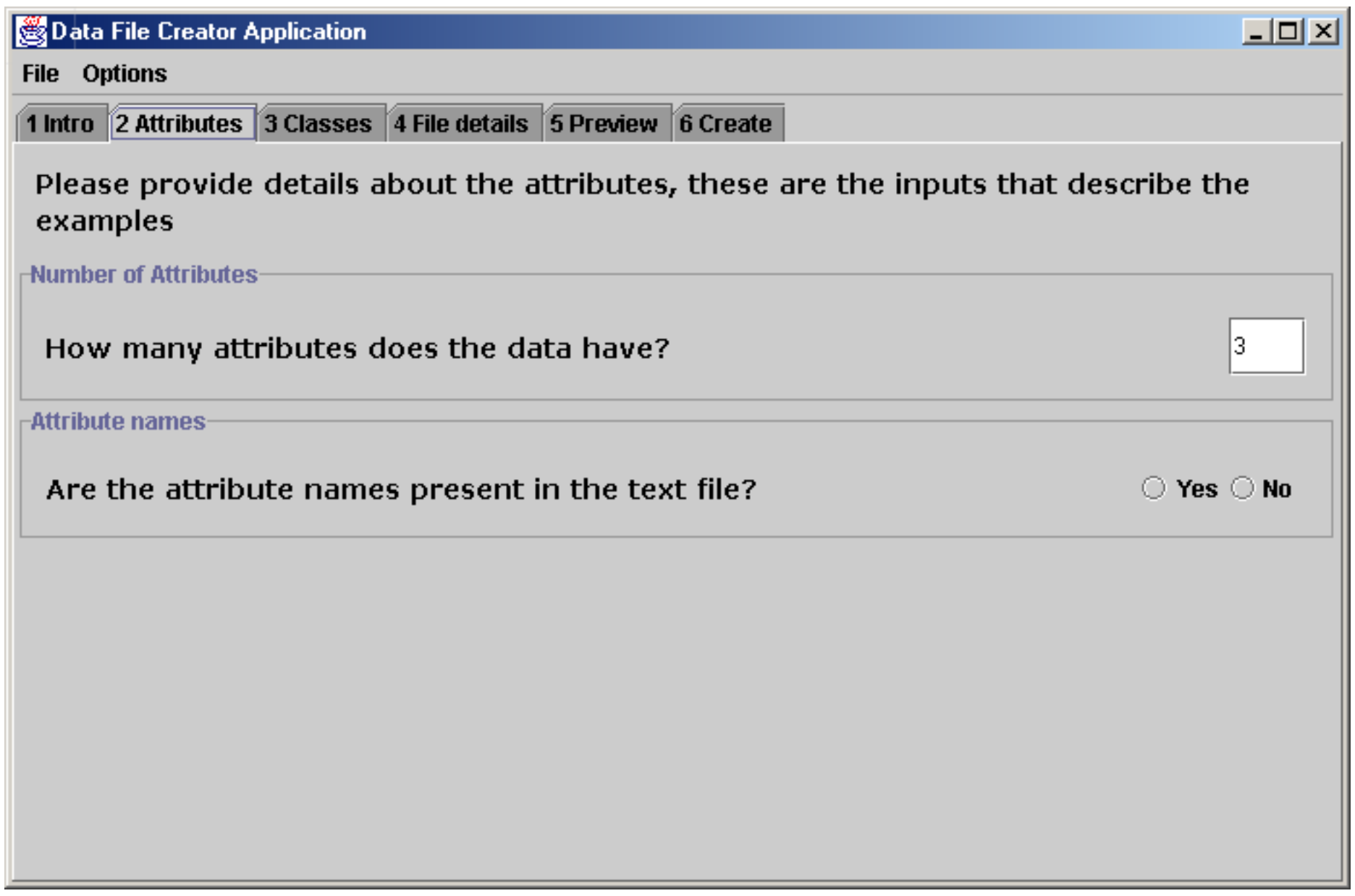

Figure 3.6: **Screen shot of the data file creator application in use:** This particular screen shot shows the tab used to enter the details about the attributes of a data set.

The application retrieves the following information from the user:

1. Details about the attributes that describe the data set.

2. Details about the possible classifications.

3. Details about the number of examples in the data set.

4. Location of the raw text file version of the data set.

The output of the application creates a structured data file as mentioned earlier. The resulting file is created with a ".data" extension and can be read by all of the subsequent TCMNN testing application. An outline of how this application and the other TCMNN systems work together is shown in Figure 3.7.

### 3.4.3 TCMNN testing applications

4 main systems were developed and much of the interface is consistent for each of the systems. All 4 applications were implemented using the TCMNN algorithm object (mentioned earlier) to perform the necessary calculations in the background. The basic functionalities of the applications are as follows:

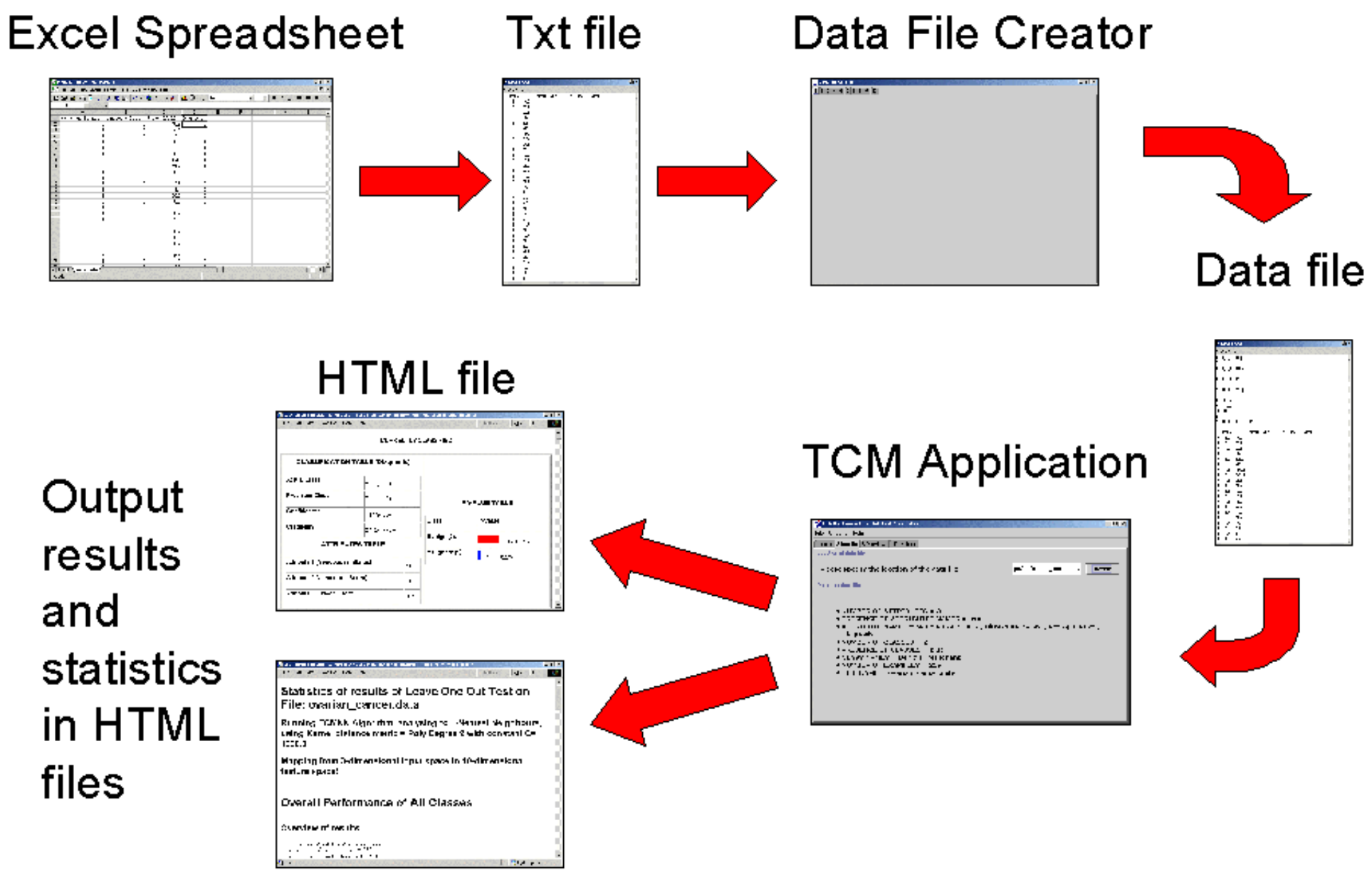

Figure 3.7: **Overview of how the TCMNN testing system works:** Initially the user creates/structures the data set text file in a spreadsheet package such as Excel. The data file creator application then uses the text file as well as queries from the user for more details about the data set. All the details are then saved in a structured data file format. The data file is then read by one of the TCMNN applications, and the user chooses options on how to analyse (selecting distance metrics, number of nearest neighbours etc). The output of results and statistics of performance are then given in HTML format as seen in appendix F.

1. Leave one out test application - This application takes a single data file which has known classes, and performs a leave one out test on the data.

2. Separate test application - This application requires 2 data files, both with known classes. This application then performs a separate training and test experiment with the data files.

3. Manual entry test application - This application requires 1 data file with known classes for training. The application constructs a table for the user to manually enter new examples to be tested.

4. Batch query of examples (where label not known) application - This application is similar to the separate application except the training data file has known classes and the test data file does not.

### 3.4.4 Output of programs

The TCMNN testing applications output two different types of files. Both files are structured as HTML files that can be viewed using any standard web browser. Keeping the output in a web based format makes the files easily portable across platforms. An example of the HTML files created can be seen in Figure 3.8

Both files start with a common header format providing the following details about the TCMNN test performed:-

1. The type of testing mode being used (eg. Leave one out, Separate and so on).
2. The number of $k$ nearest neighbours analysed to.
3. The particular distance metric being used.
4. The names of the data file(s).

#### Results file

This file displays the predictions of the TCMNN algorithm for each of the examples in the test set. The results are formatted using simple HTML tables. The file provides the following details:-

- actual class (if classes known for test set)
- predicted class
- whether the example is correctly classified (if classes known for test set)
- confidence values
- credibility values
- $p$-values for each class (displayed graphically as bar chart)
- attribute values (can be disabled)

#### Statistics file

This file outputs the statistical performance of the TCMNN test. This file provides the following details:-

- overall and individual class performance assessment.
- the number of training and testing examples provided in the data sets.
- average and confidence and credibility values.
- confidence and credibility histograms.

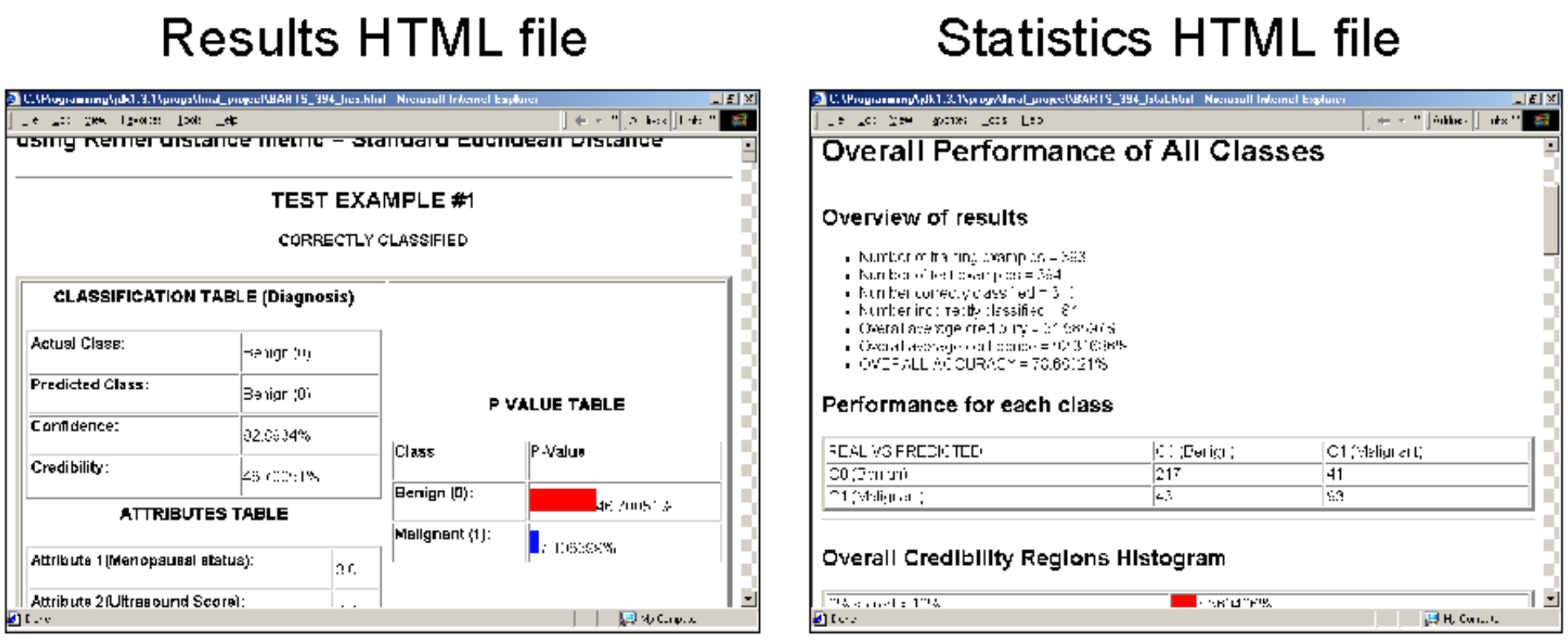

Figure 3.8: **Examples of the results and statistics HTML files generated by the TCMNN testing applications**

## 3.5 TCMNN pattern classification system (image data sets)

This system was not developed as fully as the other systems were during the course of this project. Rather, it was an extension to principles learnt from the original TCMNN system.

### 3.5.1 Image data set file structure

The image data set attributes are the numeric interpretation of each pixel in the images. The header of the image data file is identical to that of the normal data file except for an extra logical tag [*IMAGE_DIRECTORY_BASE*]. This tag defines where the directories of images can be found for each class.

Another crucial difference to the data set structure is that the first entry for each example specifies the filename of each individual image. These details are required for linking in the images to the HTML results file mentioned later.

### 3.5.2 Image data file creator application

This application is used to convert file directories containing images into an image data file which can in turn be used with the TCMNN image testing applications.

Once the user has provided the location of each image directory, the application reads in each image from the directory and computes a numeric pixel colour value, outputting it to the data file. The application also offers the functionality of being able to re-scale the images to fixed dimensions (specified by the user). The data file

```
[NUMBER_OF_EXAMPLES]
110
[NUMBER_OF_ATTRIBUTES]
361
[NUMBER_OF_CLASSES]
2
[PRESENCE_OF_CLASSES]
true
[CLASS_NAMES]
Dave     Sian
[IMAGE_FILE]
true
[PRESENCE_OF_ATTRIBUTE_NAMES]
false
[IMAGE_DIRECTORY_BASE]
C:\Programming\jdk1.3.1\progs\project\images\combined\david …
DCP01514.JPG     105     112     113     118     128     …
DCP01515.JPG     147     161     163     177     184     …
DCP01516.JPG     143     154     156     186     196     …
```

Figure 3.9: **Example layout of a image data file**

output by the application has a ".data" extension.

Figure 3.10 shows how the image data file creator is used in conjunction with the rest of the TCMNN image testing applications.

### 3.5.3 TCMNN Image testing application

These applications are identical to the interfaces of the original TCMNN testing applications mentioned earlier. The only difference between the two systems is the way the results are displayed. At present the system only offers the ability to perform a leave one out test on image data files.

### 3.5.4 Output of application

The output of the TCMNN image testing application offers the same detail content as given by the normal TCMNN testing applications. The only difference can be seen in the results file, where the original images are linked into the file.

It would not make sense to view the individual numeric pixel values for each example as this would make the output unreadable (a 16×16 image would have 256 attributes!). Instead, the application uses the file locations of the images (encapsulated in the image data file) to link in the original image. An example of the results file generated can be seen in Figure 3.11.

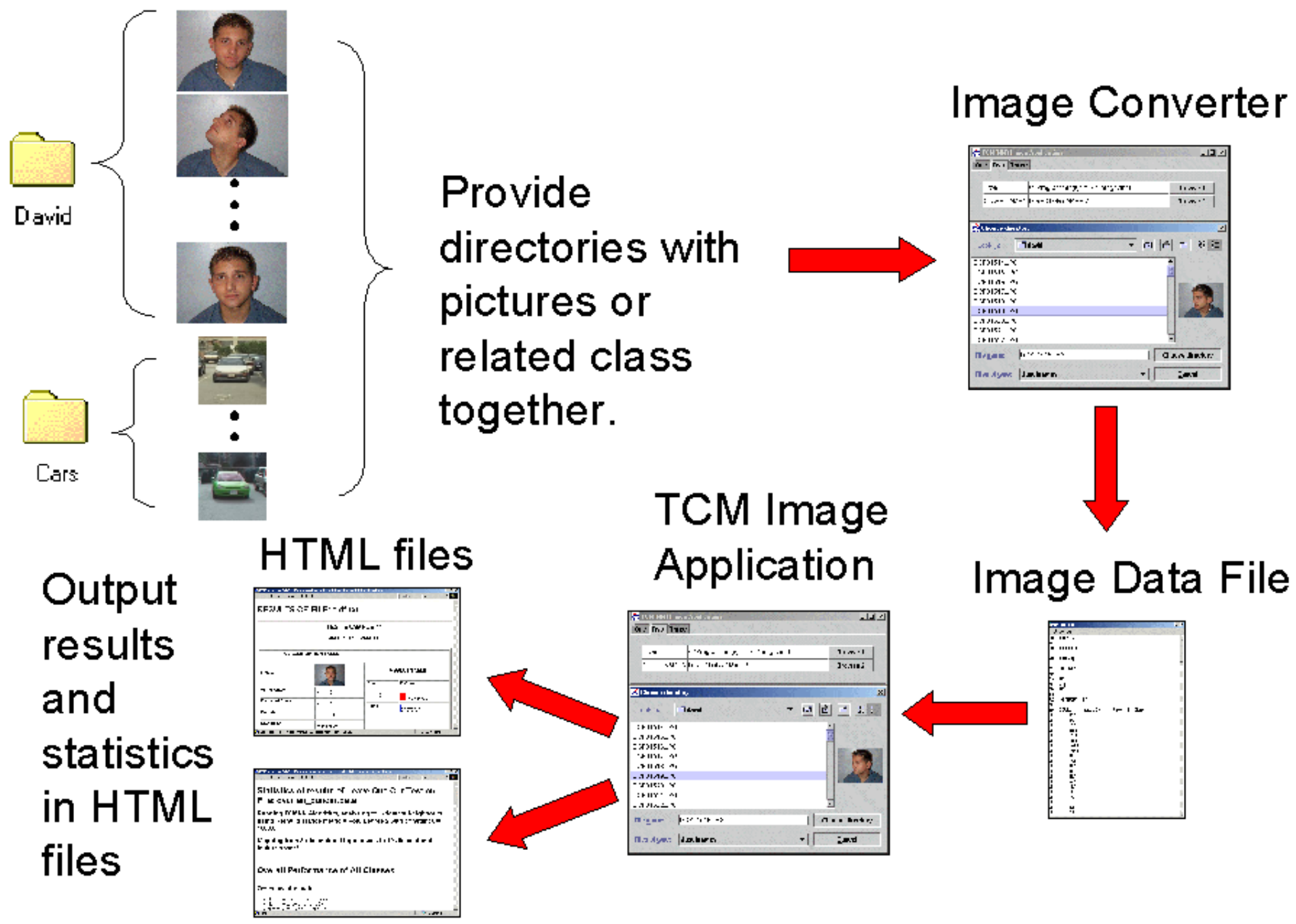

Figure 3.10: **Overview of how the TCMNN image testing system works:** Initially the user must keep their images for each class in separate files directories (eg. pictures of David in David folder, pics of cars in cars). These images are then converted into numerical pixel data using the Image Data File Creator application. This then creates an image data file to encapsulate all the information about the image data set. The data file can then be read by the TCMNN image testing applications. The results of the test are output in a user friendly HTML format.

## 3.6 DWKNN regression implementation

This class has the basic functionality of making bare predictions using the distance weighted $k$-Nearest Neighbour algorithm. Using the algorithm it was possible to create a simple 2 dimensional example application that could demonstrate the effectiveness of the algorithm. A screen shot of this application is shown in Figure 3.12.

## 3.7 Neural network implementation

This class was developed primarily for part of my coursework for the Neural Networks course. A simple fully interconnected neural network was implemented that used the

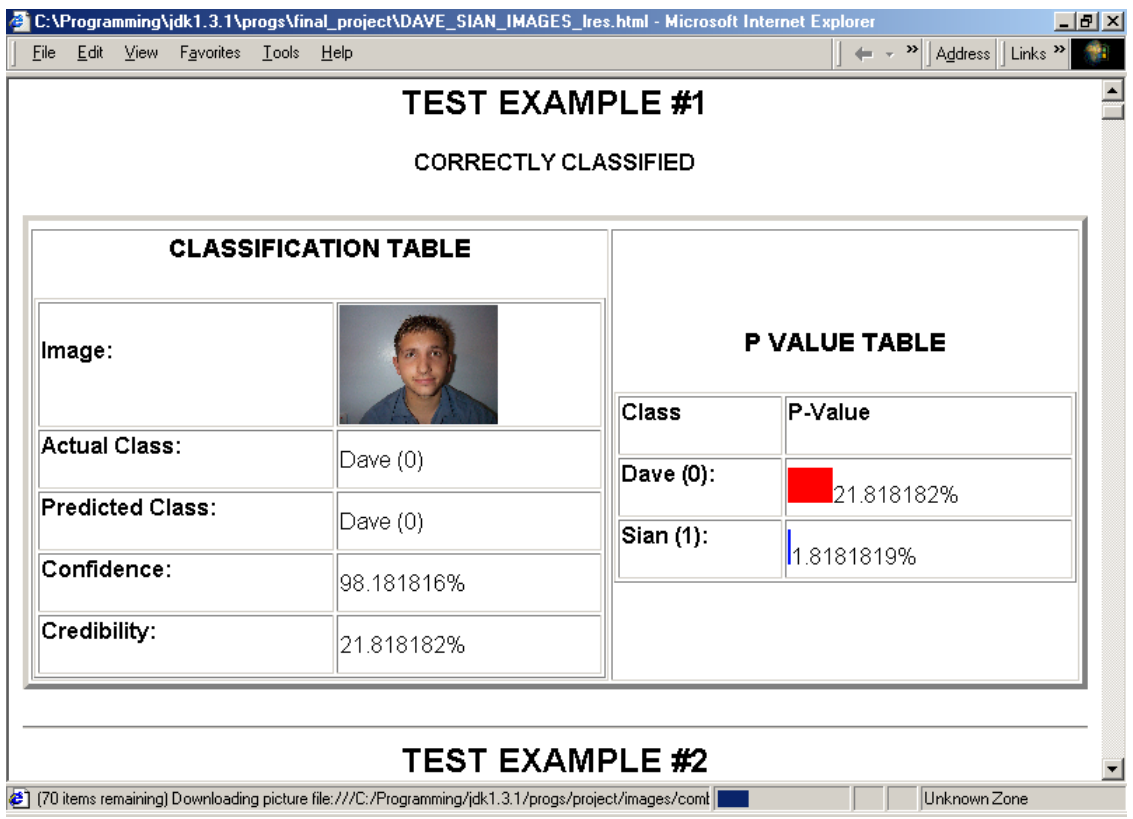

Figure 3.11: **Examples of the results HTML file generated by the TCMNN image testing applications**

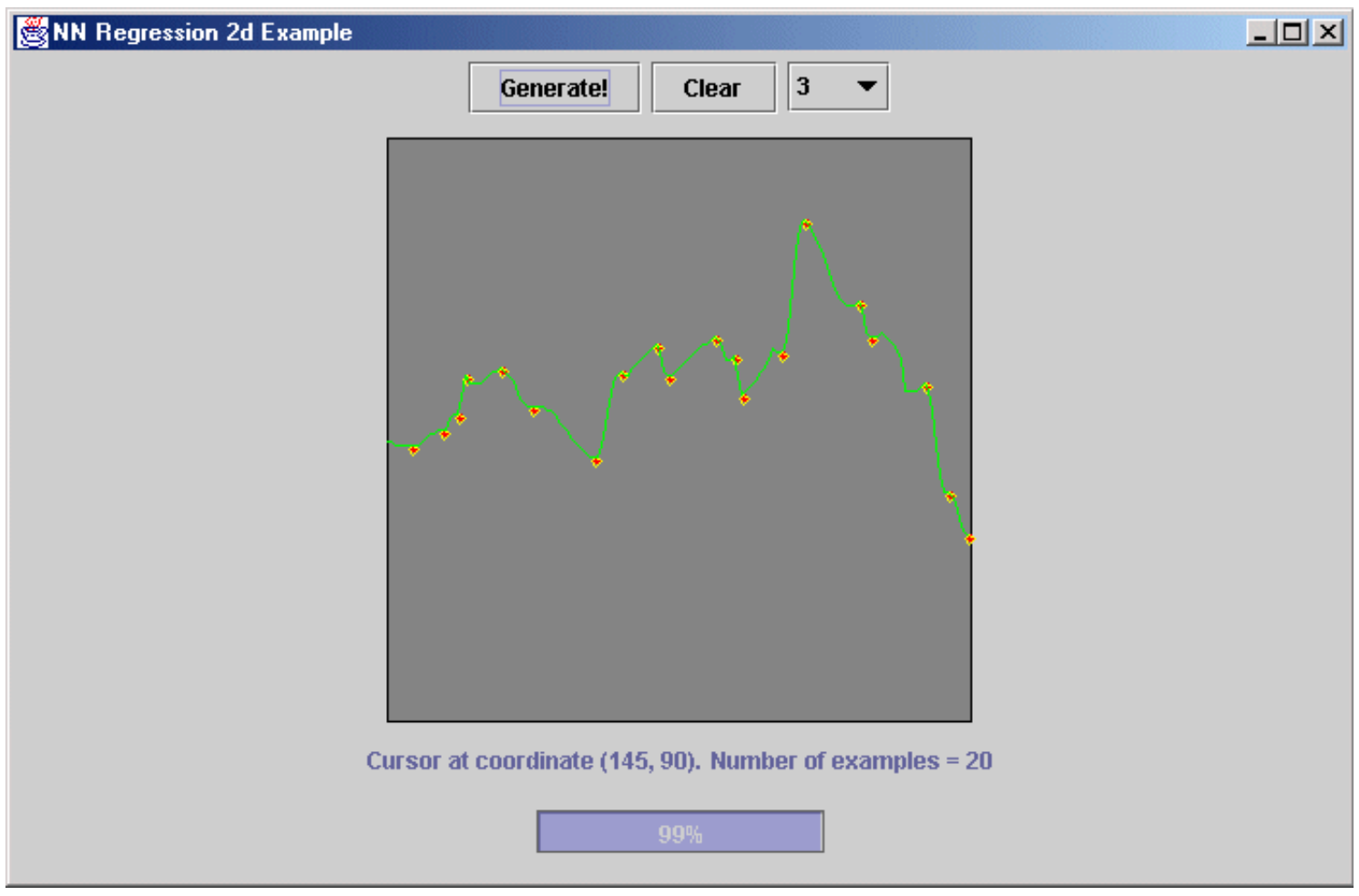

Figure 3.12: **Example of the 2d DWKNN regression application:** This application is used in a similar way to the TCMNN 2d application mentioned earlier. The user clicks within the training area to add training points (signified by red circles). Once the user has added the training data, they can then select the 'generate' button to calculate and display the best fit line. The algorithm can be tested to various numbers of $k$ nearest neighbours by selecting from the combo box. The progress of the calculations is displayed using a progress bar. For examples of simple tests carried out on artificial data sets see Figure B.8

stochastic version of the back-propagation algorithm (described in [4]) to train the weights. The basic functionalities of this class are outlined below:-

- Ability to create a fully interconnected network with any size hidden layers, with any number of hidden layers.
- Caches the weights of the network, so that they can be reloaded for later re-use.
- Ability to set various learning rates.
- Ability to set weight decay rates.
- Performs separate training and testing on data sets.
- Outputs of the neural networks sum of the squared errors over the total training set to a simple 2d Matlab plot. This plot can be used to assess convergence of the algorithm.

## 3.8 (DW)KNN pattern classification implementation

Both the normal $k$-Nearest Neighbour (KNN) and the distance weighted $k$-Nearest Neighbour (DWKNN) algorithms were implemented for use in the experiments. These classes had much the same functionality as the TCMNN algorithm, as listed below:-

- Perform separate, leave one out and random split tests
- Analyse to any number of $k$ nearest neighbours.
- Mark performance of the algorithm and output results to a simple text file.

# Chapter 4

# Experiments and Results

The experiments performed in this chapter are designed to test various topics covered earlier in the previous background theory chapter.

## 4.1 Introduction to data sets

The experiments will we be carried out on the following data sets. For each data set all the classifications of the examples were known.

### 4.1.1 Ovarian cancer

This data set was compiled by St. Bartholomews hospital in London. The data set has 3 attributes, which are menopausal status (0-3), ultrasound score (0-3) and CA125 level (continuous positive real value). Each example describes the particular measurements of the attributes taken from a patient, with the overall outcome of being either benign or malignant.

Initially we were presented with a training set of 139 examples (BARTS139 ) and a test set of 146 examples (BARTS146). These examples were combined to create a data set of 285 examples (BARTS285). Over the course of time a further 109 examples were presented, bringing the total to 394 (BARTS394). Unlike with the other data sets, the medical staff at St. Bartholomews hospital had already developed a basic formula of processing the attributes, to enable them to predict whether each case was benign or malignant.

They calculated a new measure called the risk of malignancy index (RMI), like so:

$$RMI = CA125 \times UltrasoundScore \times MenopausalStatus \tag{4.1}$$

Any patients with an RMI value greater than 200 are considered at risk of malignancy.

### 4.1.2 Abdominal pain

The second data set was provided by Bangour hospital. This data set is more complex because it has 135 attributes (values ranging from 0 to 1) describing various symptoms of abdominal pain cases with a respective diagnosis. Each example describes the particular measurements taken from a patient. There are 9 different diagnoses appendicitis, diverticulitis, perforated peptic ulcer, non-specific abdominal pain, cholisistitis, intestinal obstruction, pancreatitis, renal colic and dyspepsia. There are 4387 training (ABDO4387), and 2000 test examples (ABDO2000) for this problem.

### 4.1.3 Wisconsin Breast Cancer database

This data set was compiled by Dr. William and H. Wolberg, University of Wisconsin Hospitals, Madison, Wisconsin USA. It was donated by Olvi Mangasarian to the UCI machine learning resources database [23] in 1992. This particular data set has been tested using more traditional versions of the $k$-Nearest Neighbour algorithm in [15].

The data set has 9 attributes to describe the problem, plus an id number for patients (which is not used when processing). The attributes : clump thickness, uniformity of cell size, uniformity of cell shape, marginal adhesion, single epithelial cell size, bare nuclei, bland chromatin and normal nucleoli, have integer values ranging from 1 to 10. The possible classifications are benign or malignant. After removing examples with missing attribute values, 683 combined examples (WBC683) remained for training and testing. I then took a random split of the combined examples to create a fixed training set of 433 examples (WBC433) and 250 test examples (W-BC250). These separate data sets were useful for running faster experiments with slower algorithms such as TCMSVM.
Note: The results of test run on the Wisconsin breast cancer can be found in appendix A.
The next section will outline the types of test to be performed on the data sets described.

## 4.2 Description of tests using the TCMNN algorithm

The TCMNN algorithm has various choices of parameters that can be configured to obtain different results. Each test has been constructed to test the affect of altering these parameters.

### 4.2.1 The $k$ nearest setting

The $k$-Nearest Neighbour algorithm (which underpins the TCMNN algorithm) has the option to set the number of $k$ nearest neighbouring examples to analyse to. This test will see what happens when the value of $k$ is set to different levels.

### 4.2.2 Minkowski distance metric

As mentioned earlier, the Minkowski distance (equation 2.6) metric $L_k$ offers an alternative to the Euclidean distance normally used with the nearest neighbours algorithm. This test will see what affect using different values of $k$ has on the results.

### 4.2.3 Polynomial kernels

As mentioned earlier, kernels of the form $K(a,b)$ can be used to calculate the standard Euclidean distance between vectors in a higher dimensional feature space.

Polynomial kernels implicitly map to a higher dimensional space $((a \cdot b) + c)^d$ and can be used with various constants $c$, and powers $d$. Setting the $c$ value to zero will cause the kernel to map into a smaller feature space that will not include the original features.

This test will concentrate on the effect of polynomial kernels with various constant and degree settings.

### 4.2.4 Significance

The credibility value calculated for each example by the TCMNN can be used to place significance with the predictions. The predictions of examples with credibility below a certain threshold are discounted. This test will examine the effect marking the results of the TCMNN to different significance levels.

## 4.3 Description of tests using different machine learning algorithms

To compare the effectiveness of the algorithm, some of the most common machine learning algorithms will be used.

### 4.3.1 Neural Network

The tests will use a fully interconnected multi layered Neural Network, with standard sigmoidal activation function $\frac{1}{1+e^{-y}}$. The stochastic version of the back propagation algorithm outlined in [4] will be used to train the network.

As Neural Networks are an inductive learning technique, the results will be interesting to compare with that of the transductive algorithms (TCMNN, TCMSVM, DWKNN and KNN).

### 4.3.2 Traditional $k$-Nearest Neighbour

This test will use the 2 versions of the traditional $k$-Nearest Neighbour algorithm:

1. Standard $k$-Nearest Neighbour (KNN).
2. Distance weighted $k$-Nearest Neighbour (DWKNN).

These tests will be important to assess the difference between the performance of traditional $k$-Nearest Neighbour against its respective transductive confidence version TCMNN.

### 4.3.3 TCM Support Vector Machine (TCMSVM)

These tests will be carried out with a TCM using support vector machines as the underlying algorithm, developed by PhD students of the Computer Learning Research Centre (Royal Holloway University of London) . The system developed is available for use over the internet using any standard web browser [22]. Wherever possible these tests will be carried out using different kernel settings.

This test will be useful to compare the performance of the different implementations of the transductive confidence machine using different learning algorithms, namely the TCMNN.

## 4.4 How to assess performance

To assess the affect of altering the parameters, I will calculate the following statistics from the results of each test.

### 4.4.1 Overall accuracy

This is calculated by using the classification labels that are already provided for the test examples, thus:

$$\frac{Number\ of\ test\ examples\ correctly\ classified}{Total\ number\ of\ test\ examples} \tag{4.2}$$

### 4.4.2 Class accuracy

This is similar to the previous, except it gives us more detail as to how accurately the prediction is for each class:

$$\frac{Number\ of\ test\ examples\ with\ that\ particular\ class\ correctly\ classified}{Total\ number\ of\ test\ examples\ with\ that\ class} \tag{4.3}$$

### 4.4.3 Not classed

This is used in the significance tests performed with the TCMNN when assessing how many test examples are not considered when marking the results to a specified credibility threshold.

$$\frac{Number\ of\ test\ examples\ with\ credibility\ lower\ than\ threshold}{Total\ number\ of\ test\ examples} \tag{4.4}$$

### 4.4.4 Average confidence

This is a simple average of the confidence readings calculated for each example.

### 4.4.5 Average credibility

This is a simple average of the credibility readings calculated for each example.

### 4.4.6 Sensitivity and specificity

When designing a test for a disease often assess its performance by the following measures.

#### Sensitivity

This is how probable that the test is correct if you have the disease. With the cancer data sets this is equivalent to the malignant class accuracy.

#### Specificity

This is the probability of the test returns a false positive prediction. With the cancer data sets this can be seen as (100 - benign class accuracy). With the particular problem of cancer diagnosis we put more emphasis on sensitivity as early detection of the disease is crucial to effective treatment.

## 4.5 Ovarian cancer data set

The initial results of testing the TCMNN algorithm on the various data sets provided for the ovarian cancer problem can be seen in Table 4.1.

The best overall accuracy achieved when training and testing with these data sets was achieved with the leave one out test on the set of 285 examples. With this particular problem the accuracy of prediction for malignant cases is more important than that for benign cases.

As more training data is introduced, accuracy of prediction of malignant cases seems to decrease from 82.1% to 68.4%, while accuracy for benign increases from 78.5% to 84.1%. Average confidence drops from 95.2% to 92.3%, while average credibility also decreases from 62.4% to 51.6%.

| Results | Percentage accuracy % | | |
|---|---|---|---|
| | Separate (139,146) | Leave one out (285) | Leave one out (394) |
| Overall | 80.1 | 80.7 | 78.7 |
| Benign | 78.5 | 82.5 | 84.1 |
| Malignant | 82.1 | 77.8 | 68.4 |
| Average Conf | 95.2 | 92.8 | 92.3 |
| Average Cred | 62.4 | 57.3 | 51.6 |

Table 4.1: **Initial TCMNN test accuracy results using ovarian cancer data sets of different sizes:** These TCMNN tests were carried out using the standard settings, analysing to 1 nearest neighbour and using the standard Euclidean distance.

We would generally expect that presenting more training examples would improve the accuracy of prediction as more of the hypothesis space is explored and defined by the data set.

Decreases in average confidence values indicate that the TCMNN algorithm is finding it harder to distinguish between classifications. Decreases in average credibility indicate that each example is becoming stranger from the rest of the examples in the data set.

As seen later in Figure 4.1, the benign cases can be seen to be more similar to each other than the malignant cases. Because of the nature of this data, it is understandable that the classification of benign cases will be strengthened by the introduction of new training examples. This will consequently force more malignant cases to misclassified as benign.

As a more thorough check of the general performance of the TCMNN algorithm on the ovarian cancer data set, tests of random splits of 150 of the complete 394 examples were carried out. When taking random splits of the whole 394 examples, there were fluctuations in the accuracy of prediction, suggesting a certain level of noise in the training examples. The average of the results gave an overall accuracy of 77.1%, benign accuracy of 82.8% and malignant 65.7%. These results of the random split tests with the TCMNN algorithm can be seen in Table D.1 extended results of appendix C.

### 4.5.1 Ovarian cancer Test 1 TCMNN - $k$ nearest setting

The results of the testing the TCMNN algorithm with different $k$ nearest neighbours were contrary to what I had expected. I would have predicted that increasing the number of nearest neighbours would increase the algorithm's fault tolerance, and hence increase the accuracy of prediction.

| K value | Percentage accuracy % | | | | |
|---|---|---|---|---|---|
| | Overall | Benign | Malignant | Average Conf | Average Cred |
| 1 | 80.1 | 78.5 | 82.1 | 95.2 | 62.4 |
| 10 | 80.1 | 84.8 | 74.6 | 94.1 | 47.4 |
| 20 | 80.8 | 92.4 | 67.2 | 92.8 | 44.9 |
| 30 | 80.8 | 97.5 | 61.2 | 91.9 | 46.2 |
| 40 | 71.9 | 100.0 | 38.8 | 88.5 | 48.5 |
| 50 | 72.6 | 100.0 | 40.3 | 89.7 | 46.6 |

Table 4.2: **Results of TCMNN testing the ovarian cancer data set with different numbers $k$ of nearest neighbours:** These tests were carried out with the separate training (BARTS139)and testing (BARTS146) files described earlier. The TCMNN algorithm was performed using the standard Euclidean distance metric.

The results in Table 4.2 show that as the number of nearest neighbours increases, the overall accuracy remains constant up until 40 nearest neighbours whereupon a decrease is observed from 80.8% to 72.6%. The accuracy of prediction for benign cases dramatically increases from 78.5% to 100.0%, whilst malignant cases decrease from 82.1% to 40.3%. The average confidence for test examples is seen to decrease from 95.2% to 89.7%. Average credibility also decreases from 62.4% to 46.6%.

To understand these unexpected results, one can visualise the problem in a 3 dimensional scatter plot as shown in Figure 4.1. This is due to the fact that this classification problem only has 3 attributes describing it.

The plot in Figure 4.1 makes it clear to see that there is more variance in the attributes of malignant cases than the benign cases. The benign cases are tightly clustered into 6 distinct regions, whereas the malignant cases are spread over a wider range.

This representation of the training data shows why the accuracy of benign cases increases and malignant decreases with increased number of nearest neighbours. Looking at the clusters of benign cases, one observes that although the malignant cases are spread over a wider range, there are clusters of malignant cases very close to the benign clusters.

When analysing to 1 nearest neighbour, the examples in this 'confused region' of Euclidean space will be roughly misclassified equally. However, when increasing the nearest neighbours, the benign test cases in this confused region will be forced to be classified correctly as the clustered benign training cases will make the example seem less strange compared to the varied malignant cases.

In contrast, a malignant test case in this confused region will become more strange to its true malignant class. This is because the total $k$ nearest distances from the benign training cases will be less than that of the varied malignant training cases, causing it to be misclassified as benign.

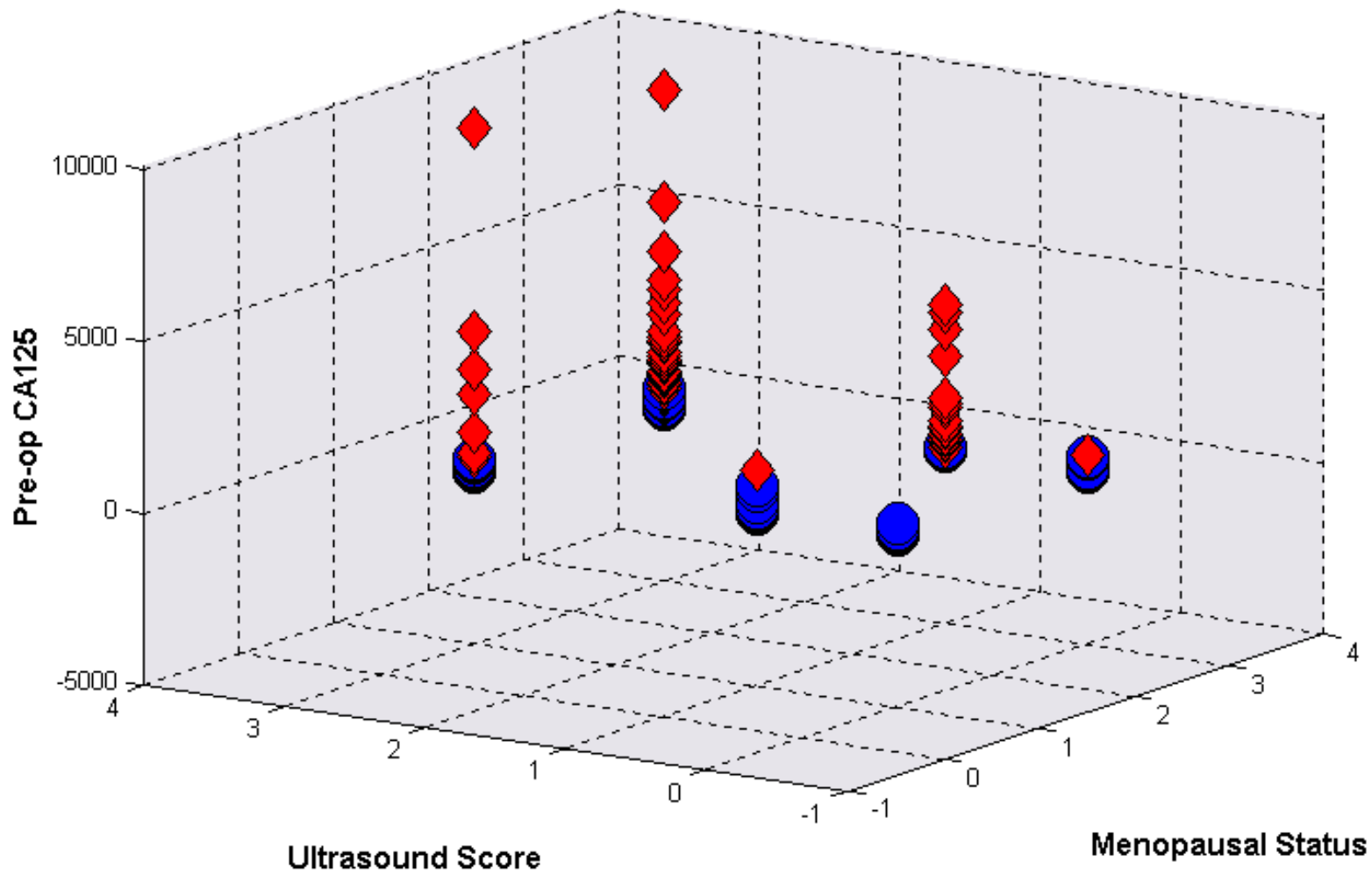

Figure 4.1: **Mapping of the ovarian cancer data set into 3-dimensional Euclidean space:** Red diamonds are malignant cases, blue circles are benign cases.

To ensure that the observed phenomena was not just due to the particular data set used in this test, the same test was repeated using a further data set this time carrying out a leave one out test on a BARTS285 data set. The results of this test shown in Table D.2 reflect the same nature of benign accuracy increasing, whilst malignant accuracy decreases with increased nearest neighbours. Clearly this is an undesirable feature, as it is not good to cause a malignant case to be misclassified as benign. The key to effective treatment of any cancer is early detection of the disease.

### 4.5.2 Ovarian cancer Test 2 TCMNN - Minkowski Metric

The use of different values of $k$ in the Minkowski distance metric with the TCMNN algorithm gave surprising results. The definition of the Minkowski metric taken from [2] did not specifically state any restrictions on the values of $k$. When running the tests I decided to test using values of $k$ such that $0 < k < 1$, along with values of $k > 1$ which are more commonly used.

The best overall accuracy of 85% was achieved using $k = \frac{1}{4}$, which was a 5% increase from that obtained from standard Euclidean distance. The results in Table 4.3 show a trend of increasing accuracy for both malignant (82.1% to 85.1%) and benign cases (78.5% to 88.6%) as the value of $k$ is decreased. In contrast, the accuracy of prediction appeared to decrease as the value of $k$ increased.

Average confidence was seen to stay roughly constant throughout all settings of $k$. From the values of $k$ ranging from 2 to $\frac{1}{5}$ the average credibility was seen to decrease from 62.4% to 58.1%. In contrast, from $k$ values 2 to 5 a slight increase in average credibility from 62.4% to 63.7% was observed.

Constant average confidence values imply that the Minkowski metric has little effect on the TCMNN algorithm's ability to distinguish between classifications. However, the results with average credibility imply that increasing the $k$ value in the Minkowski metric to values greater than 2 cause examples to become less strange from one another. In contrast, decreasing the value $k$ to less than 2 appears to cause the examples to become stranger from one another.

Notice that the values of the Minkowski metric $L_k$ set to $k = 2$ give exactly the same performance as that achieved using the standard Euclidean distance as shown earlier in Table 4.1. This is expected as the $L_2$ metric is mathematically equivalent to that of the Euclidean distance.

The results in Table 4.3 show that different values of $k$ in the Minkowski metric yield better performance than that obtained from testing with the standard Euclidean distance ($k = 2$) commonly used with the TCMNN algorithm.

| K value | Percentage accuracy % | | | | |
|---|---|---|---|---|---|
| | Overall | Benign | Malignant | Average Conf | Average Cred |
| $\frac{1}{5}$ | 84.9 | 88.6 | 80.6 | 95.4 | 58.1 |
| $\frac{1}{4}$ | 85.6 | 87.3 | 83.6 | 95.5 | 58.3 |
| $\frac{1}{3}$ | 84.2 | 83.5 | 85.1 | 95.6 | 59.5 |
| $\frac{1}{2}$ | 82.1 | 79.7 | 85.1 | 95.8 | 60.7 |
| 1 | 81.5 | 79.7 | 83.6 | 95.4 | 61.3 |
| 2 | 80.1 | 78.5 | 82.1 | 95.2 | 62.4 |
| 3 | 79.5 | 77.2 | 82.1 | 95.1 | 63.2 |
| 4 | 79.5 | 77.2 | 82.1 | 95.2 | 63.3 |
| 5 | 78.8 | 77.2 | 80.6 | 95.1 | 63.7 |

Table 4.3: **Results of TCMNN testing the ovarian cancer data set using the Minkowski distance metric with different values of $k$:** These tests were carried out with the separate training (BARTS139), testing (BARTS146) files described earlier. The TCMNN algorithm was performed analysing to 1 nearest neighbour.

On further inspection, it can be realised with a simple example that using the Minkowski distance metric with fractional powers of $\frac{1}{k}$ instead of $k$ means that the metric no longer is a valid distance metric.

Referring back to the discussion earlier, where the conditions of a distance metric are specified on page 26. The triangular inequality property states that, for all vectors **a**, **b**, **c**, the following must hold:

$$d(\mathbf{a}, \mathbf{b}) + d(\mathbf{b}, \mathbf{c}) \geq d(\mathbf{a}, \mathbf{c})$$

Consider the vectors $\mathbf{a} = \begin{pmatrix} 1 \\ 0 \end{pmatrix}$, $\mathbf{b} = \begin{pmatrix} 0 \\ 0 \end{pmatrix}$ and $\mathbf{c} = \begin{pmatrix} 0 \\ 1 \end{pmatrix}$ using the measure $L_{\frac{1}{2}}$ defined as:

$$L_{\frac{1}{2}}(\mathbf{a}, \mathbf{b}) = \Big( \sum_{i=1}^{n} |a_i - b_i|^{\frac{1}{2}} \Big)^2$$

Using vectors $\mathbf{a}, \mathbf{b}, \mathbf{c}$ with the measure $L_{\frac{1}{2}}$ we get:

$$d(\mathbf{a}, \mathbf{b}) + d(\mathbf{b}, \mathbf{c}) = 2 \ngeq d(\mathbf{a}, \mathbf{c}) = 4$$

This example clearly highlights the fact that the Minkowski metrics of the form $L_{\frac{1}{k}}$ that were used in the experiments were not valid according to the formal definitions of Euclidean distance metrics mentioned earlier. For this reason one must think of the function as a 'measure' computed between vectors. This of course does not change the fact that using this measure improves the performance of the TCMNN algorithm.

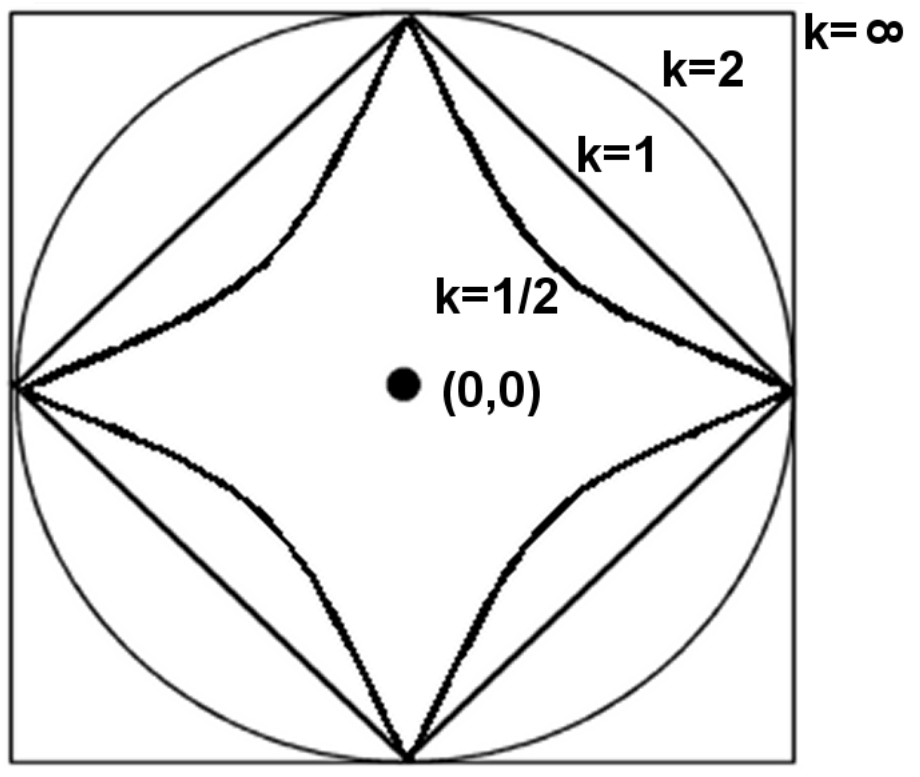

Figure 4.2: **Diagram of Minkowski metric representation in 2 dimensions:** The diagram is a plot of all vectors that lie a unit distance away from the origin, using different powers $k$ of the Minkowski metric $L_k$.

By looking at a simple plot of the Minkowski metrics as in Figure 4.2, it is clear to see what the geometrical representations are that these metrics represent. Notice the Euclidean distance ($k = 2$) considers all vectors in a unit circle around the origin as we would expect. This behaviour of the distance measure can be seen in Figure **??** on simple 2 dimensional data problems.

By using $L_k$ with values $k \leq 2$ the measure is actually putting more emphasis on vectors that lie perpendicular to an arbitrary vector in Euclidean space. In contrast, using values of $k \leq 2$, the metric can be seen to put less emphasis on vectors lying perpendicular to a target vector.

The nature of this measure and the 3 dimensional plot of the data in Figure 4.1 can be used to explain why there is an improvement in the performance of the TCMNN algorithm on the ovarian cancer data set. The menopausal status and ultrasound score attributes are discrete integer values ranging between 1 and 3. This causes vectors to lie perpendicular with one another. Looking at the plot of the data we can see that benign and malignant cases lie perpendicular to one another, and that by using this measure the algorithm will be helped to associate the correct vectors with new test cases.

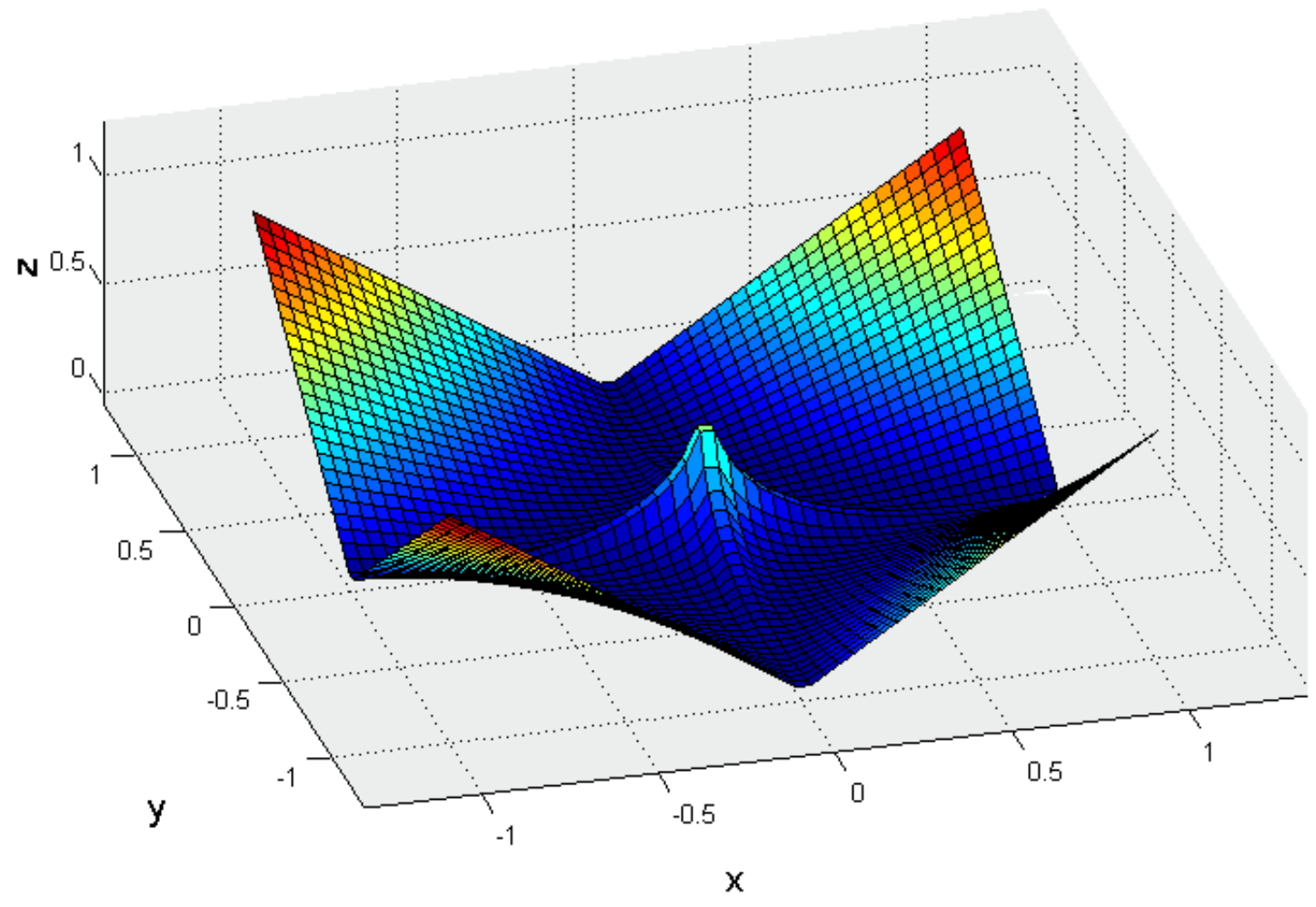

Figure 4.3: **Diagram of Minkowski metric $L_{\frac{1}{2}}$ representation in 3 dimensions:** The diagram is a plot of all vectors that lie a unit distance away from the origin, using the measure $L_{\frac{1}{2}}$.

Looking at a particular example of one of these invalid metrics $L_{\frac{1}{2}}$ in 3 dimensions, one can see the true extent to the bizarre nature of this measure. It is easy to think that the standard Euclidean distance considers vectors lying on the surface of a unit sphere to be equally as close. Figure 4.3 shows that the measure considers vectors that lie on the hyperplanes seen jutting out of the plot to be infinitely closer than would have been originally considered using a standard distance metric.

### 4.5.3 Ovarian cancer Test 3 TCMNN - Kernel distance metric

The results of tests using various settings of polynomial kernels on this data set were very disappointing. Unlike the test with the Minkowski metric, no marked improvements on the results of the standard Euclidean performance were observed.

| Results | Percentage accuracy % | | | | |
|---|---|---|---|---|---|
| | Euclid | Poly deg 2 | Poly deg 3 | Poly deg 4 | Poly deg 5 |
| Overall | 80.1 | 78.8 | 78.1 | 78.8 | 78.1 |
| Benign | 78.5 | 75.9 | 75.9 | 77.2 | 77.2 |
| Malignant | 82.1 | 82.1 | 80.6 | 80.6 | 79.1 |
| Average Conf | 95.2 | 95.0 | 94.5 | 94.4 | 94.3 |
| Average Cred | 63.3 | 63.4 | 64.0 | 63.8 | 63.1 |

Table 4.4: **Results of TCMNN testing the ovarian cancer data set using a polynomial kernel with various degrees $d$:** These tests were carried out with the separate training (BARTS139) and testing (BARTS146) files described earlier. The TCMNN algorithm was performed analysing to 1 nearest neighbour, using the standard Euclidean distance. All polynomial kernel constants in these tests were set $c = 0$.

The results in Table 4.4 show that changing the degree used in the polynomial kernel had little effect on the performance of the TCMNN algorithm. One strange observation was that odd powers gave a consistent overall performance of 78.8%, and even powers similarly gave 78.1%. However, when looking closer at the individual class performance in these tests it can be seen that they differ between odd and even powers results.

To test the effect of altering the constant in the polynomial kernel, tests were run fixing the polynomial to degree 2 and setting the constant to various values. The results of the tests can be seen in Table 4.5.

The results of further use of polynomial kernels in Table 4.5 were also disappointing. All results were exactly the same as that seen when the constant was set to zero earlier in Table 4.4.

By considering the binomial expansion of the polynomial kernel mentioned earlier on 29, we can build a better intuition of what the polynomial kernel feature mapping actually represents. Given $n$ dimensional vectors $\mathbf{x}$ and $\mathbf{z}$ the polynomial kernel maps to the feature space defined by the terms:

| Results | Percentage accuracy % | | | | | |
|---|---|---|---|---|---|---|
| | Euclid | $c$=0 | $c$=1/2 | $c$=10 | $c$=100 | $c$=1000 |
| Overall | 80.1 | 78.8 | 78.8 | 78.8 | 78.8 | 78.8 |
| Benign | 78.5 | 75.9 | 75.9 | 75.9 | 75.9 | 75.9 |
| Malignant | 82.1 | 82.1 | 82.1 | 82.1 | 82.1 | 82.1 |
| Average Conf | 95.2 | 95.0 | 95.0 | 95.0 | 94.9 | 94.9 |
| Average Cred | 63.3 | 63.4 | 63.4 | 63.4 | 62.8 | 62.2 |

Table 4.5: **Results of TCMNN testing the ovarian cancer data set using a polynomial kernel with various constants $c$:** These tests were carried out with the separate training (BARTS139) and testing (BARTS146) files described earlier. The TCMNN algorithm was performed analysing to 1 nearest neighbour, using the standard Euclidean distance. All polynomial kernel degrees were set $d = 2$.

$$(\langle \mathbf{x}\cdot\mathbf{z}\rangle + c)^d = (\langle \mathbf{x}\cdot\mathbf{z}\rangle)^d + \binom{d}{1}(\langle \mathbf{x}\cdot\mathbf{z}\rangle)^{d-1}c + \binom{d}{2}(\langle \mathbf{x}\cdot\mathbf{z}\rangle)^{d-2}c^2 + \ldots + \binom{d}{d-1}(\langle \mathbf{x}\cdot\mathbf{z}\rangle)c^{d-1} + c^d$$

Where $(\langle \mathbf{x} \cdot \mathbf{z}\rangle)^d$ can be thought of as the all possible $d$ tuples of the $n$ attributes describing each example. In the application to the medical data sets, it can be thought of as looking at the problem in terms of $d$ combinations of symptoms. This can be useful if considering the interaction between attributes, and hence will aid the learning process.

The constant $c$ can be thought of as how much weighting to put on smaller sets of $d$ tuples, including the original feature vectors. Setting the value of $c = 0$ causes the problem to be considered in terms of just the $d$ tuples as the new features of the problem.

Because the ovarian cancer data set problem has a very small number of attributes (where much of the values are similar between examples), the effect of cross validating the attributes using a polynomial kernel would be negligible.

### 4.5.4 Ovarian cancer Test 4 TCMNN - Significance level

The results of testing the TCMNN algorithm's performance with different significance levels were disappointing. Table 4.6 shows results of testing the performance of the TCMNN algorithm with significance levels ranging from 5% to 90%.

The best overall performance achieved was 89.9% with a significance level of 50%, which is a 9.8% increase on the overall performance achieved without marking to any significance. The setting of 50% significance improved accuracy of benign cases from 78.5% to 90.0% and malignant cases from 82.1% to 89.8%. However at this setting 32.2% of examples are not classified.

| Significance level | Percentage accuracy % | | | |
|---|---|---|---|---|
| | Overall | Benign | Malignant | Not Classified |
| 5% | 80.1 | 78.5 | 82.1 | 0.0 |
| 10% | 80.1 | 78.5 | 82.1 | 0.0 |
| 20% | 80.7 | 78.9 | 82.8 | 4.1 |
| 30% | 84.1 | 79.4 | 89.7 | 13.7 |
| 40% | 88.4 | 86.4 | 90.6 | 23.3 |
| 50% | 89.9 | 90.0 | 89.8 | 32.2 |
| 60% | 89.7 | 90.2 | 89.4 | 39.7 |
| 70% | 87.7 | 87.1 | 88.1 | 50.0 |
| 80% | 88.4 | 87.5 | 88.9 | 70.5 |
| 90% | 80.0 | 50.0 | 100.0 | 93.2 |

Table 4.6: **Results of TCMNN testing the ovarian cancer data set using different significance levels:** These tests were carried out with the separate training (BARTS139) and testing (BARTS146) files described earlier. The TCMN-N algorithm was performed analysing to 1 nearest neighbour, using the standard Euclidean distance.

When comparing these results to the effectiveness of the screening system that the hospital already use as shown in Table D.3, one notices that marking with significance does indeed improve on both the specificity and sensitivity of the results. The overall performance of the hospital's RMI screening process on the same test set used above gives an overall performance of 86.9%, with accuracy for benign cases of 92.4% and malignant 80.6%.

There is a degree of subjectivity as to whether one considers this improvement on the existing screening system valid or not. On one hand there is greater specificity and sensitivity when marking the TCMNN results with significance than that of the RMI index. Conversely, marking with significance would mean that 32.2% of patients would not be given any (valid) prediction of diagnosis at all. Perhaps new test patients who fall into this category of being not classed could undergo further rigorous examination to ascertain their true diagnosis.

### 4.5.5 Ovarian cancer Test 5 Performance of other learning algorithms

Table 4.7 shows results of comparing the effectiveness of different machine learning algorithms on the separate BARTS139/BARTS146 test data sets. This table also in-cludes details of the performance of the RMI screening technique currently employed by the hospital.

The best overall performance of 88.4% was achieved using neural networks. On closer inspection we see that this is due to high accuracy of benign cases of 98.7% compared to malignant cases 73.1%. The RMI method performs with an overall accuracy of 86.9%, which is slightly less than that achieved by the neural network. In contrast, the RMI method gives greater accuracy for malignant cases than the neural network does.

| Algorithm Used | Overall % | Benign % | Malignant % |
|---|---|---|---|
| Neural Network | 88.4 | 98.7 | 73.1 |
| DWKNN | 78.8 | 78.5 | 79.1 |
| KNN | 78.8 | 75.9 | 82.1 |
| TCMNN | 80.1 | 78.5 | 82.1 |
| TCMSVM | 84.2 | 84.8 | 83.6 |
| St. Barts RMI | 86.9 | 92.4 | 80.6 |

Table 4.7: **Results of testing the ovarian cancer data set using different learning algorithms:** Each of these learning algorithms were trained and tested using the same BARTS139 and BARTS146 data sets to ensure a fair comparison. All versions of the $k$-Nearest Neighbour algorithm (TCMNN, KNN and DWKNN) were set to standard settings of 1 nearest neighbour, using Euclidean distance. The TCMSVM results were obtained using the standard inner product.

I did not expect the neural network to outperform the other transductive learning algorithms tested. The neural network was the only inductive algorithm that was used in the tests. From the discussion earlier about inductive and transductive approaches to machine learning, one would expect that the transductive algorithms would be more powerful and outperform inductive learning algorithms.

The TCMNN algorithm outperformed the more traditional versions of the $k$-Nearest Neighbour algorithm by roughly 2%. It is important to remember that the functionality of the TCMNN algorithm was not primarily designed to improve performance of the algorithms predictions but to address the problem of density estimation.

The results of the TCMSVM algorithm performed 4.1% better than of the TCMNN algorithm. The results given in Table 4.7 for TCMSVM were calculated using a standard inner product. Using the inner product is equivalent to running the SVM in the original input space. As SVM's are kernel based learning algorithms one would be hopeful that the results could be improved even further by the choice of an appropriate kernel.

One practical difference between the running of these two different implemen-tations of the TCM is the computational complexity. The test took roughly 10 seconds to complete running on the TCMNN, whilst the TCMSVM took over 2 hours to complete.

### 4.5.6 Ovarian cancer - Further experiments with neural networks

After observing the interesting performance results for neural networks mentioned earlier in Table 4.7, I decided to run further tests with the algorithm to test if this performance could be improved, or if it could be harnessed to improve the performance of the TCMNN algorithm.

The results given in Table 4.7 were calculated using a simple 3-5-2 feed forward network as shown in Figure 4.4. The learning rate $\eta$ was set to 0.1 and the weights were initially set randomly between $\pm 0.05$. To increase training time, the attributes were normalised between 0 and 1. The results were achieved after training for 100,000 weight updates.

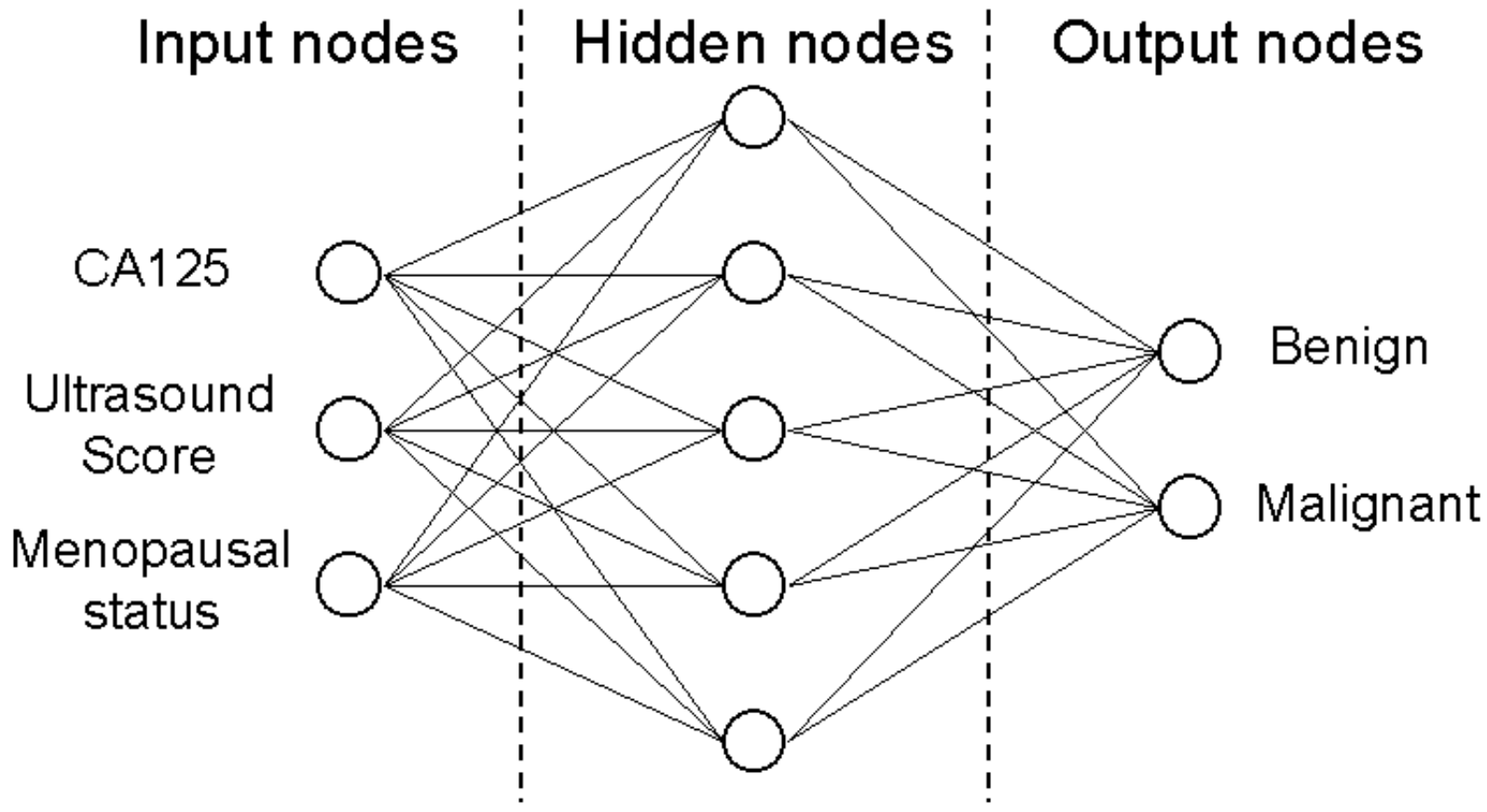

Figure 4.4: **Neural network design for testing the ovarian cancer data set:** The network was designed as a simple 3-5-2 fully interconnected feed forward network. The outputs correspond to each of the classifications. The prediction from the network is taken as the highest value of all the output nodes.

I chose a 1 of N output encoding so as to enable a much greater hypothesis space than would be possible by encoding the output as a single thresholded unit (eg. output less than 0.5 predict benign, greater than 0.5 predict malignant).

Each possible classification was given a respective output unit. The index of the highest output from the network was taken as the prediction. The target values were initially set to 0.999 for the correct classifications, and 0.001 for the others.

As neural networks are parametric algorithms, there could be a more optimal solution possible by configuring any of the many parameters such as learning rate, the size of weights and the dimensions of the network. I will not be focussing in great detail of these specific parameters as the primary concern in this project is the TCMNN algorithm.

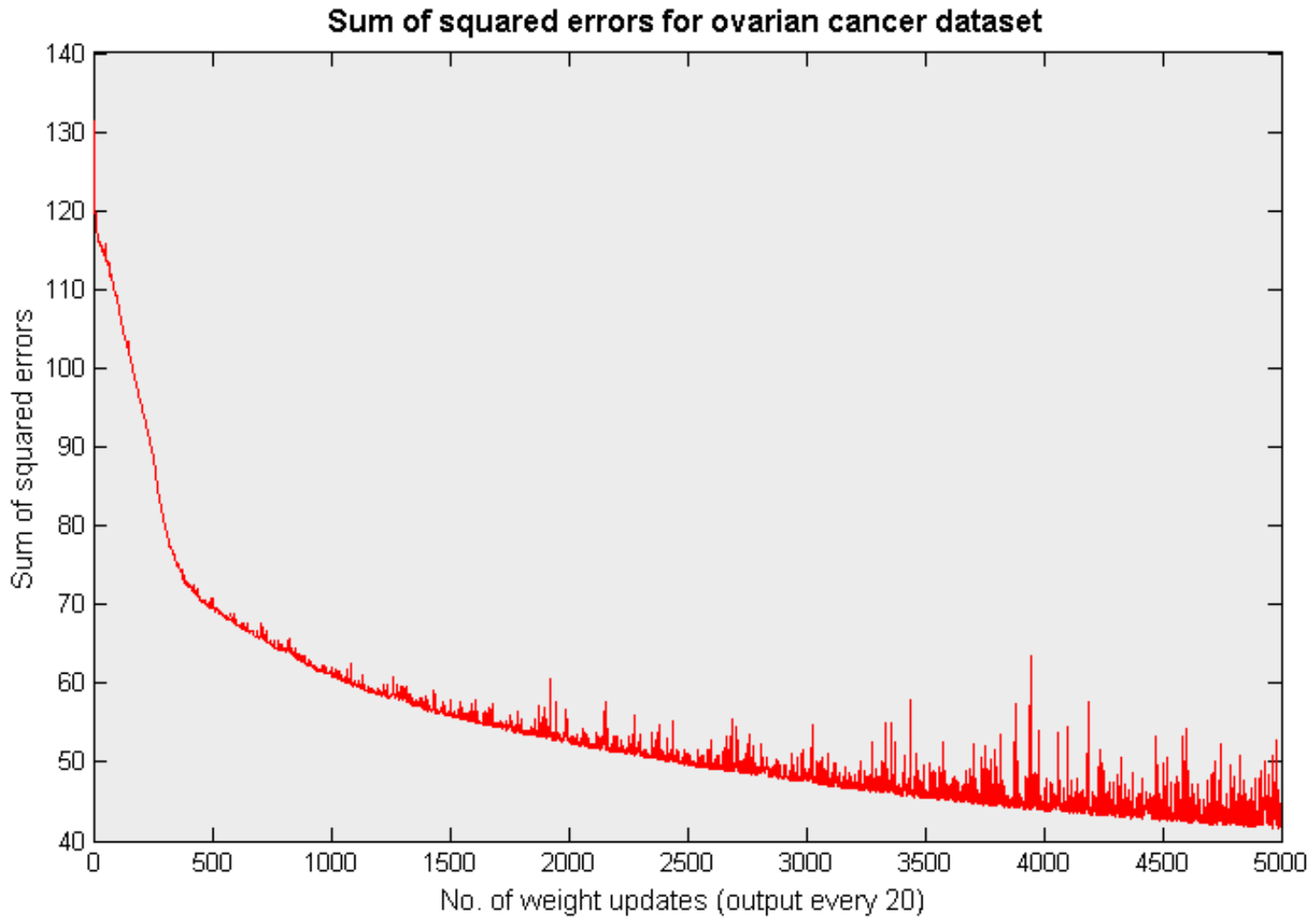

Figure 4.5: **Plot of neural networks sum of squared errors for ovarian cancer data set:** The above graph demonstrates the convergence of the back propagation algorithm on the ovarian cancer data set to the solution given. There were in total 100,000 weight updates. The sum of the squared errors over the total training set (BARTS149) was calculated every 20 weight updates and plotted respectively.

One special property of neural networks is their flexibility. As mentioned earlier, tests were run with 0.999 as the target value of the true classification in the output nodes of the network. If this value is altered for one of the classes, it can be seen as trying to bias the learning process and also trying to force the learning algorithm to concentrate on a particular classification.

As mentioned earlier, the performance of the neural network on the BARTS146 test set gave class accuracies of 98.7% for benign cases and 73.1% for malignant cases. It was clearly disappointing that the accuracy for malignant cases was much lower than that of benign. These tests were run using no bias on either classification. It is therefore interesting to question the possibility of improving this accuracy by placing more bias on malignant cases when training the network.

The results of Table 4.8 demonstrate the effect of setting these target values of the neural network for benign cases to various levels. The results show that there is improvement in the malignant cases, at the cost of benign accuracy. The best accuracy for malignant cases of 83.6% was achieved with a setting of benign cases with target values of 0.5.

| Target values | Average Percentage Accuracy % | | |
|---|---|---|---|
| | Overall | Benign | Malignant |
| B = 0.999, M = 0.999 | 88.4 | 98.7 | 76.1 |
| B = 0.8, M = 0.999 | 89.7 | 98.7 | 79.1 |
| B = 0.7, M = 0.999 | 88.4 | 93.7 | 82.1 |
| B = 0.6, M = 0.999 | 88.4 | 92.4 | 83.6 |
| B = 0.5, M = 0.999 | 88.4 | 92.4 | 83.6 |

Table 4.8: **Effect of various target values using Neural Networks on ovarian cancer data set:** This table displays the results of setting the target values for each class benign (B) and malignant (M) to different levels. Each of the values in this table are the best taken from repeated separate tests using the BARTS139 and BARTS146 data sets. The results of these tests can be seen in Tables D.4 , D.5, D.6, D.7 and D.8.

The results of Table 4.8 reflect the flexibility of neural networks in their ability to configure many different parameters and enhance the results. An intriguing point is raised as to why the prediction accuracy is higher than that of the transductive algorithms (TCMNN, DWKNN, KNN and TCMSVM), which are supposed to be capable of creating equally, if not more complicated decision rules due to their ability to estimate the decision rules locally.

Another special property of training multi-layered networks is that the values of the nodes in the hidden layer (as see Figure 4.4) can be seen to have the ability to discover useful intermediate representations [4].

If the network is capable of making better predictions using the intermediate hidden representations, then these hidden values could be used to make better predictions with the TCMNN classifier. These hidden values could be interpreted as a way of establishing the most important details about the data to be learnt. This is similar to the problem posed earlier of finding a relevant way of scaling the original attributes according to their real influence to the classification.

This could be achieved by creating an augmented version of the original training set as outlined in Figure 4.6.

Table D.9 in appendix D shows results of repeated tests on the ovarian cancer data set using the augmented training set technique outlined in Figure 4.6. The tests were carried out using repeated random splits of 150 examples of the BARTS285 data set.

The TCMNN performance was tested before and after as a fair comparison of us-ing the augmentation process. The neural network's performance was also tested to see if it was a factor involved with the improvement of accuracy of the TCMNN algorithm. Table D.9 indicated that the greatest improvement in the overall performance of 8% difference was achieved in test 2. This test also correlated with the highest neural network overall accuracy achieved of 94.0%. It would seem therefore that this augmentation process is somewhat reliant on the neural networks performance on the original data set.

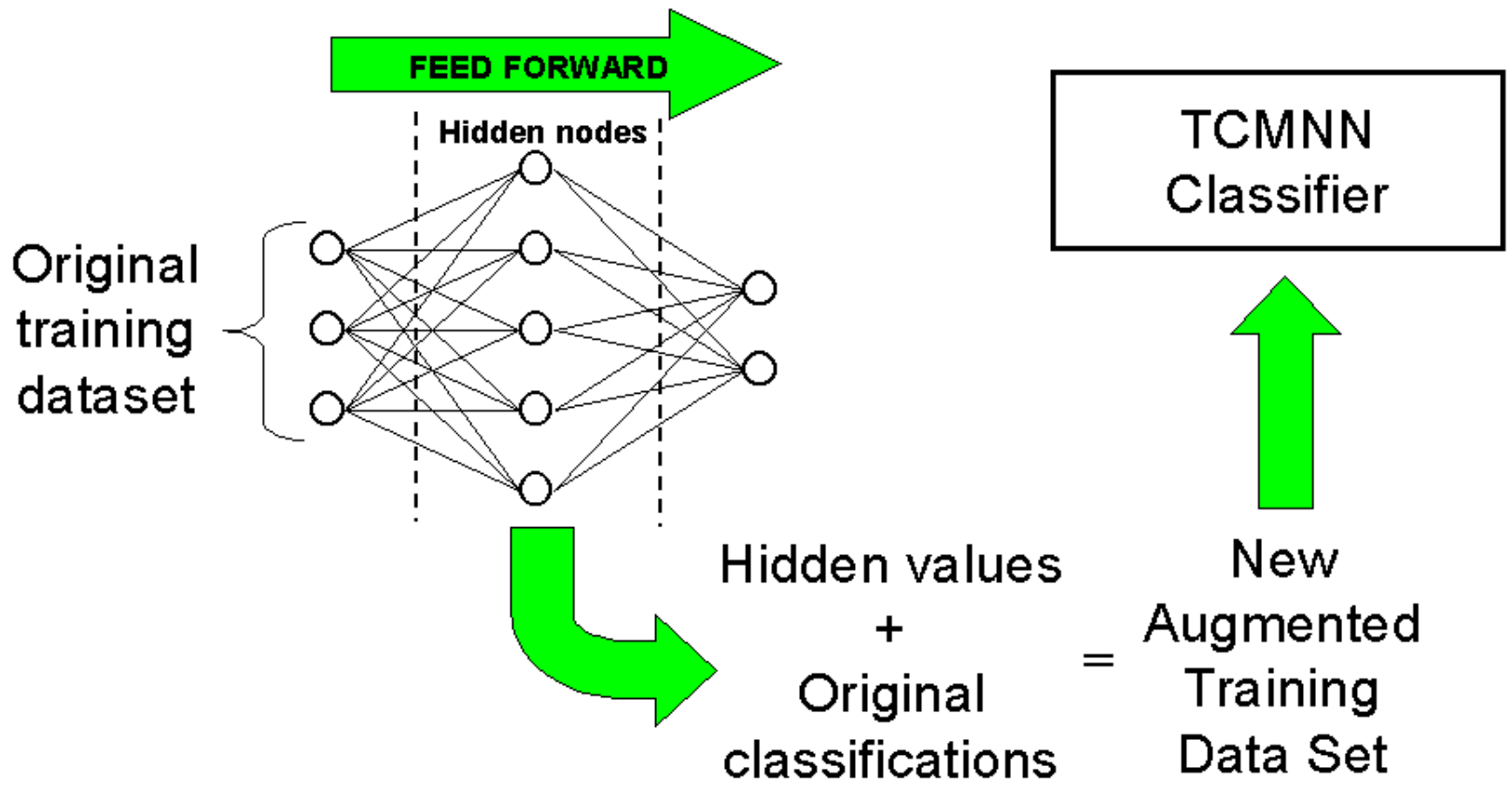

Figure 4.6: **Creating augmented data sets:** The above diagram outlines how to use a neural network to create augmented versions of the data sets that can be used by a different machine learning classifier. Initially, the network is trained using the original training set. Then the original training vectors are input to the network and fed forward through the network. The values that are held in the hidden nodes of the network are then extracted, along with the original training vectors real classification. This process is repeated for all the vectors in the training set to create the augmented version of the data set. In effect, the network is used to create a transformation into an alternative feature space, which is hopefully easier to learn from. Once this new data set is created it can then be used to train different learning algorithms such as the TCMNN classifier.

Table 4.9 gives the averages of the values taken in the repeated tests shown in Table D.9 of appendix D. The results show an average increase of 3.8% in overall accuracy. Benign accuracy is seen to increase by 3.6% and malignant cases similarly by 3.5%. Slight increases were observed in both average confidence and credibility, which indicated that the augmentation process was assisting in making the examples less strange from one another hence making it easier for the algorithm to classify.

As a final check to see if these differences were significant and not due to chance, I performed a two tailed $t$-test to the results in Table D.9. The equation for the test

| Training Data | Percentage Accuracy % | | | | |
|---|---|---|---|---|---|
| | Overall | Benign | Malignant | Average Conf | Average Cred |
| Original | 79.2 | 85.5 | 69.5 | 92.2 | 59.1 |
| Augmented | 83.0 | 89.1 | 73.0 | 92.9 | 60.4 |

Table 4.9: **Summary of augmented data set tests with TCMNN on the ovarian cancer data set:** The results are averages of the values observed in the tests run with augmented data sets shown in Table D.9. The TCMNN was set to standard setting of 1 nearest neighbour, using Euclidean distance.

is as follows:

$$t = \frac{\sum D}{\sqrt{\frac{n \sum D^2 - (\sum D)^2}{(n-1)}}} \tag{4.5}$$

where $D$ is the difference between groups, and $n$ is the number of pairs of observations.

Calculating the $t$-test statistic comparing the overall performance of the TCMNN before and after we get:

$$t = \frac{38}{\sqrt{\frac{10(244) - (38)^2}{(9)}}} = 3.612$$

Because we are testing dependent samples, the degrees of freedom are taken to be $n - 1 = 9$. Analysing to 9 degrees of freedom and taking a risk of 0.01 gives a critical value of 3.25 from statistical tables. Our observed value $t$=3.612 $>$ 3.25, therefore we can be sure that the differences observed are 99% significant.

### 4.5.7 Extended tests with TCM Support vector machine (TCMSVM)

To further test the comparison of the TCMSVM and TCMNN algorithms performance on the ovarian cancer data set, I ran a leave one out test on the 285 example data set using the TCMSVM. The computation time for running the same respective leave one out test with TCMSVM took just over a day to compute, compared to my implementation of TCMNN which took roughly 90 seconds.

As indicated earlier in the 3 dimensional plot of the ovarian cancer data set in Figure 4.1, there was some overlap between the benign and the malignant examples, which could have caused the SVM to have difficulty in finding a separating hyperplane between the two classes.

| Results | Percentage accuracy % | | |
|---|---|---|---|
| | Separate (139,146) | Leave one out (285) | Leave one out (394) |
| TCMNN | | | |
| Overall | 80.1 | 80.7 | 78.7 |
| Benign | 78.5 | 82.5 | 84.1 |
| Malignant | 82.1 | 77.8 | 68.4 |
| TCMSVM | | | |
| Overall | 84.2 | 79.3 | 71.1 |
| Benign | 84.8 | 84.2 | 77.1 |
| Malignant | 83.6 | 71.3 | 59.6 |

Table 4.10: **Results comparing performance of TCMNN and TCMSVM on various ovarian cancer data sets:** These TCMNN tests were carried out using the standard settings, analysing to 1 nearest neighbour and using the standard Euclidean distance. The TCMSVM tests were run using the standard inner product.

I had then hoped that by using a simple polynomial kernel, separation of the two classes would be made possible, but this was not the case. When using the pCoMa system none of the polynomial kernel results were successful, as the results gave invalid $p$-values. When discussing the problem with the PhD students who designed the system, they stated that this was due to problems with the optimiser used in the underlying SVM implementation.

## 4.6 Abdominal pain data set

Due to the size of this particular data set the required calculations took considerably longer to complete that the calculations using the ovarian cancer data set did. Initial results of this data set were disappointing. As mentioned earlier, this could be due to problem of the curse of dimensionality. Some of the less relevant attributes could be distorting the formulation of the problem, causing the examples to be misclassified.

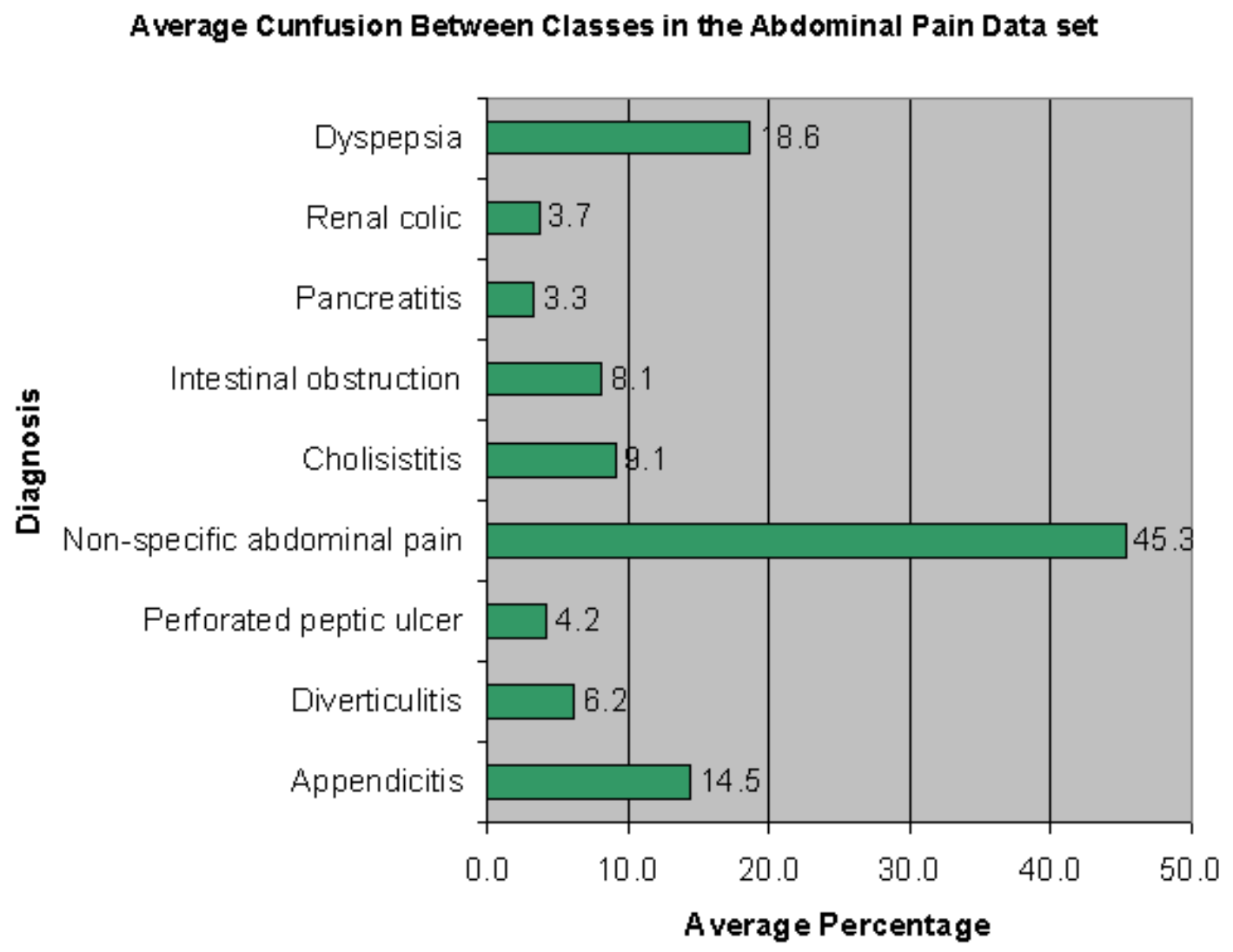

Figure 4.7: **Bar chart representing the average number of examples misclassified as each class for abdominal pain data set:** The percentage of examples misclassified for each class (confusion values) were taken from the performance of the TCMNN algorithm training with ABDO4387 and testing with ABDO2000 data sets. The test was run using 1 nearest neighbour with standard Euclidean distance.

As this data set had more than two classes, it was interesting to express how the different diagnoses became confused with each other as a percentage. This data was taken directly from the HTML output from my programs as seen in appendix F.

Individual Class Analysis for Class 0 (Appendicitis)
There were 259 Appendicitis examples in the test set
150 of the 259 were correctly classified
Accuracy of prediction for Appendicitis = 57.9%
Of the 109 examples incorrectly classified:

- 4.6% of them were misclassified as class (1) Diverticulitis

- 2.8% of them were misclassified as class (2) Perforated peptic ulcer

- 74.3% of them were misclassified as class (3) Non-specific abdominal pain

- 2.8% of them were misclassified as class (4) Cholisistitis

- 5.5% of them were misclassified as class (5) Intestinal obstruction

- 0.9% of them were misclassified as class (6) Pancreatitis

- 0.9% of them were misclassified as class (7) Renal colic

- 8.3% of them were misclassified as class (8) Dyspepsia

By looking at the incorrectly classified examples for each class like this we can build up a picture of which particular classifications are causing the errors in our predictions.

To see which classifications had more difficulty being classified than others, I took an average of the percentages of examples misclassified as each classification, giving the results shown in the bar chart of Figure 4.7.

As we can see, most cases seem to be misclassified as class 3 (non-specific abdominal pain). This classification is not the most useful of the diagnoses, as it does not actually specify what the problem is with the patient.

### 4.6.1 Abdominal pain - Test 1 $k$ nearest setting

The results in Table 4.11 show the effect of increasing the number of nearest neighbours used with the TCMNN algorithm on the abdominal pain data set. Overall accuracy of prediction is seen to increase from 60.4% to around 71.7% as the number of nearest neighbours is increased to 40.

Unfortunately with this improvement of the overall accuracy, the accuracy of prediction for diverticulitis cases decrease from 20.0% to 0.0%, and also Pancreatitis from 12.9% to 0.0%.

With analysing to increased nearest neighbours, average confidence is observed to increase from 76.9% to 83.9%. In contrast, average credibility is seen to decrease from 72.9% to 61.0%.

The fact that the results improve as the nearest neighbours are increased indicates that there may be a certain level of noise in the data. The performance achieved with 40 nearest neighbours of 71.7% is one of the best improvements achieved with all the tests.

| Neighbours | Percentage Accuracy % | | | | | |
|---|---|---|---|---|---|---|
| | 1 | 10 | 20 | 30 | 40 | 50 |
| Overall | 60.4 | 70.8 | 71.0 | 71.7 | 71.3 | 70.7 |
| Appendicitis | 57.9 | 58.3 | 58.3 | 59.4 | 60.2 | 59.0 |
| Diverticulitis | 20.0 | 0.0 | 0.0 | 0.0 | 0.0 | 0.0 |
| Perforated peptic ulcer | 50.0 | 57.1 | 52.4 | 50.0 | 50.0 | 45.2 |
| Non-specific abdominal pain | 71.2 | 87.6 | 89.0 | 90.3 | 90.5 | 90.4 |
| Cholisistitis | 49.0 | 60.0 | 60.0 | 59.5 | 56.5 | 56.5 |
| Intestinal obstruction | 44.9 | 54.3 | 51.2 | 51.2 | 47.2 | 44.1 |
| Pancreatitis | 12.9 | 0.0 | 0.0 | 0.0 | 0.0 | 0.0 |
| Renal colic | 53.1 | 61.9 | 59.9 | 58.5 | 58.5 | 56.5 |
| Dyspepsia | 58.9 | 67.2 | 67.2 | 68.7 | 68.7 | 68.7 |
| Average Conf | 76.9 | 85.1 | 84.9 | 84.4 | 83.9 | 83.5 |
| Average Cred | 72.9 | 62.1 | 61.2 | 61.0 | 60.9 | 61.0 |

Table 4.11: **Results of testing the abdominal pain data set using the TCMNN algorithm with different numbers of nearest neighbours:** The tests above were run using the TCMNN algorithm using the standard Euclidean distance with training ABDO4387 and testing ABDO2000 data sets mentioned earlier.

### 4.6.2 Abdominal pain - Test 2 Kernel distance metric

The results in Table 4.12 show that testing the use of polynomial kernels with the TCMNN algorithm on the abdominal pain data set .

The best overall performance observed was 61.7% obtained using a polynomial kernel of degree 5 with constant $c = 0$. This is a slight increase of 1.3% from that obtained using the standard Euclidean distance. As discussed earlier, this increase in accuracy using the polynomial degree 5 feature mapping implies that the abdominal pain problem can be better formulated by considering all possible combinations of 5 symptoms as the features. This then implies that there is a slight co-dependence between the 135 attributes, which can be used to improve learning.

The worst overall performance of 47.9% was observed when using polynomial kernel of degree 3 with constant $c = 0$. This gives a more drastic decrease in accuracy of 12.5%. Looking closely at the accuracy of prediction for each class, slight variations with each polynomial degree setting are observed.

As mentioned earlier, the use of the polynomial kernel allows the algorithm to compute the Euclidean distance between higher dimensional feature vectors without explicitly constructing the vectors. The usage of polynomial kernels on this example clearly highlights the computational shortcut achieved with this property.

The number of features created on the abdominal pain data set using a poly-nomial kernel of degree 2 with no constant $c = 0$ is $\binom{136}{2}$ which is $\frac{136!}{2\times134!} = 9180$ features.

| Results | Percentage Accuracy % | | | | |
|---|---|---|---|---|---|
| | Euclid | Poly deg 2 | Poly deg 3 | Poly deg 4 | Poly deg 5 |
| Overall | 60.4 | 60.4 | 47.9 | 60.4 | 61.7 |
| Appendicitis | 57.9 | 57.1 | 68.7 | 57.9 | 55.9 |
| Diverticulitis | 20.0 | 17.1 | 8.6 | 20.0 | 17.1 |
| Perfor peptic ulcer | 50.0 | 38.1 | 21.5 | 50.0 | 50.0 |
| Non-spec abdo pain | 71.2 | 71.9 | 58.7 | 71.2 | 73.0 |
| Cholisistitis | 49.0 | 48.5 | 21.0 | 49.0 | 49.0 |
| Intestinal obstr | 44.9 | 44.9 | 16.5 | 44.9 | 45.7 |
| Pancreatitis | 12.9 | 9.7 | 3.2 | 12.9 | 16.1 |
| Renal colic | 53.1 | 53.1 | 38.1 | 53.1 | 58.5 |
| Dyspepsia | 58.9 | 60.4 | 46.8 | 58.9 | 61.1 |
| Average conf | 76.9 | 78.6 | 74.8 | 77.1 | 78.7 |
| Average cred | 72.9 | 71.7 | 78.4 | 72.7 | 71.5 |

Table 4.12: **Results of testing the abdominal pain data set using the TCMNN algorithm with polynomial kernels of different degrees:** These tests were run using the TCMNN algorithm with the standard Euclidean distance metric and analysing to 1 nearest neighbour. The results were achieved training with ABDO4387 and testing with ABDO2000 data sets.

The results implied that mapping into these high dimensional feature space did not cause the Euclidean distance between the higher dimensional feature vectors to reflect anything more meaningful than in the original feature space.

### 4.6.3 Abdominal pain - Test 3 Significance level

Changing the significance level with the abdominal pain data set had far less effect on the accuracy of prediction than observed with the same test on the ovarian cancer data set. Table 4.13 shows the results of assessing the performance of the TCMNN algorithm predictions on the 4387 training and 2000 test sets to different significance level settings.

Up to 50% significance no increase in accuracy is observed. When looking at the plot of the credibility interval histogram in Figure D.4 in appendix C we can see that this is because the majority of examples are observed to have credibility greater than 40%.

At a significance level of 90% the overall accuracy is seen to increase from 60.1% to 81.5%. On the downside, at this setting 36.9% of examples are not classified.

| Significance | Percentage Accuracy % | | | | | | | | | |
|---|---|---|---|---|---|---|---|---|---|---|
| | 5 | 10 | 20 | 30 | 40 | 50 | 60 | 70 | 80 | 90 |
| Overall | 60.4 | 60.4 | 60.4 | 60.4 | 60.4 | 61.1 | 69.9 | 73.4 | 77.4 | 81.5 |
| Appendicitis | 57.9 | 57.9 | 57.9 | 57.9 | 57.9 | 56.9 | 56.4 | 55.5 | 55.9 | 64.3 |
| Diverticulitis | 20.0 | 20.0 | 20.0 | 20.0 | 20.0 | 20.0 | 17.4 | 17.6 | 21.4 | 28.6 |
| Perf peptic ulcer | 50.0 | 50.0 | 50.0 | 50.0 | 50.0 | 54.0 | 61.5 | 61.9 | 54.5 | 60.0 |
| Non-spec abdom | 71.3 | 71.3 | 71.3 | 71.3 | 71.3 | 72.0 | 81.3 | 84.5 | 87.9 | 91.9 |
| Cholisistitis | 49.0 | 49.0 | 49.0 | 49.0 | 49.0 | 49.7 | 55.9 | 60.2 | 63.5 | 69.7 |
| Intestinal obstr | 44.9 | 44.9 | 44.9 | 44.9 | 44.9 | 45.4 | 63.2 | 65.9 | 85.8 | 100.0 |
| Pancreatitis | 12.9 | 12.9 | 12.9 | 12.9 | 12.9 | 13.3 | 18.8 | 16.7 | 0.0 | 0.0 |
| Renal colic | 53.1 | 53.1 | 53.1 | 53.1 | 53.1 | 54.2 | 63.3 | 71.4 | 77.8 | 79.3 |
| Dyspepsia | 58.9 | 58.9 | 58.9 | 58.9 | 58.9 | 59.4 | 69.4 | 74.1 | 78.2 | 68.8 |
| Not classed | 0.0 | 0.0 | 0.0 | 0.0 | 0.0 | 3.3 | 30.0 | 46.2 | 63.4 | 80.9 |

Table 4.13: **Results of testing the abdominal pain data set using the TCMNN algorithm marking with different significance levels:** These tests were run using the TCMNN algorithm with the standard Euclidean distance metric and analysing to 1 nearest neighbour. The results were achieved training with ABDO4387 and testing with ABDO2000 data sets.

These results show that by setting a threshold value for the credibility of an example the overall accuracy of prediction can be increased quite considerably by 21.4%.

### 4.6.4 Abdominal pain - Test 4 performance of other learning algorithms

The results shown in Table 4.14 demonstrate the performance of different learning algorithms on the abdominal pain data sets.

Both KNN and DWKNN performed identically on the test sets. This is surprising as we would expect the results of these two algorithms to differ slightly due to the fact the DWKNN algorithm takes into consideration the actual distance between vectors, whereas KNN merely uses the majority of examples in each class to predict.

Another surprising result is that both DWKNN and KNN are seen to outperform the TCMNN algorithm slightly by 2.1%. Looking at the class accuracies for the $k$-Nearest Neighbour learning algorithms we notice that they perform similarly as well for each class.

The best overall accuracy on the abdominal pain test set of 67.5% was achieved by the neural network. Repeated tests with this algorithm shown in Table D.15 indicate a consistent performance at this level. This result is 7.1% higher than that of the TCMNN performance.

| Class | Percentage Accuracy % | | | | |
|---|---|---|---|---|---|
| | Neural Network | KNN | DWKNN | TCMNN | TCMSVM |
| Overall | 67.5 | 62.5 | 62.5 | 60.4 | - |
| Appendicitis | 74.1 | 53.3 | 53.3 | 57.9 | - |
| Diverticulitis | 0.0 | 14.3 | 14.3 | 20.0 | - |
| Perforated peptic ulcer | 0.0 | 45.2 | 45.2 | 50.0 | - |
| Non-spec abdom pain | 78.1 | 73.8 | 73.8 | 71.3 | - |
| Cholisistitis | 71.0 | 51.5 | 51.5 | 49.0 | - |
| Intestinal obstr | 0.0 | 46.5 | 46.5 | 44.9 | - |
| Pancreatitis | 0.0 | 12.9 | 12.9 | 12.9 | - |
| Renal colic | 74.1 | 63.9 | 63.9 | 53.1 | - |
| Dyspepsia | 78.4 | 63.0 | 63.0 | 58.9 | - |

Table 4.14: **Results of using different learning algorithms on the abdominal pain data set:** All $k$-Nearest Neighbour algorithms (TCMNN, KNN, DWKNN) were set analysing to 1 nearest neighbour, using the standard Euclidean distance. The neural network result was achieved using a 135-10-10-9 network. No result was available for the TCMSVM as it took too long to calculate. All tests were trained using ABDO4387 and tested using ABDO2000 data sets for fair comparisons.

Looking closer at the results of the neural network in Table 4.14 we see that the class accuracies are different to that obtained by the $k$-Nearest Neighbour based algorithms. The performance of the neural network was more worse than the $k$-Nearest Neighbour based algorithms with certain cases such as diverticulitis, perforated peptic ulcer, intestinal obstruction and pancreatitis.

No results with the TCMSVM algorithm were available due to the sheer size of the problem taking too long to compute the results. This is because the problem has 135 attributes and 9 different classifications and is far more complex than the binary classifications problems posed by the cancer data sets.

Some results using an inductive confidence machine SVM (ICMSVM) have been worked on by PhD students working in the Computer Learning Research Centre (Royal Holloway University of London) . However, these results are no better than that achieved by the TCMNN algorithm.

The abdominal pain data set could not be tested with the TCMNN using different Minkowski metrics because its inputs were binary. This meant that the difference between individual features $|a_i - b_i|$ was also 0 or 1. Using different powers with the Minkowski metric would therefore return the same answer.

### 4.6.5 Abdominal pain - Test 5 Further experiments with neural networks on abdominal pain data set

Due to the surprisingly good results of the neural networks performance on the abdominal pain data set seen in Tables D.15 and 4.14, I decided to run further tests using neural networks.

The neural network used in the tests earlier was of 135-10-10-9 dimension. As with the neural network used on the ovarian cancer data set, a 1 of N output encoding was used. The results were obtained after 100,000 weight updates.

| | Percentage Accuracy % | |
|---|---|---|
| Performance | Original | Augmented |
| Overall | 57.5 | 59.6 |
| Average Conf | 77.4 | 82.8 |
| Average Cred | 71.8 | 67.6 |

Table 4.15: **Summary of augmented data set tests with TCMNN on abdominal pain data set:** The results are averages of the values observed in the random split tests run with augmented data sets shown in Table D.16. The TCMNN was set to standard setting of 1 nearest neighbour, using Euclidean distance.

Table 4.15 shows the results of performing the tests using the augmented data set technique outlined earlier in Figure 4.6 on the abdominal pain data set. The results in this table are averages of the values taken from repeated random split tests.

The results of these individual random split tests can be seen in Table D.16 of appendix D. These tests were carried out using a random split of 1000 examples of the ABDO4387 data set. The performance of the TCMNN algorithm was recorded before using the original version of the data sets, and after using the augmented version of the data sets created by the neural network.

The summary of these results in Table 4.15 show an average increase in overall performance of 2.1%. Using a simple two tailed $t$-test (outlined earlier in Equation 4.5) on the individual experiments results in Table D.16 gives a critical value of:

$$t = \frac{15}{\sqrt{\frac{7(71.24)-(15)^2}{(6)}}} = 2.221$$

Analysing to 6 degrees of freedom and taking a significance level of 0.1 gives a critical value of 1.943 from statistical tables. Our observed value $t$=2.221 > 1.943, therefore the differences observed are 90% significant. This difference is not as significant as observed with the ovarian data set.

Average credibility is seen to increase from 77.4% to 82.8%, whilst average credibility is seen to decrease from 71.8% to 67.6%. These results indicate that the augmentation process makes the examples stranger from one another while making it easier to distinguish between classes.

The most surprising consequence of these results is not the fact that the network improves the accuracy of the results slightly, but that the network maps the full 135 features to just 20. Although this mapping using the network is a complex function combining the inputs, the fact that the inputs can be successfully mapped to a much smaller set of inputs indicates that there is redundancy in the inputs.

The fact that the TCMNN algorithm can improve its performance using this smaller augmented data set means that the process is indeed capturing the important features of the problem as I had hoped. This once again implies co-dependence between the input features used to describe the data set.

# Chapter 5

# Conclusion and Future Work

## 5.1 Self assessment of progress

The initial aims of the project were to develop a system for transductive confidence techniques and apply it to the medical data set provided by St. Bartholomews hospital.

All of the original objectives outlined earlier on page 18 have been completed successfully:

1. The underlying theory of the TCM was thoroughly investigated. The $k$-Nearest Neighbour algorithm was fully explored.

2. The TCMNN algorithm was successfully implemented in Java. A simple user interface was developed as a front end to this algorithm.

3. The TCMNN testing system was applied to various medical data sets, namely the abdominal pain, ovarian cancer and Wisconsin breast cancer data sets.

The project has also achieved the extensions initially outlined, investigating ideas such as:

- Neural networks
- Polynomial kernels
- Minkowski metric
- Analysing to increased nearest neighbours
- Marking TCMNN results with significance
- Testing TCMNN on 2-dimensional artificial data sets (pattern and regression)

Overall I am pleased with the user friendliness of the system developed. Although the system is not in use by the hospital, the simple windows interface should be common and familiar to the staff. Unfortunately there was no time to obtain feedback from the hospital to assess any further requirements to the system.

In retrospect I believe that the system could have been developed quicker if I used a visual Java development environment such as Symantec's Visual Cafe to create the user interfaces. Coding all of the programs interfaces manually has developed my understanding of the Java language.

It was disappointing that the system was not able to run as an applet in the users web browser, as achieved by the *pCoMa* system [22].

The experiments with the data sets gave very interesting results. In particular using neural networks and using the Minkowski metric with the TCMNN algorithm gave results that were very surprising. I feel that I have thoroughly investigated the applications of the transductive confidence technique and have opened up new areas for further research.

Over the course of the project I feel that there have been many learning outcomes on my part. For instance I have learnt many new skills such as:

- How to structure and perform large scale experiments.
- How to develop large systems by breaking problems down into simpler component objects.
- How to develop professional looking and functional graphical user interfaces.
- Writing large technical documents with LaTex.

I also feel that I have developed my knowledge in the theoretical and mathematical background to machine learning.

In conclusion a summary of the results and the possible reasons for the observations are as follows:-

## 5.2 TCMNN results

### 5.2.1 $k$ nearest setting

Changing the number of $k$ nearest neighbours that the TCMNN algorithm analysed to had different effects on each of the data sets. In the case of the ovarian cancer and USPS data sets, an increase in the number of nearest neighbours decreased the accuracy of the TCMNN. In contrast, an increase in the number of nearest neighbours with the abdominal pain and Wisconsin breast cancer data sets did improve accuracy.

### 5.2.2 Minkowski metric

Using the Minkowski metric with powers $k$ such that $0 < k < 2$ improved the accuracy performance of the TCMNN on all medical data sets as compared to the results achieved using the standard Euclidean distance measure. In contrast, the more commonly used powers of $k$ such that $k > 2$ decreased the accuracy of prediction on all results.

The improvement observed with fractional powers of $k$ could be due to the geometrical properties of this measure assisting problems with the discrete nature of the attributes of the data sets.

Tests with the Minkowski metric had adverse affects on the USPS data set. Fractional powers of $k$ were observed to decrease accuracy, whilst the standard values $k > 2$ increase accuracy. This adverse results could be due to the features of the USPS data set being continuous real values.

### 5.2.3 Polynomial kernels

Using the polynomial kernels to allow the TCMNN algorithm to perform in a higher dimensional space gave mixed results. No marked improvements over the standard Euclidean distance were observed with the use of polynomial kernels on the ovarian cancer data sets. In contrast slight improvements were observed with the abdominal pain, Wisconsin breast cancer and USPS data sets. This improvement suggests that the classification task is dependent on interactions of attributes in these data sets.

### 5.2.4 Significance

Marking the results of the TCMNN algorithm with significance improved the accuracy of prediction for all data sets to varying degrees.

## 5.3 Results with other learning algorithms

### 5.3.1 Neural networks

The results showed that using a neural network outperformed all the other learning algorithms on all of the medical data sets tested. The results of the neural networks were further improved by placing bias on particular classifications that the algorithm had problems with.

Using neural networks to create the augmented versions of the data sets improved the classification accuracy of the TCMNN on all of the medical data sets.

### 5.3.2 Traditional $k$-Nearest Neighbour

Both DWKNN and KNN algorithms performed equally as well on each of the data sets. This implies that the Euclidean distance between vectors did not give much extra meaning to the true 'difference' between the examples in the data sets.

The TCMNN algorithm was seen to outperform the traditional versions DWKNN and KNN algorithms on the ovarian cancer data set. In contrast the DWKNN and KNN algorithms outperformed the TCMNN algorithm on the abdominal pain and Wisconsin breast cancer data sets.

### 5.3.3 TCMSVM

On average there was very little difference observed between the performance of the SVM and nearest neighbours implementation of the transductive confidence machine.

The results of the Wisconsin breast cancer data set suggested that the TCMNN and TCMSVM performed similarly using polynomial kernels.

## 5.4 Summary of results of each data set

### 5.4.1 Ovarian cancer

Ovarian cancer is a disease produced by the rapid growth and division of cells within one or both ovaries [28].

The ovarian cancer data set was the smallest of all the data sets tested. Adjusting all of the parameters explored in the TCMNN algorithm did not improve the accuracy to match the performance of the hospitals' simple RMI screening technique. In contrast, the neural network was the only algorithm seen to improve on the RMI results.

The addition of more examples to this data set was seen to decrease the accuracy of prediction by the learning algorithms. Overall the data set proved to be a very difficult problem to learn from. This could be due to the lack of attribute information available to effectively describe the problem.

What must be understood when weighing up the effectiveness of the algorithm on the ovarian cancer data set is that the disease has inherent complexities due to its multi step development. Human tumours usually develop in progressive stages, resulting finally in an invasive, metastasizing, malignant cancer [1].

Natural cancers are thought to result from the interaction of multiple events over time. In addition, it is believed that those individuals with a genetic pre-disposition to cancer tend to 'skip' a few steps required for a tissue to become metastatic and malignant. They are therefore at a higher risk of developing cancer at some stage in their life.

Perhaps the training data contained a mixture of examples of patients with and without the predisposition to cancer, which caused confusion with diagnosis. Perhaps the 3 attributes given were not enough to distinguish those patients with malignant tumours and those with benign tumours.

Factors such as genetic pre-disposition, lifestyle, environment and age may need to be taken into account in addition to the 3 attributes to give more distinguishable results. Introducing features such as these may give results that are more clustered and distinguishable from each other.

Specifically known linked risk factors [27] of this disease could be incorporated as new attributes, for example:-

- Family history of the disease
- A history of infertility
- Never having children
- Early menstruation
- History of breast, endometrial or colorectal cancer
- Late menopause
- Obesity

### 5.4.2 Abdominal pain

This was the most complex of all the medical data sets tested. The results of the augmented version of the data set using the TCMNN suggested that there was inherent redundancy in the attributes.

The tests analysing to increased numbers of nearest neighbours suggested that there was a certain level of 'noise' in the data. Results from testing with polynomial kernels and the augmented data set indicated that there was some interaction between the 135 symptoms used in the problem.

This data set demonstrated the computational limitations of the TCMSVM algorithm, in that calculation time for this problem was not feasible. Perhaps the use of the augmented version of the data set using only 20 attributes could be used to obtain results with the TCMSVM algorithm.

The abdominal pain data set proved the hardest to learn from. All the algorithms tested performed consistently worse with this data set. The problem could be because different attributes used to describe the problem are more important to some diagnoses than others, causing confusion when considering the problem with all 135 attributes.

### 5.4.3 Wisconsin breast cancer

The performance achieved using this data set was far greater than the abdominal pain and ovarian cancer data sets tested. Configuring the TCMNN parameters were seen to give slight increases in the data set's accuracy of prediction.

Tests using the augmented version of the training set provided the least improvement to the performance as compared to the other data sets. This suggested that the data set was a lot 'cleaner' than the other two data sets. Results with polynomial kernels indicated that there was some interaction between the symptoms, which could be exploited to improve performance.

Unlike with the other data sets tested, marking the results of the TCMNN with significance enabled specificity and sensitivity to be increased to 100%.

### 5.4.4 USPS

The results of performing TCMNN tests on the USPS data sets can be seen in appendix C.

Increasing the number of nearest neighbours with the TCMNN on this data set decreased accuracy. This suggested that the data was very 'clean'. The use of polynomial kernels on the data reflected very little change in performance, implying that there was negligible dependence between pixels. Using Minkowski metrics on this data set supported my hypothesis that fractional powers would mainly assist problems with discrete inputs. The Minkowski tests also implied that data sets with real valued inputs may be assisted by powers $k > 2$.

## 5.5 Summary of conclusions

- The performance of the TCMNN did not significantly improve the prediction of the underlying $k$-Nearest Neighbour algorithm.
- Using different learning algorithm implementations of the TCM did not seem to affect the performance.
- The TCM appeared to suffer more than traditional inductive algorithms on 'noisy' data sets.
- Some improvement was made to the TCMNN algorithm performance by using different distance metrics and polynomial kernel feature mappings.
- Neural networks outperformed transductive algorithms on noisy medical data sets.
- Marking the TCMNN results with significance helped to improve specificity and sensitivity of the test.

## 5.6 Future extensions

### 5.6.1 Deliver existing system over the Web

One obvious extension to my existing system would be to convert its functionality to work as an applet in a web page. As discussed earlier in the design chapter, there are a number of techniques that could be employed to provide the ability to read and write to files.

Given sufficient time I would choose to design some Java servelet/cgi/perl backbone to the system to enable communication between the server and the user's machine. By making the system available over the web would mean that it would be far easier to use and access.

Because the current system is written in Java and constructed using Swing components, very little changes to the code would be needed to convert the application into an applet.

### 5.6.2 Implementing TCMSVM

The project implemented the TCMNN algorithm. I had also hoped to implement the TCMSVM for using in the experiments. The major problem that was encountered was that the SVM algorithm normally requires a quadratic optimisation package such as MINOS [25] or LOQO [26] to obtain its solutions. If I tried to develop a similar system for the TCMSVM in Java, then I would require the program to communicate with the optimiser.

This would also mean that when distributing the system, the user would also need access to the optimiser for the system to function. This of course causes further complications in using the system. There is also the problem that the free versions of these optimisers have limitations to the size of the problems they can solve.

In order to get around this problem, the SMO optimiser could be implemented to find solutions for the SVM [3]. This optimiser is very simple to implement and is specifically designed to solve problems posed by the SVM. Another even simpler option would be to implement the kernel adatron algorithm which is an SVM approach based on basic linear perceptrons.

### 5.6.3 Extending neural networks for generating confidences

As mentioned earlier the neural networks used in the experiments used a 1 of $N$ output encoding. This meant that each possible classification was designated an output unit in the network. After feeding an example through the network, the highest values in the output units decided the prediction. These values could then be seen as crude measures of confidence and credibility.

Since neural networks were observed to outperform the TCMNN in the experiments, it would be interesting to compare the confidence and credibility of the neural network compared to the values constructed by the TCM.

Taking these values further one could then use them as strangeness values for that completion. These strangeness values could then be used by the $p$-value function (Equation 2.2) to construct a TCM implementation of neural networks.

Special care would have to be taken to ensure that the exchangeability conditions of the TCM framework are withheld to return valid $p$-values. Neural networks provide solutions which are dependent on the order in which the examples are presented, therefore an arbitrary step of sorting the training examples would have to be performed before use. Another complication is that the weights of the neural network are usually initialised to random values. For TCM Neural Networks the initial weights would have to be kept constant as differences would affect the predictions of the network.

### 5.6.4 Using kernels with the Minkowski metric

The Euclidean distance is mathematically equivalent to Minkowski metric degree 2, and thus belongs to the set of Minkowski metrics. It would therefore be interesting to investigate the effect of polynomial (or other kernels) with the Minkowski metric to various degrees.

### 5.6.5 Further work with image data sets

From the initial work carried out using the TCMNN algorithm on image data sets, it would be interesting to investigate techniques to help learning algorithms cope with translations of the images.

### 5.6.6 Developing more complex kernels

From the initial results obtained using polynomial kernels with the TCMNN algorithm it would be interesting to investigate the use of more complex kernels. It would also be interesting to compare the effectiveness of using the same kernel on both SVM and nearest neighbours algorithms.

### 5.6.7 Automated configuration of the TCMNN using genetic algorithms

When testing the TCMNN at present with the various parameters, it is necessary to manually configure and run the test. This process is then repeated and the results are compared to search for an optimal configuration. One practical extension would be to automate this process so that these optimal settings can be found for any given data set.

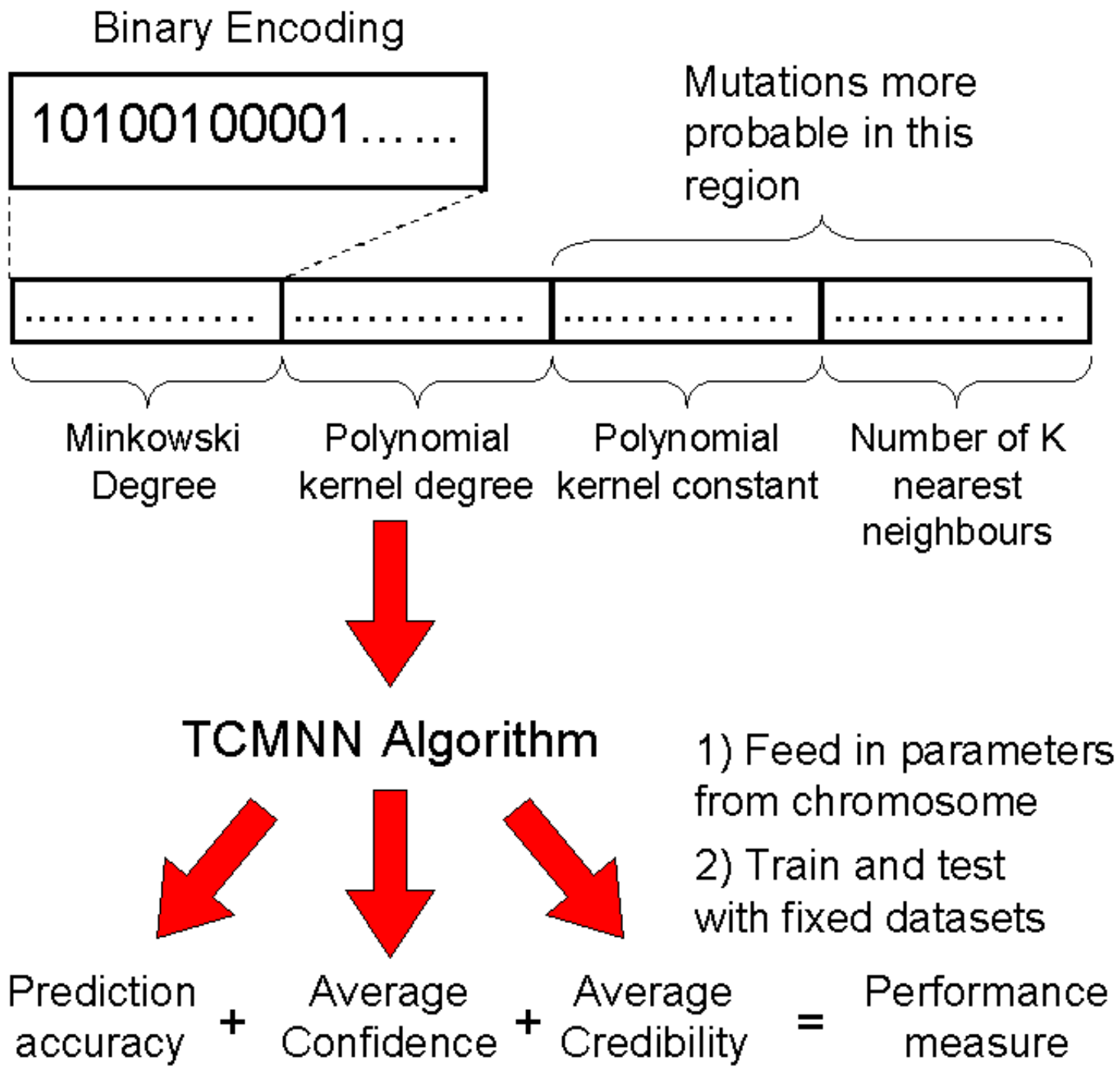

Figure 5.1: **An example of using a genetic algorithm to automatically configure the TCMNN algorithm**

This problem could be solved using a genetic algorithm as a heuristic to configure these parameters as shown in Figure 5.1 by encoding the various parameters of the TCMNN as a binary string 'chromosome'. A genetic algorithm would then create a genetic population of these chromosomes and perform random mutations and crossovers. These parameters could then be extracted from the chromosomes to configure and run a TCMNN on a fixed data set. The performance of the TCMNN could then be used to rank how successful that mutation was amongst the rest of the gene pool of settings.

Particular weighings could be introduced to make the mutations of specific regions of the chromosome more likely, such as the number of nearest neighbours.

This approach to the problem has the advantage that the genetic algorithm automatically keeps track of the best configuration of the TCMNN. These crossover and mutation processes of the genetic algorithm could be seen as combining successful combinations of TCMNN settings, and slightly altering parameters. Since the experiments show that results of traditional nearest neighbours are similar to the results of the TCMNN, one could possibly evaluate the ranking of the chromosome using the DWKNN (instead of the TCMNN) to accelerate the technique.

### 5.6.8 Implementing TCMNN regression

This project has concentrated on pattern recognition problems. If given more time I could also investigate using the TCM on regression cases.

As mentioned previously, I implemented the distance weighted $k$-Nearest Neighbour regression algorithm. This algorithm was demonstrated using a simple 2 dimensional example application as shown earlier in Figure 3.12.

An obvious extension would be to develop from this a transductive confidence version of this algorithm (TCMNN regression). This would be achieved by breaking the targeted continuous real valued output into intervals and using the normal TCM technique to predict confidence for each interval.

# Bibliography

[1] *Molecular Cell Biology* $2^{nd}$ *edition* , Damell, Lodish, Baltimore, WH Freeman & Company, (1990), 994, 995

[2] *Pattern Classification* $2^{nd}$ *edition* , R. Duda, P. Hart, D. Stork, J. Wiley & Sons, (2001) 187, 184

[3] *An Introduction to Support Vector Machines* N. Cristianini, J. Shawe-Taylor, Cambridge University Press, (2000), 26

[4] *Machine Learning* T. Mitchell, McGraw-Hill, (1997), 97, 98, 106, 231, 238

[5] *Computational Learning and Probabilistic Reasoning* edited by A. Gammerman, John Wiley & Sons, (1996), 4

[6] *An Introduction to Kolmogorov Complexity and Its Application* $2^{nd}$ *edition* M. Li, P. Vitányi, Springer, (1997), 127, 136

[7] *Pattern recognition via linear programming: Theory and application to medical diagnosis* O. L. Mangasarian, R. Setiono, and W.H. Wolberg:, SIAM Publications, Philadelphia 1990, pp 22-30.

[8] *Statistics for people who (think they) hate statistics* , N. Salkind, Sage Publications (2000) 169, 193

[9] *Java in a Nutshell, Third edition* D. Flanagan, O'Reilly Publishers, (1999) 3, 6, 9

[10] *The JFC Swing Tutorial, Third edition* K. Walrath, M. Campione, (2000) 5, 17, 19, 343, 387, 443

[11] *Statistical Learning Theory* V. Vapnik, J. Wiley & Sons, (1998), 501

[12] Gammerman A., Vovk V., *Kolmogorov Complexity: Sources, Theory and Applications*, (1999), The Computer Journal: 42

[13] I. Jacobs, D. Oram, J. Fairbanks, J. Turner, C. Frost, J. G. Grudzinskas, A risk of malignancy index incorporating CA 125, ultrasound and menopausal status for the preoperative diagnosis of ovarian cancer, *British Journal of Obstetrics and Gynaecology* (1990), 97, 922-929

[14] Proedrou K., Nouretdinov I., Volodya V., Gammerman A., (2001), *Transductive Confidence Machine for Pattern Recognition*, Department of Computer Science, Royal Holloway, University of London, Surrey, TW20 OEX

[15] Sarkar M., Leong T., Medical Computing Laboratory, Department of Computer Science, National University of Singapore, Lower Kent Ridge Road (2000) *Application of K-Nearest Neighbors Algorithm on Breast Cancer Diagnosis Problem*

[16] Mangasarian O. L. and Wolberg W. H.: *Cancer diagnosis via linear programming*, (1990), SIAM News, 1, 18, 23

[17] Wolberg W.H. and Mangasarian O.L.: *Multisurface method of pattern separation for medical diagnosis applied to breast cytology* , (1990), Proc. Nat. Aca. Sci: 87, 9193-9196.

[18] Bennett K. P., Mangasarian O. L., : *Robust linear programming discrimination of two linearly inseparable sets*, (1992), Optimization Methods and Software: 1, 23-34.

[19] Gammerman A., Vovk V., *Prediction algorithms and confidence measures based on algorithmic randomness theory*, (2002), to appear in Theoretical Computer Science.

[20] Saunders C., Gammerman A., Vovk V., Transductive with confidence and credibility, in: *Proceedings of the Sixteenth International Joint Conference on Artificial Intelligence II* (Morgan Kaufmann, 1999) 722-726.

[21] Vovk V., Gammerman A. and Saunders C., Machine learning applications of algorithmic randomness, in: Bratko I. and Dzeroski S., eds., *Procceedings of the Sixteenth International Joint Conference on Machine Learning*, (Morgan Kaufmann, 1999, San Francisco, CA, 1999) 444-453

[22] `http://nostradamus.cs.rhul.ac.uk/~leo/pCoMa/` *Transductive Confidence Machine web site* developed by I. Nouretdinov, D. Surkov, L. Gordon.

[23] `http://www.ics.uci.edu/~mlearn/MLSummary.html` *UCI machine learning resource web site.*

[24] `http://www.webdeveloper.com/java/java_jj_read_write.html` *Java Jive: File I/O with Java: It can be done*

[25] http://www.sbsi-sol-optimize.com/Minos.htm *Minos 5.5 quadratic optimiser web site*

[26] http://www.orfe.princeton.edu/~loqo/ *LOQO optimisation and applications web site*

[27] http://www.ovariancanada.org/facts/what/canada.php *Canadian national ovarian cancer association web site: What is ovarian cancer*

[28] http://www.oncologychannel.com/ovariancancer/ *Oncology world web resource*

[29] http://java.sun.com/docs/books/tutorial/security1.2/tour1/ *Sun web site, Java Tutorial : Quick tour of controlling applets - Observe applet restrictions, Grant the required permission, See the policy file effects*

[30] http://developer.java.sun.com/developer/onlineTraining/Programming/... BasicJava1/data.html *Sun web site, Java Developer Connection - File access and permissions*

[31] http://www.tek271.com/articles/java_2_applet_security.htm *Java 2 Applet Security, online article by Abdul Habra (2000)*

# Appendix A

# Results with Wisconsin breast cancer data set

This was the last medical data set to be tested. The data set was obtained using the UCI machine learning resource web site [23]. The data originally came as one combined data set of 683 examples (WBC683). To run fast and separate tests I randomly created two further 433 training (WBC433) 250 test (WBC250) data sets.

| Results | Percentage accuracy % | |
|---|---|---|
| | Separate (433,250) | Leave one out (683) |
| Overall | 96.4 | 95.5 |
| Benign | 97.2 | 97.3 |
| Malignant | 94.4 | 92.1 |
| Average Conf | 99.5 | 99.5 |
| Average Cred | 64.9 | 59.1 |

Table A.1: **Initial TCMNN test accuracy results using Wisconsin breast cancer data sets of different sizes:** These TCMNN tests were carried out using the standard settings, analysing to 1 nearest neighbour and using the standard Euclidean distance.

Table A.1 shows initial results of testing the TCMNN on the different Wisconsin breast cancer data sets. The results of the leave one out test on the full 683 examples were slightly worse by 0.9% than the separate training 433 test 250.

Average confidence was the same for both tests. The average credibility was less by 5.8% for the leave one out test. One would expect that the more training data presented in the learning process would improve accuracy. These results show the contrary, implying that perhaps there was some level of noise in the data, probably introduced by the fact that the data had been compiled by several hospitals.

## A.1 Test 1 TCMNN - $k$ nearest setting

The results of running the TCMNN analysing to different numbers of nearest neighbours can be seen in Table . The results show that this setting does improve performance.

| K value | Percentage accuracy % | | | | |
|---|---|---|---|---|---|
| | Overall | Benign | Malignant | Average Conf | Average Cred |
| 1 | 96.4 | 97.2 | 94.4 | 99.5 | 64.9 |
| 5 | 98.0 | 98.3 | 97.2 | 99.6 | 57.4 |
| 10 | 98.0 | 98.3 | 97.2 | 99.5 | 55.2 |
| 15 | 98.0 | 98.9 | 95.8 | 99.5 | 54.5 |
| 20 | 98.0 | 98.9 | 95.8 | 99.5 | 54.4 |
| 25 | 98.0 | 98.9 | 95.8 | 99.5 | 54.4 |

Table A.2: **Results of TCMNN testing the Wisconsin breast cancer data set with different numbers of nearest neighbours:** These tests were carried out with the separate training WBC433, testing WBC250 files described earlier. The TCMNN algorithm was performed using the standard Euclidean distance metric.

Increasing to any number greater than 5 nearest neighbours improved the overall accuracy from 96.4% to 98.0%. Looking carefully at the results, we notice that although $k \geq 5$ gave the same performance, 5 and 10 nearest neighbours had a higher malignant accuracy of 97.2% compared to $k \geq 15$ which gave 95.8%.

The results showed little change in average confidence with respect to changes in number of nearest neighbours. Average credibility was seen to decrease from 64.9% to 54.4%, which indicated that increasing the number of nearest neighbours caused the examples to become stranger from one another.

## A.2 Test 2 TCMNN - Minkowski Metric

The results from testing the Minkowski metric on the Wisconsin breast cancer data set are shown in Table A.3. The best overall accuracy of 97.2% was achieved using the Minkowski measure to degree $\frac{1}{3}$, $\frac{1}{4}$ and $\frac{1}{5}$. This accuracy was an improvement of 0.8% on the standard Euclidean distance.

Notice once again that the results of the Minkowski metric degree 2 were the same as those obtained using normal Euclidean distance. The results show that as the Minkowski degree is increased to values greater than 2, the overall accuracy decreases from 96.4% to 95.6%.

| K value | Percentage accuracy % | | | | |
|---|---|---|---|---|---|
| | Overall | Benign | Malignant | Average Conf | Average Cred |
| $\frac{1}{5}$ | 97.2 | 97.8 | 95.8 | 99.5 | 63.6 |
| $\frac{1}{4}$ | 97.2 | 97.8 | 95.8 | 99.5 | 63.6 |
| $\frac{1}{3}$ | 97.2 | 97.8 | 95.8 | 99.5 | 63.5 |
| $\frac{1}{2}$ | 96.8 | 98.3 | 92.9 | 99.5 | 63.6 |
| 1 | 96.8 | 98.3 | 92.9 | 99.5 | 64.6 |
| 2 | 96.4 | 97.2 | 94.4 | 99.5 | 64.9 |
| 3 | 96.0 | 96.6 | 94.4 | 99.4 | 65.5 |
| 4 | 95.6 | 96.1 | 94.4 | 99.3 | 65.8 |
| 5 | 95.6 | 96.1 | 94.4 | 99.3 | 65.8 |

Table A.3: **Results of TCMNN testing the Wisconsin breast cancer data set using the Minkowski distance metric with different powers $k$:** These tests were carried out with the separate training WBC433, testing WBC250 files described earlier. The TCMNN algorithm was performed analysing to 1 nearest neighbour.

The results in Table A.3 showed little change in average confidence as the Minkowski metric degree was changed. In contrast, average credibility decreases by 1.3% with Minkowski metric degrees $d < 2$, and increases by 1.2% for values $d > 2$.

This indicated that using the Minkowski metric to fractional powers caused the examples to become stranger from one another. As mentioned earlier this improvement using these powers could be due to the discrete nature of the features of this data set (each input is an integer between 1 to 10).

## A.3 Test 3 TCMNN - Kernel distance metric

The results of using polynomial kernels of different degrees $d$ and constant $c = 0$ with the Wisconsin breast cancer data set can be seen in Table A.4.

The results showed that using a polynomial kernel with degree 2 decreased the overall accuracy by 0.9% compared to that obtained using the standard Euclidean distance in the original input space. In contrast, improvements were observed when using polynomial kernels of degree 3, 4 and 5.

The best improvement was obtained using a polynomial kernel of degree 5, increasing the accuracy from 96.4% to 98.0%. With all these kernel results, very little

| Results | Percentage accuracy % | | | | |
|---|---|---|---|---|---|
| | Euclid | Poly deg 2 | Poly deg 3 | Poly deg 4 | Poly deg 5 |
| Overall | 96.4 | 95.5 | 97.6 | 97.6 | 98.0 |
| Benign | 97.2 | 97.1 | 98.3 | 98.3 | 98.3 |
| Malignant | 94.4 | 92.5 | 95.8 | 95.8 | 97.2 |
| Average Conf | 99.5 | 99.5 | 99.5 | 99.5 | 99.5 |
| Average Cred | 64.9 | 59.0 | 65.2 | 65.3 | 65.2 |

Table A.4: **Results of testing the Wisconsin breast cancer data set using a polynomial kernel with various degrees $d$:** These tests were carried out with the separate training WBC433, testing WBC250 data sets described earlier. The TCMNN algorithm was performed analysing to 1 nearest neighbour, using the standard Euclidean distance. All polynomial kernel constants in these tests were set $c = 0$.

change in the average confidence and credibility values are observed.

| Results | Percentage accuracy % | | | | | |
|---|---|---|---|---|---|---|
| | Euclid | $c$=0 | $c$=1/2 | $c$=10 | $c$=100 | $c$=1000 |
| Overall | 96.4 | 95.5 | 97.2 | 96.8 | 96.4 | 96.4 |
| Benign | 97.2 | 97.1 | 97.7 | 97.8 | 97.8 | 97.2 |
| Malignant | 94.4 | 92.5 | 95.8 | 94.4 | 93.0 | 94.4 |
| Average Conf | 99.5 | 99.5 | 99.5 | 99.4 | 99.5 | 99.5 |
| Average Cred | 64.9 | 59.0 | 65.2 | 65.2 | 64.9 | 64.8 |

Table A.5: **Results of TCMNN testing the Wisconsin breast cancer data set using a polynomial kernel with various constants $c$:** These tests were carried out with the separate training WBC433, testing WBC250 data sets described earlier. The TCMNN algorithm was performed analysing to 1 nearest neighbour, using the standard Euclidean distance. All polynomial kernel degrees were set $d = 2$.

Table A.5 shows further tests using polynomial kernels, this time fixing the degree to 2 and trying various constants $c$.

The best accuracy of the TCMNN was observed when setting the constant $c = 1/2$. This produced an overall accuracy of 97.2% which was better than both that of the standard Euclidean distance (96.4%) and the polynomial constant $c = 0$ (95.5%).

## A.4 Test 3 Significance level

The results with assessing the TCMNN performance on the Wisconsin breast cancer data set to different significance levels can be seen in Table A.6.

| | Percentage accuracy % | | | |
|---|---|---|---|---|
| Significance level | Overall | Benign | Malignant | Not Classified |
| 5% | 96.4 | 97.2 | 94.4 | 0.0 |
| 10% | 97.9 | 97.7 | 98.5 | 3.2 |
| 15% | 98.7 | 98.8 | 98.4 | 5.2 |
| 20% | 99.1 | 98.8 | 100.0 | 9.6 |
| 25% | 99.1 | 98.8 | 100.0 | 10.0 |
| 30% | 100.0 | 100.0 | 100.0 | 13.6 |

Table A.6: **Results of TCMNN testing the Wisconsin breast cancer data set using different significance levels:** These tests were carried out with the separate training WBC433, testing WBC250 data sets described earlier. The TCMNN algorithm was performed analysing to 1 nearest neighbour, using the standard Euclidean distance.

The results show that to achieve an accuracy of prediction of 100% we can mark to 30% significance. At this level only 13.6% of examples are not classed. This result is impressive, meaning that marking with significance could potentially guarantee correct predictions every time.

To test if these results were not due to the effect of the particular split of the data I made, I reran the test using a leave one out test on the whole 683 examples. These results can be seen in Table D.18 in appendix D. These results also show the same phenomena of having 100% accuracy when marking at 30% significance.

## A.5 Test 4 Performance of other learning algorithms

Table A.7 compares the performances of different algorithms on the Wisconsin breast cancer data set. Once again, the best results were achieved by the Neural network with 98.0% overall accuracy.

All of these tests were run on the same fixed separate training and testing data sets so as to make the comparison fair.

Once again the traditional versions of the $k$-Nearest Neighbour algorithm outperform the TCMNN by 0.8%. Both DWKNN and KNN implementations achieve the same accuracy of prediction.

The results of the tests run in [15] boast an accuracy of 98.3% ,slightly better than my neural network result. However the results in [15] do not mention specifically what size split of training and test sets are being used. Therefore these results cannot be fairly compared to the results in my tests.

| Algorithm Used | Overall % | Benign % | Malignant % |
|---|---|---|---|
| Neural Network | 98.0 | 97.8 | 98.6 |
| DWKNN | 97.2 | 97.2 | 97.2 |
| KNN | 97.2 | 97.2 | 97.2 |
| TCMNN | 96.4 | 97.2 | 94.4 |
| TCMSVM | 96.8 | 96.6 | 97.2 |
| $\diamondsuit$ Results from [15] | 98.3 | 99.6 | 85.8 |

Table A.7: **Results of testing the Wisconsin breast cancer data set using different learning algorithms:** Each of these learning algorithms were trained using WBC433 and tested with WBC250 data sets to ensure a fair comparison. All versions of the $k$-Nearest Neighbour algorithm (TCMNN, KNN and DWKNN) were set to standard settings of 1 nearest neighbour, using Euclidean distance. The TCMSVM results was obtained using the standard inner product. The size of training and testing sets for results obtained from $\diamondsuit$ are not specified.

## A.6 Test 6 Further experiments with neural networks on Wisconsin breast cancer data set

As with the abdominal pain and ovarian cancer data sets, I decided to try the same technique of using augmented data sets to improve performance of the TCMNN algorithm. The results in Table A.8 show the averages of the repeated tests of this augmentation process.

The individual tests can be seen in Table D.19 in appendix C. When running a two tailed $t$-test statistic in these results it was found that there was no significance in the differences.

The results of this test were disappointing compared to that observed with the other medical data sets. The results in Table A.8 showed a very small average increase of only 0.3%.

The poor performance in this test may be because there was not as much redundancy in the inputs as found with the other medical data sets, therefore the technique could not improve the TCMNN results. This would link in with the fact that the accuracy for the Wisconsin data set was much higher than the others.

| Training Data | Percentage Accuracy % | | | | |
|---|---|---|---|---|---|
| | Overall | Benign | Malignant | Average Conf | Average Cred |
| Original | 96 | 97.3 | 92.2 | 99.5 | 63.4 |
| Augmented | 96.3 | 96.9 | 93.5 | 99.6 | 64.7 |

Table A.8: **Summary of augmented data set tests with TCMNN on Wisconsin breast cancer data set:** The results are averages of the values observed in the tests run with augmented data sets shown in Table D.19. The TCMNN was set to standard setting of 1 nearest neighbour, using Euclidean distance.

## A.7 Comparing performance of TCMSVM with TCMN-N on Wisconsin breast cancer data set

| Results | Percentage accuracy % | |
|---|---|---|
| | Separate (433,250) | Leave one out (683) |
| TCMNN | | |
| Overall | 96.4 | 95.5 |
| Benign | 97.2 | 97.3 |
| Malignant | 94.4 | 92.1 |
| TCMSVM | | |
| Overall | 96.8 | 96.6 |
| Benign | 96.6 | 97.1 |
| Malignant | 92.1 | 95.8 |

Table A.9: **Results comparing performance of TCMNN and TCMSVM on various Wisconsin breast cancer data sets:** These TCMNN tests were carried out using the standard settings, analysing to 1 nearest neighbour and using the standard Euclidean distance. The TCMSVM tests were run using the standard inner product.

| Results | Percentage accuracy % | |
|---|---|---|
| | Inner product | Poly Deg 2 |
| TCMNN | | |
| Overall | 96.4 | 95.5 |
| Benign | 97.2 | 97.1 |
| Malignant | 94.4 | 92.5 |
| TCMSVM | | |
| Overall | 96.8 | 94.4 |
| Benign | 96.6 | 95.0 |
| Malignant | 97.2 | 93.0 |

Table A.10: **Results comparing performance of TCMNN and TCMSVM on various Wisconsin breast cancer data sets using polynomial kernels:** These TCMNN tests were carried out using the standard settings, analysing to 1 nearest neighbour and using the standard Euclidean distance. The results were obtained by training with WBC433 and tested with WBC250 data sets.

# Appendix B

# Results with 2-dimensional artificial data set

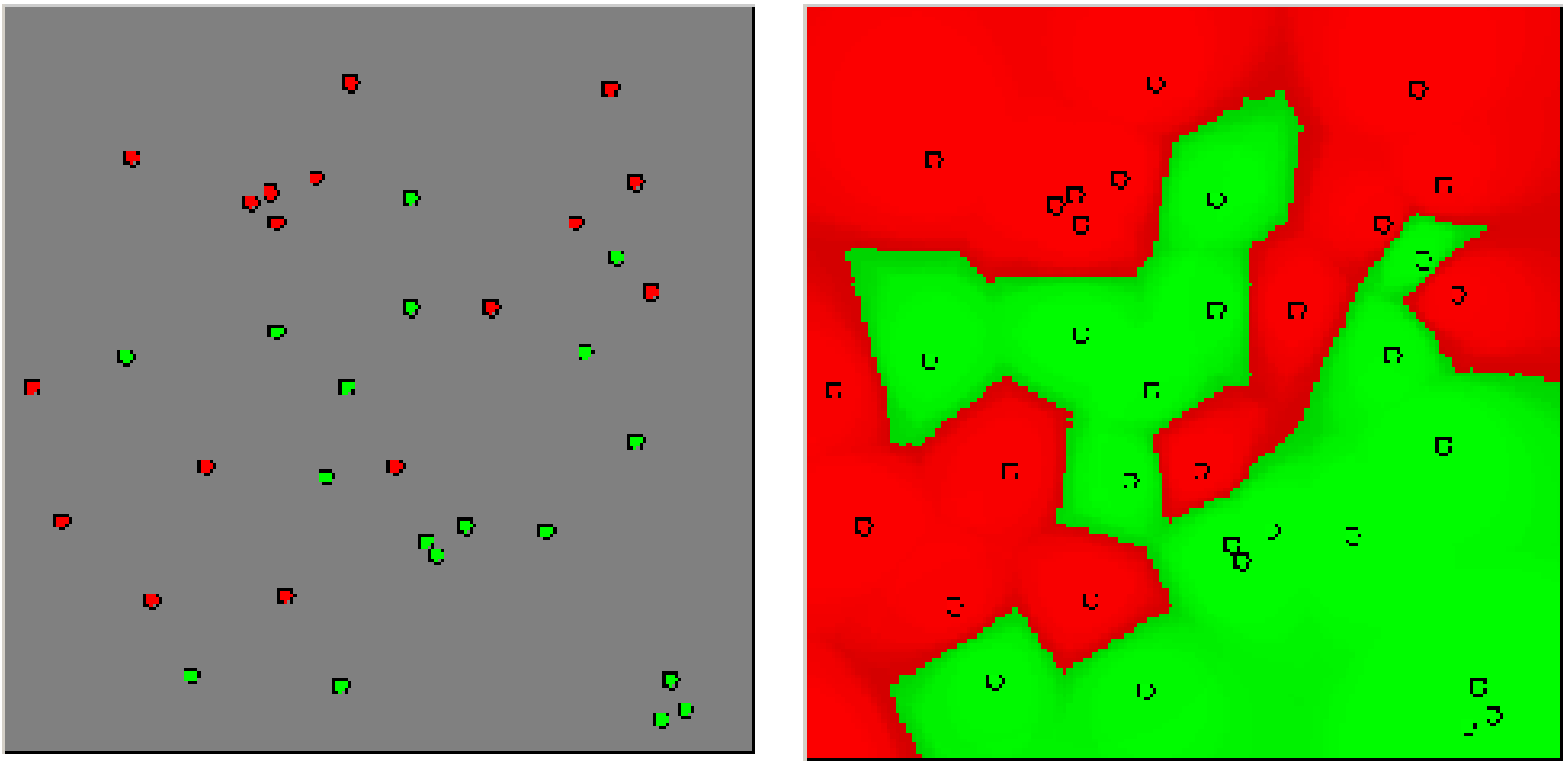

Figure B.1: **Example of the decision surface generated using the TCMNN algorithm with default settings on a simple artificial data set:** The plot on the left shows the training points class 0 = red, class 1 = green, specified before the use of the TCMNN. The plot on the right shows the predictions made by the TCMNN. The intensity of the colour shows how confident the prediction is. The closer to black the less confident it is. This same training data will be used in further experiments shown later.

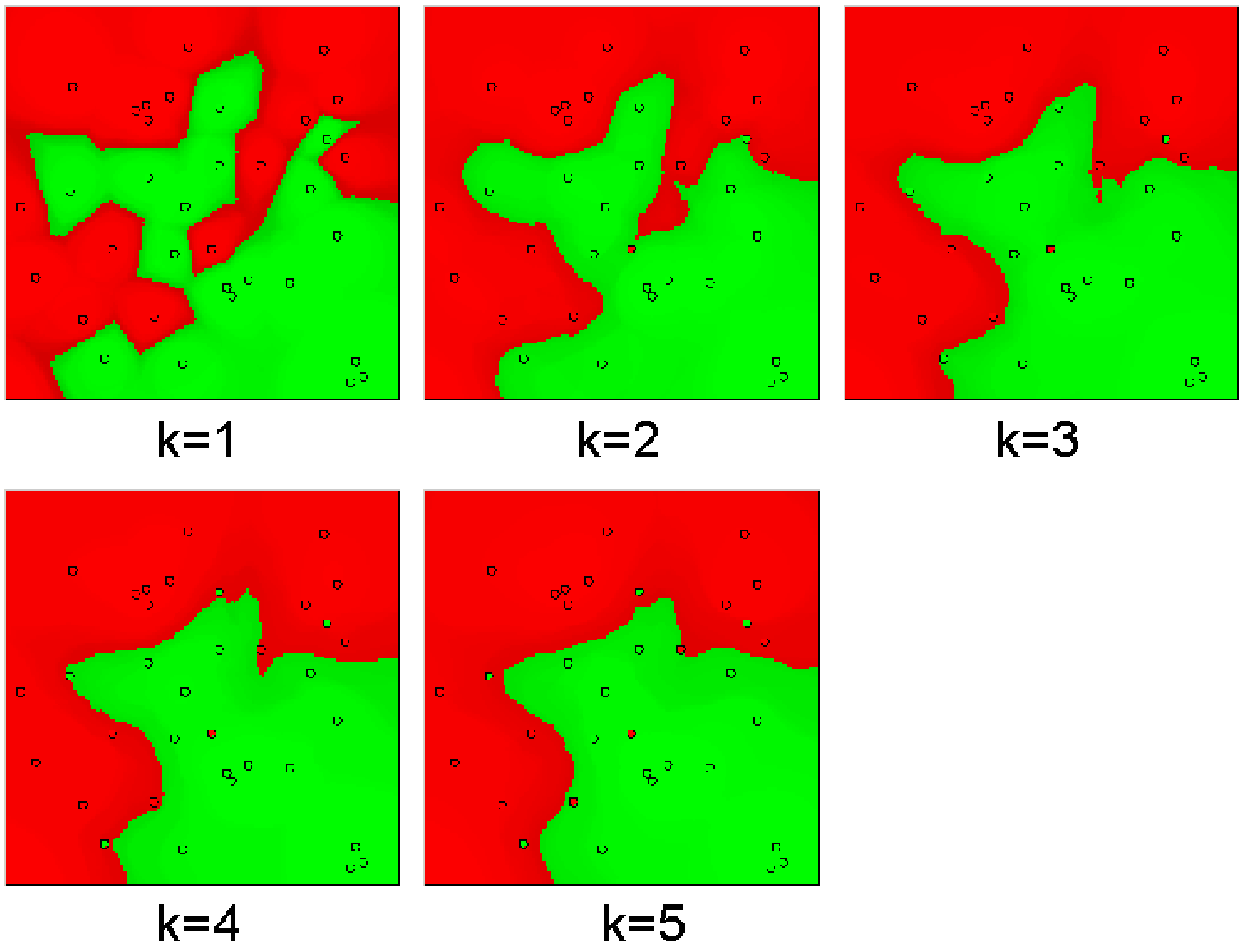

Figure B.2: Examples of the TCMNN decision surface for 2d artificial example analysing to different numbers of $k$ nearest neighbours

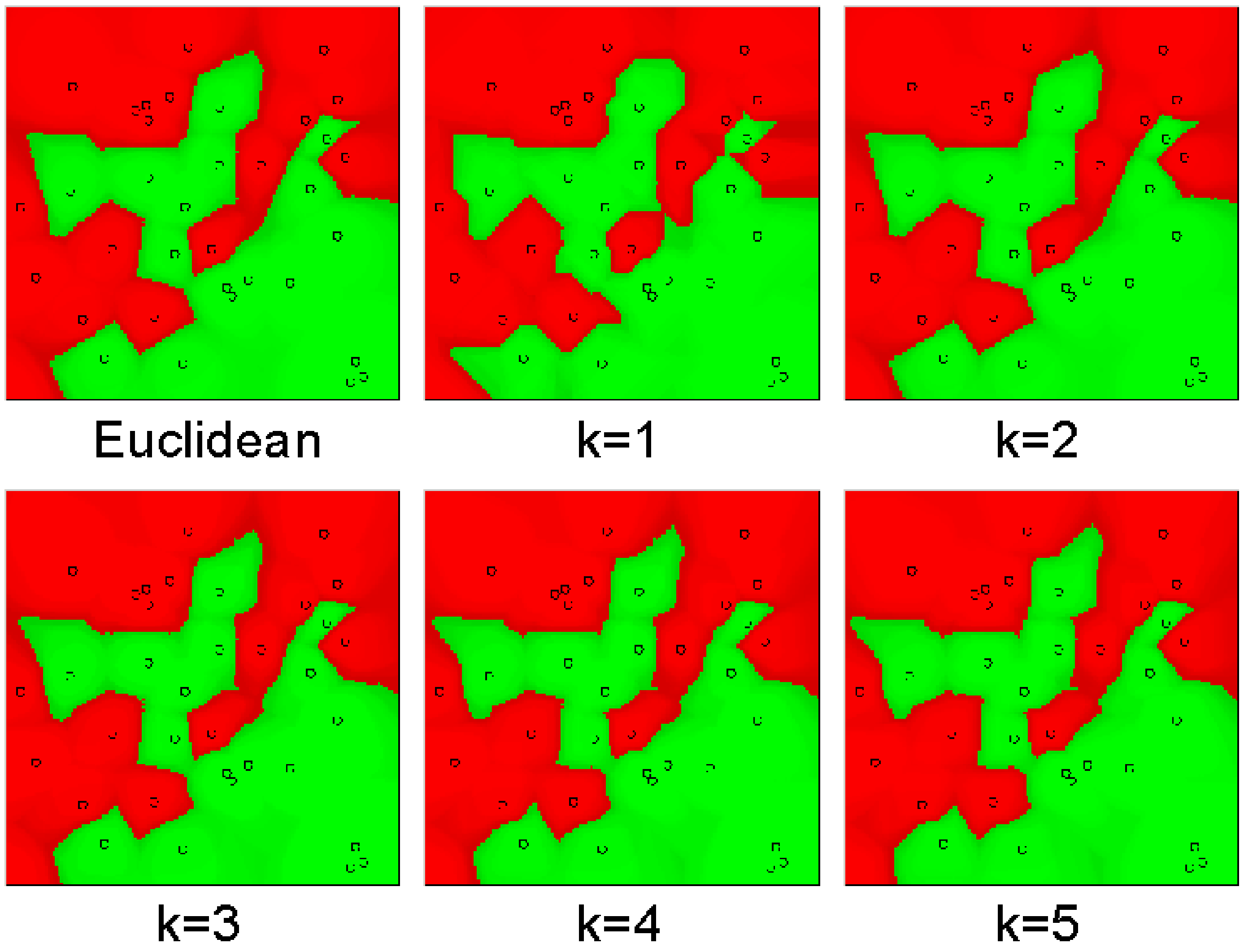

Figure B.3: Examples of the TCMNN decision surface for 2d artificial example analysing to different positive powers $k \geq 2$ of the Minkowski metric

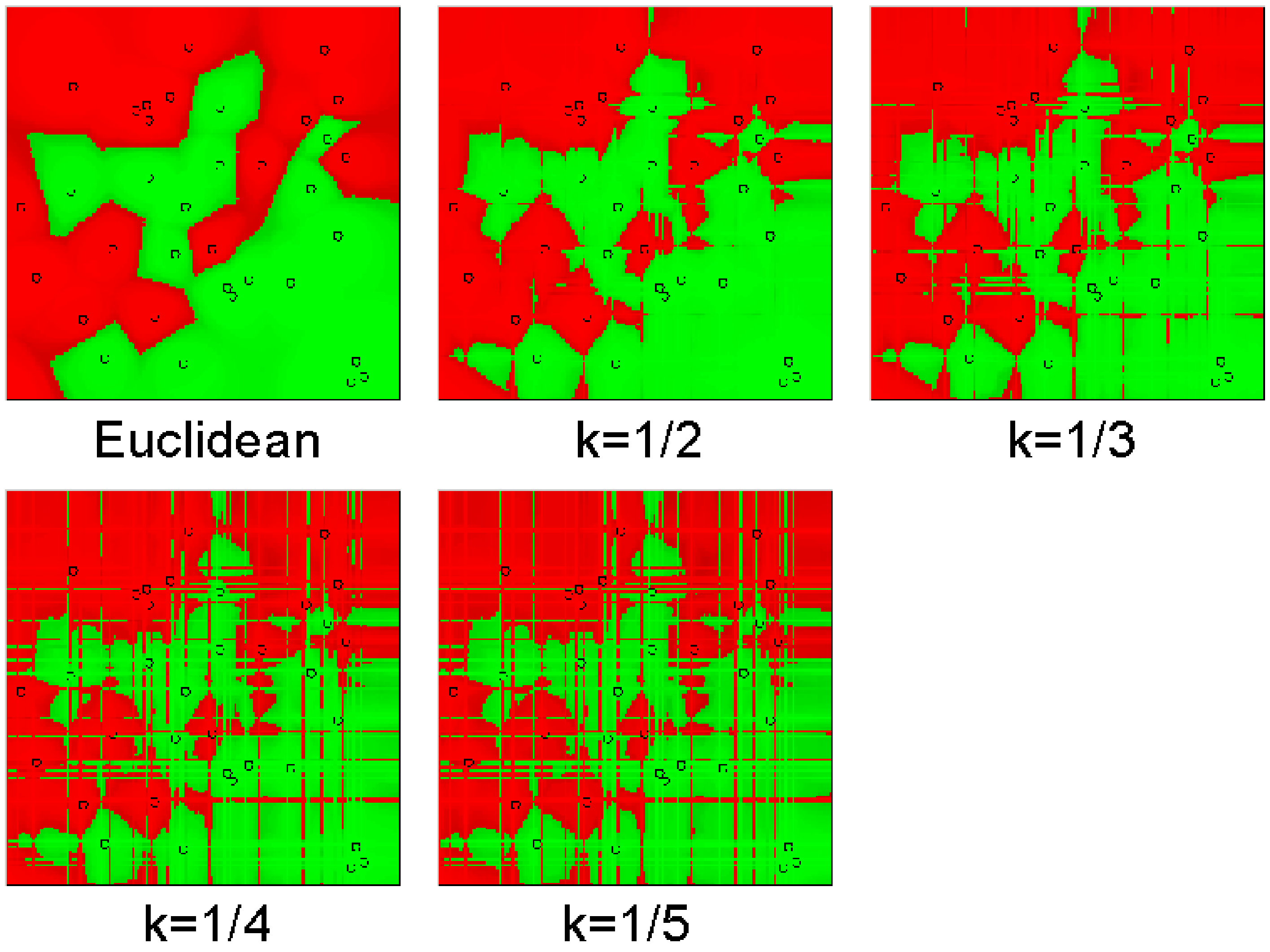

Figure B.4: Examples of the TCMNN decision surface for 2d artificial example analysing to different fractional positive powers $\frac{1}{k}$ of the Minkowski metric

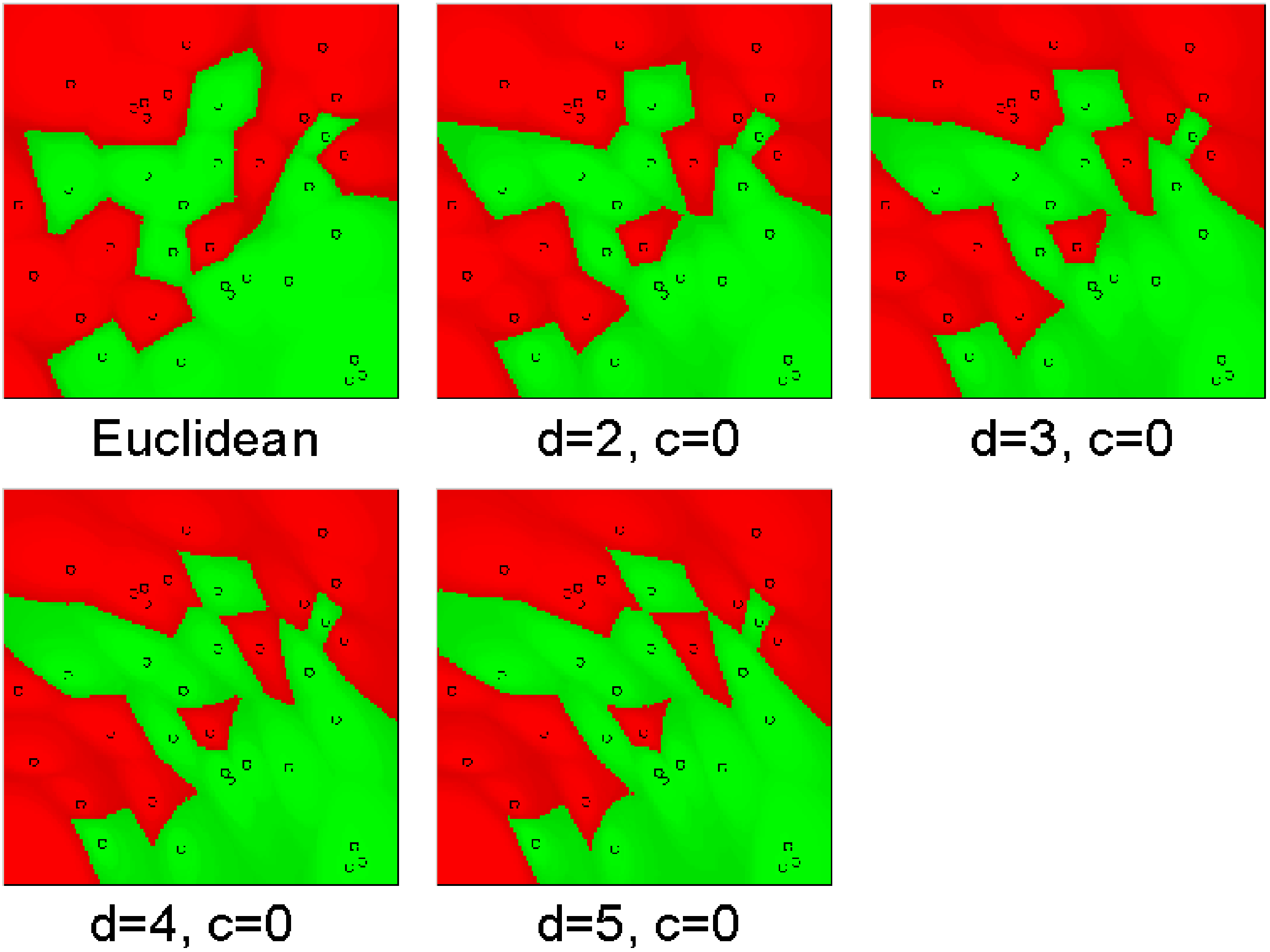

Figure B.5: Examples of the TCMNN decision surface for 2d artificial example using polynomial kernels with different degrees $d$, with constants $c = 0$

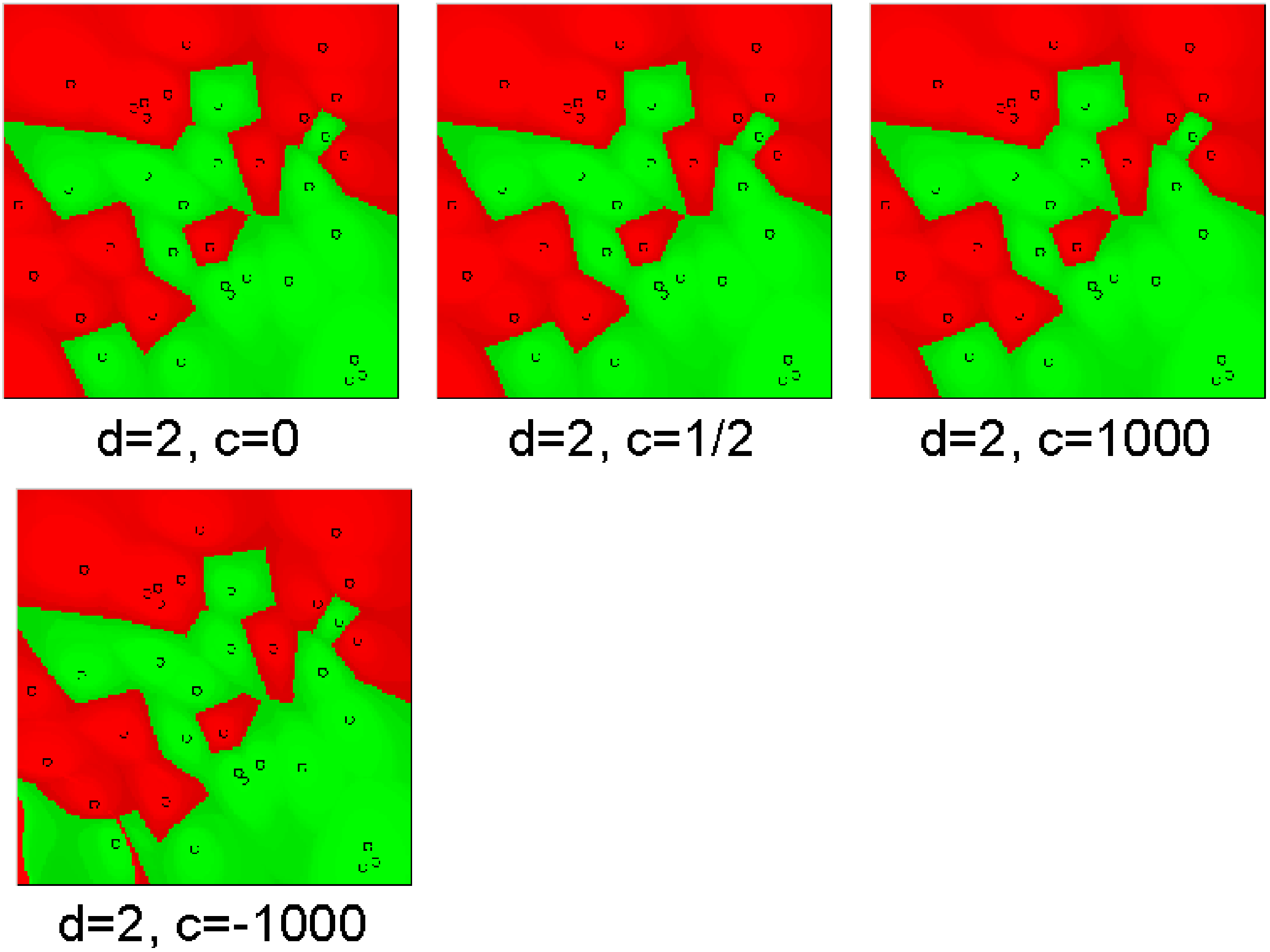

Figure B.6: Examples of the TCMNN decision surface for 2d artificial example using polynomial kernels with degrees $d = 2$, with constants $c$ set to various values

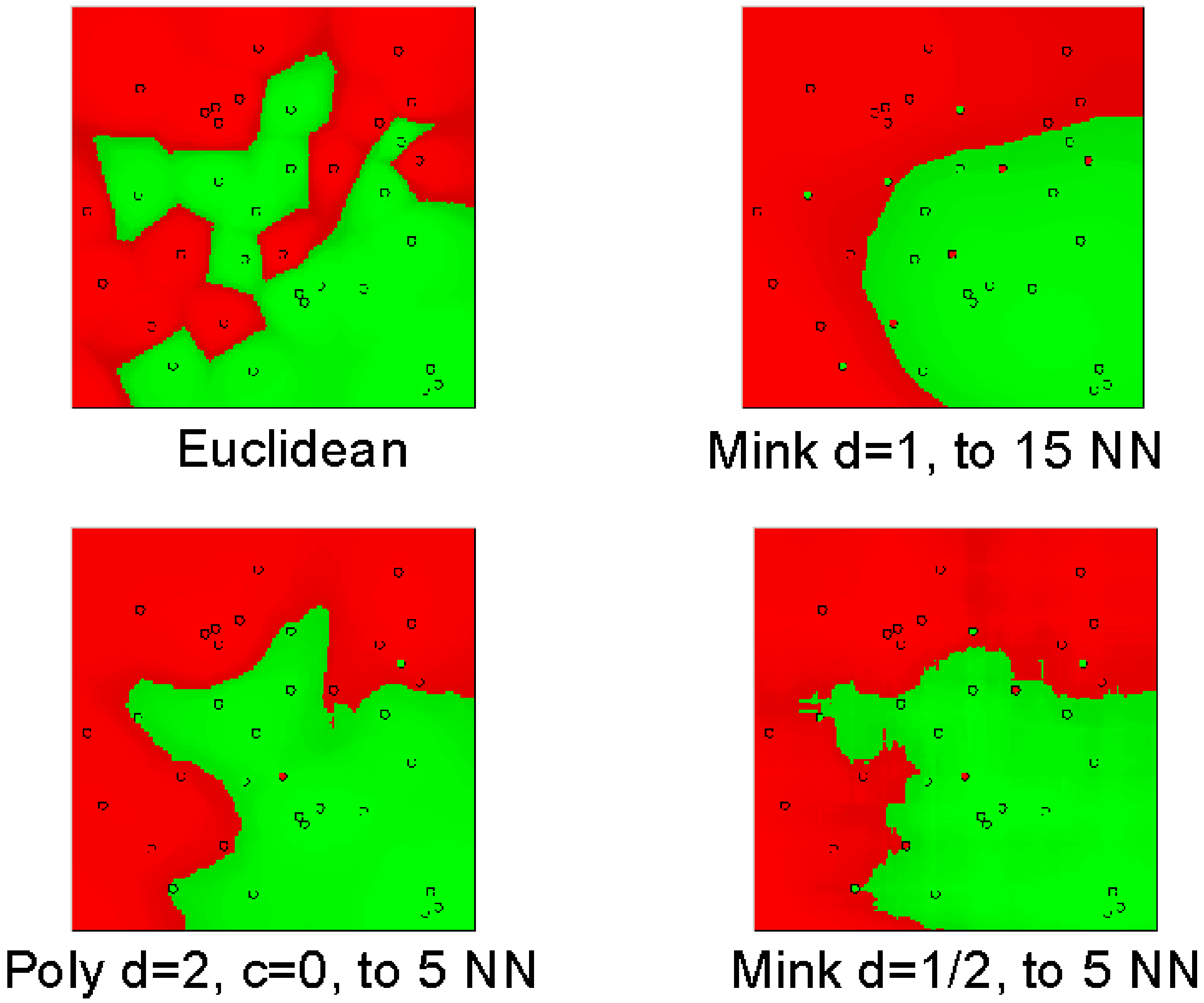

Figure B.7: Examples of the TCMNN decision surface for 2d artificial example with mixtures of settings

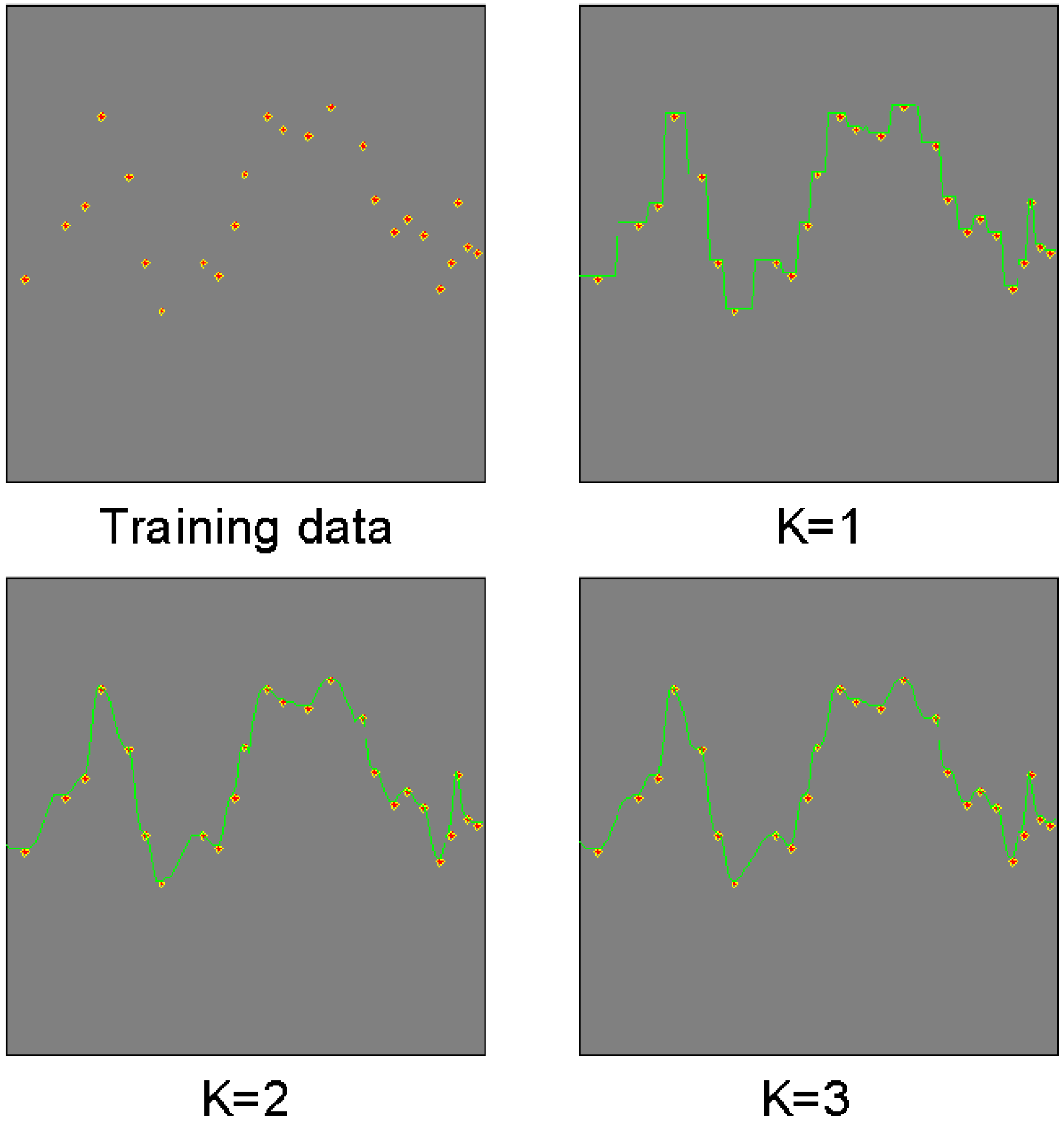

Figure B.8: Examples of DWKNN regression algorithm testing with a simple 2 dimensional artificial data set. These tests were run using various numbers of $k$ nearest neighbours.

# Appendix C

# Results with USPS data set

## C.1 Test 1 TCMNN - $k$ nearest setting

| | Percentage Accuracy % | | | | | |
|---|---|---|---|---|---|---|
| Neighbours | 1 | 10 | 20 | 30 | 40 | 50 |
| Overall | 94.9 | 94.8 | 93.7 | 93.4 | 92.6 | 91.7 |
| Zero | 98.6 | 98.9 | 98.6 | 98.3 | 98.3 | 98.1 |
| One | 97.0 | 97.7 | 97.7 | 97.7 | 97.7 | 97.7 |
| Two | 93.9 | 90.9 | 90.9 | 89.9 | 87.4 | 85.9 |
| Three | 90.7 | 92.8 | 90.4 | 90.4 | 90.4 | 89.8 |
| Four | 92.0 | 90.0 | 87.5 | 87.5 | 85.5 | 82.5 |
| Five | 91.9 | 91.9 | 88.1 | 86.9 | 84.4 | 83.1 |
| Six | 97.6 | 96.5 | 96.5 | 95.3 | 94.7 | 94.7 |
| Seven | 95.2 | 95.2 | 93.9 | 93.2 | 92.5 | 91.8 |
| Eight | 91.0 | 92.8 | 90.4 | 91.0 | 90.4 | 89.8 |
| Nine | 96.6 | 97.2 | 96.6 | 96.6 | 96.6 | 95.5 |
| Average Conf | 99.6 | 99.7 | 99.5 | 99.4 | 99.3 | 99.1 |
| Average Cred | 58.0 | 48.3 | 47.9 | 47.9 | 47.9 | 47.9 |

Table C.1: **Testing the USPS data set with TCMNN analysing to different numbers of $k$ nearest neighbours:** These TCMNN tests were carried out using the standard Euclidean distance. The TCMNN was trained with the USPS7291 and tested with the USPS2007 data sets.

## C.2 Test 2 TCMNN - Minkowski Metric

| Minkowski degree | Percentage Accuracy % | | | | | | | | |
|---|---|---|---|---|---|---|---|---|---|
| | $\frac{1}{5}$ | $\frac{1}{4}$ | $\frac{1}{3}$ | $\frac{1}{2}$ | 1 | 2 | 3 | 4 | 5 |
| Overall | 90.1 | 91.2 | 91.9 | 92.8 | 94.5 | 94.9 | 95.1 | 95.3 | 95.0 |
| Zero | 97.5 | 98.1 | 98.3 | 98.3 | 98.6 | 98.6 | 98.6 | 98.9 | 98.9 |
| One | 97.7 | 97.3 | 97.3 | 96.9 | 97.3 | 97.0 | 97.0 | 97.0 | 97.0 |
| Two | 87.4 | 89.4 | 90.4 | 91.4 | 93.4 | 93.9 | 93.4 | 92.4 | 91.9 |
| Three | 80.7 | 82.5 | 83.7 | 88.6 | 90.9 | 90.4 | 91.0 | 90.4 | 89.8 |
| Four | 84.5 | 86.5 | 87.0 | 87.5 | 90.5 | 92.0 | 92.5 | 92.5 | 92.0 |
| Five | 87.5 | 87.5 | 88.8 | 89.4 | 91.9 | 91.9 | 92.5 | 94.4 | 95.0 |
| Six | 94.7 | 95.3 | 95.3 | 95.9 | 95.9 | 97.6 | 97.6 | 97.1 | 96.5 |
| Seven | 93.2 | 93.2 | 93.2 | 92.5 | 93.9 | 95.2 | 94.6 | 95.2 | 94.6 |
| Eight | 80.1 | 83.1 | 84.9 | 86.7 | 90.4 | 90.9 | 91.6 | 92.8 | 92.2 |
| Nine | 87.0 | 89.3 | 90.4 | 92.7 | 96.0 | 96.6 | 97.7 | 97.7 | 97.7 |
| Average Conf | 98.5 | 98.7 | 98.9 | 99.2 | 99.5 | 99.6 | 99.6 | 99.6 | 99.6 |
| Average Cred | 55.9 | 56.2 | 56.5 | 57.1 | 58.1 | 58.0 | 57.6 | 57.3 | 57.0 |

Table C.2: **Testing the USPS data set with TCMNN using Minkowski metric with various degrees $d$ :** These TCMNN tests were carried out analysing to 1 nearest neighbour. The TCMNN was trained with the USPS7291 and tested with the USPS2007 data sets.

## C.3 Test 3 TCMNN - Kernel distance metric

| Minkowski degree | Percentage Accuracy % | | | | |
|---|---|---|---|---|---|
| | Euclid | Poly deg 2 | Poly deg 3 | Poly deg 4 | Poly deg 5 |
| Overall | 94.9 | 95.0 | 94.8 | 94.5 | 92.6 |
| Zero | 98.6 | 98.6 | 98.3 | 97.8 | 96.1 |
| One | 97.0 | 96.6 | 97.0 | 97.0 | 97.0 |
| Two | 93.9 | 93.9 | 93.9 | 91.9 | 89.4 |
| Three | 90.7 | 92.2 | 91.0 | 91.0 | 89.2 |
| Four | 92.0 | 90.5 | 90.0 | 89.5 | 84.5 |
| Five | 91.9 | 91.3 | 90.0 | 88.8 | 85.6 |
| Six | 97.6 | 97.6 | 97.1 | 96.5 | 93.5 |
| Seven | 95.2 | 95.2 | 93.9 | 93.9 | 92.5 |
| Eight | 91.0 | 92.2 | 94.0 | 95.8 | 96.4 |
| Nine | 96.6 | 97.2 | 97.7 | 98.3 | 97.2 |
| Average Conf | 99.6 | 99.6 | 99.6 | 99.5 | 99.2 |
| Average Cred | 58.0 | 57.6 | 57.1 | 56.3 | 55.4 |

Table C.3: **Testing the USPS data set with TCMNN using polynomial kernel feature mappings of various degrees** $d$**:** These TCMNN tests were carried out analysing to 1 nearest neighbour. The TCMNN was trained with the USPS7291 and tested with the USPS2007 data sets.

## C.4 Test 4 Significance level

| Minkowski degree | Percentage Accuracy % | | | | |
|---|---|---|---|---|---|
| | Euclid | $c$=0 | $c = 1/2$ | $c = 10$ | $c = 100$ |
| Overall | 94.9 | 95.0 | 95.0 | 95.1 | 95.0 |
| Zero | 98.6 | 98.6 | 98.6 | 98.6 | 98.6 |
| One | 97.0 | 96.6 | 96.6 | 96.6 | 96.6 |
| Two | 93.9 | 93.9 | 93.9 | 93.9 | 93.9 |
| Three | 90.7 | 92.2 | 92.2 | 92.2 | 92.2 |
| Four | 92.0 | 90.5 | 90.5 | 90.5 | 90.5 |
| Five | 91.9 | 91.3 | 91.3 | 91.9 | 91.3 |
| Six | 97.6 | 97.6 | 97.6 | 97.6 | 97.6 |
| Seven | 95.2 | 95.2 | 95.2 | 95.2 | 95.2 |
| Eight | 91.0 | 91.0 | 91.0 | 92.2 | 91.0 |
| Nine | 96.6 | 96.6 | 96.6 | 97.2 | 96.6 |
| Average Conf | 99.6 | 99.6 | 99.6 | 99.6 | 99.6 |
| Average Cred | 58.0 | 57.6 | 57.6 | 57.6 | 57.6 |

Table C.4: **Testing the USPS data set with TCMNN using polynomial degree 2 kernel feature mappings with various constants $c$:** These TCMNN tests were carried out analysing to 1 nearest neighbour. The TCMNN was trained with the USPS7291 and tested with the USPS2007 data sets.

| Significance | Percentage Accuracy % | | | | | | | | |
|---|---|---|---|---|---|---|---|---|---|
| | 10 | 20 | 30 | 40 | 50 | 60 | 70 | 80 | 90 |
| Overall | 98.5 | 99.3 | 99.5 | 99.6 | 99.7 | 99.8 | 99.9 | 99.8 | 99.8 |
| Zero | 99.4 | 100.0 | 100.0 | 100.0 | 100.0 | 100.0 | 100.0 | 100.0 | 100.0 |
| One | 98.0 | 98.0 | 98.8 | 98.8 | 99.6 | 99.6 | 100.0 | 100.0 | 100.0 |
| Two | 98.9 | 99.4 | 99.3 | 100.0 | 100.0 | 100.0 | 100.0 | 100.0 | 100.0 |
| Three | 97.9 | 99.2 | 100.0 | 100.0 | 100.0 | 100.0 | 100.0 | 100.0 | 100.0 |
| Four | 96.6 | 97.9 | 97.6 | 97.8 | 97.0 | 97.9 | 96.9 | 94.4 | 87.5 |
| Five | 97.0 | 98.1 | 98.8 | 98.2 | 100.0 | 100.0 | 100.0 | 100.0 | 100.0 |
| Six | 100.0 | 100.0 | 100.0 | 100.0 | 100.0 | 100.0 | 100.0 | 100.0 | 100.0 |
| Seven | 98.5 | 100.0 | 100.0 | 100.0 | 100.0 | 100.0 | 100.0 | 100.0 | 100.0 |
| Eight | 97.2 | 100.0 | 100.0 | 100.0 | 100.0 | 100.0 | 100.0 | 100.0 | 100.0 |
| Nine | 100.0 | 100.0 | 100.0 | 100.0 | 100.0 | 100.0 | 100.0 | 100.0 | 100.0 |
| Not Classed | 8.4 | 16.4 | 23.5 | 31.4 | 40.3 | 48.2 | 57.6 | 67.2 | 79.5 |

Table C.5: **Marking the USPS data set results with TCMNN to various significance levels :** These TCMNN tests were carried out analysing to 1 nearest neighbour, using the standard Euclidean distance. The TCMNN was trained with the USPS7291 and tested with the USPS2007 data sets.

# Appendix D

# Extended Results

This appendix will give more details about additional tests that were carried out.

## D.1 Ovarian cancer data set

| Test No. | Percentage accuracy % | | | | |
|---|---|---|---|---|---|
| | Overall | Benign | Malignant | Average Conf | Average Cred |
| 1 | 72.0 | 83.9 | 52.6 | 91.9 | 63.8 |
| 2 | 76.7 | 80.7 | 65.9 | 91.5 | 53.9 |
| 3 | 79.3 | 85.3 | 66.7 | 91.1 | 51.9 |
| 4 | 79.3 | 84.0 | 70.0 | 91.9 | 63.2 |
| 5 | 74.0 | 79.4 | 62.5 | 91.4 | 53.2 |
| 6 | 80.0 | 83.7 | 71.7 | 92.0 | 61.0 |
| 7 | 73.3 | 80.9 | 60.7 | 91.9 | 60.1 |
| 8 | 80.0 | 86.9 | 66.7 | 92.5 | 66.2 |
| 9 | 74.0 | 76.9 | 67.4 | 91.7 | 56.8 |
| 10 | 82.0 | 86.3 | 72.9 | 90.8 | 63.2 |
| Mean | 77.1 | 82.8 | 65.7 | 91.6 | 59.3 |

Table D.1: **Effect of TCMNN algorithm running random split tests with ovarian cancer data set:** These tests were run using the standard settings of analysing to 1 nearest neighbour, using the standard Euclidean distance. The accuracy of prediction fluctuates wildly between each test. Each test was run with a random split of 150 examples of the BARTS285 data set.

| | Percentage Accuracy % |
|---|---|

| Training Data | Overall | Benign | Malignant | Average Conf | Average Cred |
|---|---|---|---|---|---|
| TEST 1 | | | | | |
| Original | 82.0 | 90.0 | 70.0 | 93.9 | 66.9 |
| Neural Network | 84.0 | 96.7 | 65.0 | - | - |
| Augmented | 84.0 | 90.0 | 75.0 | 94.4 | 73.5 |
| TEST 2 | | | | | |
| Original | 82.0 | 80.5 | 85.7 | 90.1 | 58.5 |
| Neural Network | 94.0 | 97.2 | 85.7 | - | - |
| Augmented | 90.0 | 91.7 | 85.7 | 90.7 | 58.0 |
| TEST 3 | | | | | |
| Original | 72.0 | 84.6 | 58.3 | 93.4 | 60.5 |
| Neural Network | 74.0 | 96.2 | 50.0 | - | - |
| Augmented | 76.0 | 84.6 | 66.7 | 93.9 | 54.7 |
| TEST 4 | | | | | |
| Original | 84.0 | 82.8 | 85.7 | 91.4 | 66.6 |
| Neural Network | 90.0 | 93.1 | 85.7 | - | - |
| Augmented | 90.0 | 89.7 | 90.5 | 92.2 | 66.4 |
| TEST 5 | | | | | |
| Original | 68.0 | 78.1 | 50.0 | 92.6 | 53.8 |
| Neural Network | 80.0 | 100.0 | 44.4 | - | - |
| Augmented | 70.0 | 81.25 | 50.0 | 93.9 | 57.5 |
| TEST 6 | | | | | |
| Original | 76.0 | 85.7 | 53.3 | 94.0 | 55.7 |
| Neural Network | 70.0 | 68.6 | 73.3 | - | - |
| Augmented | 86.0 | 97.1 | 60.0 | 94.7 | 62.2 |
| TEST 7 | | | | | |
| Original | 84.0 | 87.1 | 78.9 | 89.9 | 53.8 |
| Neural Network | 88.0 | 100.0 | 68.4 | - | - |
| Augmented | 88.0 | 93.5 | 78.9 | 90.8 | 54.8 |
| TEST 8 | | | | | |
| Original | 76.0 | 85.7 | 63.6 | 91.7 | 52.3 |
| Neural Network | 80.0 | 92.9 | 63.7 | - | - |
| Augmented | 76.0 | 85.7 | 63.6 | 91.7 | 52.3 |
| TEST 9 | | | | | |
| Original | 86.0 | 83.9 | 89.5 | 91.8 | 60.7 |
| Neural Network | 88.0 | 93.5 | 78.9 | - | - |
| Augmented | 86.0 | 87.1 | 84.2 | 92.7 | 61.2 |
| TEST 10 | | | | | |
| Original | 82.0 | 96.7 | 60.0 | 93.4 | 61.8 |
| Neural Network | 84.0 | 96.7 | 65.0 | - | - |
| Augmented | 84.0 | 90.0 | 75.0 | 94.2 | 63.8 |

Table D.10: **Effect of using augmented ovarian cancer data sets with the TCMNN algorithm:** These were the results of testing the TCMNN algorithm on both the augmented and original versions of the ovarian cancer data sets. The tests were repeated so that a two tailed t-test could be performed to see how significant these differences in prediction were. Each test was run with a random split of 150 examples from the BARTS285 data set to ensure that the effect of the augmentation process was properly tested.

| K value | Percentage accuracy % | | | | |
|---|---|---|---|---|---|
| | Overall | Benign | Malignant | Average Conf | Average Cred |
| 1 | 80.7 | 82.5 | 77.8 | 92.8 | 57.3 |
| 10 | 82.8 | 94.4 | 63.9 | 95.1 | 54.7 |
| 20 | 83.2 | 94.9 | 63.9 | 95.1 | 54.8 |
| 30 | 83.5 | 95.4 | 63.9 | 94.3 | 55.5 |
| 40 | 83.9 | 96.0 | 63.9 | 93.9 | 55.9 |
| 50 | 83.9 | 97.2 | 62.0 | 93.6 | 56.3 |

Table D.2: **Results of TCMNN leave one out testing the ovarian cancer data set with different numbers of nearest neighbours:** These tests were carried out with a leave one out test on the BARTS285 data set. The TCMNN algorithm was performed using the standard Euclidean distance metric.

| Test Set | Percentage accuracy % | | |
|---|---|---|---|
| | Overall | Benign | Malignant |
| BARTS146 | 86.9 | 92.4 | 80.6 |
| BARTS285 | 90.2 | 95.5 | 81.5 |
| BARTS394 | 87.3 | 89.1 | 83.8 |

Table D.3: **Results of testing the various ovarian cancer data sets using the RMI index screening technique** These tests were carried out using the same ovarian cancer data sets that were used in my other experiments.

## D.2 Abdominal pain data set

| Training | Percentage Accuracy % | | |
|---|---|---|---|
| | Overall | Average Conf | Average Cred |
| TEST 1 | | | |
| Original | 59.7 | 77.6 | 71.7 |
| Neural Network | 64.0 | - | - |
| Augmented | 58.0 | 82.1 | 67.2 |
| TEST 2 | | | |
| Original | 58.5 | 78.2 | 72.2 |
| Neural Network | 65.0 | - | - |
| Augmented | 58.3 | 83.4 | 66.9 |
| TEST 3 | | | |
| Original | 57.7 | 76.5 | 69.7 |
| Neural Network | 65.8 | - | - |
| Augmented | 59.0 | 82.6 | 67.8 |
| TEST 4 | | | |
| Original | 55.9 | 76.9 | 69.8 |
| Neural Network | 63.0 | - | - |
| Augmented | 58.6 | 83.9 | 65.7 |
| TEST 5 | | | |
| Original | 57.6 | 77.6 | 72.6 |
| Neural Network | 67.7 | - | - |
| Augmented | 61.0 | 82.4 | 69.6 |
| TEST 6 | | | |
| Original | 55.9 | 77.9 | 71.7 |
| Neural Network | 64.3 | - | - |
| Augmented | 61.8 | 83.0 | 67.8 |
| TEST 7 | | | |
| Original | 57.1 | 77.2 | 72.3 |
| Neural Network | 64.6 | - | - |
| Augmented | 60.7 | 82.1 | 68.1 |

Table D.16: **Effect of using augmented abdominal pain data sets with the TCMNN algorithm:** These were the results of testing the TCMNN algorithm on both the augmented and original versions of the abdominal pain data sets. The tests were repeated so that a two tailed t-test could be performed to see how significant these differences in prediction were. Each test was run with a random splits of 1000 examples of the ABDO4387 data set to ensure that the effect of the augmentation process was properly tested.

| Test No. | Percentage accuracy % | | |
|---|---|---|---|
| | Overall | Benign | Malignant |
| 1 | 86.9 | 98.7 | 73.1 |
| 2 | 88.4 | 98.7 | 76.1 |
| 3 | 89.7 | 98.7 | 79.1 |
| 4 | 84.9 | 98.7 | 68.7 |
| 5 | 87.7 | 98.7 | 74.6 |
| Mean | 87.5 | 98.7 | 74.3 |

Table D.4: **Testing neural network performance with equal weighting on benign and malignant cases for ovarian cancer data set :** These tests were run using a 3-5-2 fully interconnected network. The output used a 1 of N output encoding, and equal bias was put on malignant and benign cases. The learning rate was set to 0.1 and the weights were initialised randomly between ±0.05. The network was trained for 100,000 weight updates on the 139 example data set and tested on the 146 example data set. Several tests were made because the random nature of the weight intialisation yields slightly different results each time.

| Test No. | Percentage accuracy % | | |
|---|---|---|---|
| | Overall | Benign | Malignant |
| 1 | 89.7 | 98.7 | 79.1 |
| 2 | 74.6 | 62.0 | 89.6 |
| 3 | 85.6 | 86.1 | 85.1 |
| 4 | 87.7 | 98.7 | 74.6 |
| 5 | 87.7 | 93.7 | 80.6 |
| Mean | 85.1 | 87.8 | 81.8 |

Table D.5: **Testing neural network performance with bias weighting on benign cases setting target to 0.8:** The setup for these tests were exactly the same as that detailed in Table D.4. The only difference between these tests is that the target values for benign cases were set to 0.8 instead of 0.999.

## D.3 Wisconsin breast cancer data set

| Training Data | Percentage Accuracy % | | | | |
|---|---|---|---|---|---|
| | Overall | Benign | Malignant | Average Conf | Average Cred |
| TEST 1 | | | | | |
| Original | 96.4 | 98.1 | 93.3 | 99.4 | 61.8 |
| Neural Network | 97.6 | 96.9 | 98.9 | - | - |
| Augmented | 96.8 | 96.3 | 97.8 | 99.5 | 63.4 |
| TEST 2 | | | | | |
| Original | 95.6 | 96.4 | 93.9 | 99.5 | 65.2 |
| Neural Network | 96.0 | 94.6 | 98.8 | - | - |
| Augmented | 95.2 | 95.2 | 95.1 | 99.6 | 66.1 |
| TEST 3 | | | | | |
| Original | 95.6 | 97.5 | 92.4 | 99.5 | 64.1 |
| Neural Network | 96.8 | 96.8 | 96.7 | - | - |
| Augmented | 96.0 | 96.8 | 94.6 | 99.4 | 64.5 |
| TEST 4 | | | | | |
| Original | 94.8 | 95.8 | 92.7 | 99.5 | 64.3 |
| Neural Network | 96.8 | 95.8 | 98.8 | - | - |
| Augmented | 96.0 | 96.4 | 95.1 | 99.6 | 65.7 |
| TEST 5 | | | | | |
| Original | 97.2 | 98.1 | 95.5 | 99.4 | 59.9 |
| Neural Network | 97.6 | 98.1 | 96.6 | - | - |
| Augmented | 96.8 | 98.1 | 94.3 | 99.6 | 62.9 |
| TEST 6 | | | | | |
| Original | 95.2 | 98.2 | 88.9 | 99.4 | 64.1 |
| Neural Network | 97.2 | 97.6 | 96.3 | - | - |
| Augmented | 94.8 | 97.6 | 88.9 | 99.5 | 66.4 |
| TEST 7 | | | | | |
| Original | 97.2 | 97.1 | 88.9 | 99.5 | 64.2 |
| Neural Network | 98.3 | 98.1 | 89.3 | - | - |
| Augmented | 98.2 | 98.1 | 88.9 | 99.7 | 64.1 |

Table D.20: **Effect of using augmented Wisconsin breast cancer data sets with the TCMNN algorithm:** These were the results of testing the TCMNN algorithm on both the augmented and original versions of the Wisconsin breast cancer data sets. The tests were repeated so that a two tailed t-test could be performed to see how significant these differences in prediction were. Each test was run with a random splits of 250 examples of the WBC683 data set to ensure that the effect of the augmentation process was properly tested.

| Test No. | Percentage accuracy % | | |
|---|---|---|---|
| | Overall | Benign | Malignant |
| 1 | 88.4 | 93.7 | 82.1 |
| 2 | 85.6 | 86.1 | 85.1 |
| 3 | 88.4 | 98.7 | 76.1 |
| 4 | 86.9 | 93.7 | 79.1 |
| 5 | 88.4 | 93.7 | 82.1 |
| Mean | 87.5 | 93.2 | 80.9 |

Table D.6: **Testing neural network performance on the ovarian cancer data set with bias weighting on benign cases setting target to 0.7:** The setup for these tests were exactly the same as that detailed in Table D.4. The only difference between these tests is that the target values for benign cases were set to 0.7 instead of 0.999.

| Test No. | Percentage accuracy % | | |
|---|---|---|---|
| | Overall | Benign | Malignant |
| 1 | 87.7 | 93.7 | 80.6 |
| 2 | 88.4 | 93.7 | 82.1 |
| 3 | 88.4 | 92.4 | 83.6 |
| 4 | 87.7 | 93.7 | 80.6 |
| 5 | 87.7 | 93.7 | 80.6 |
| Mean | 87.9 | 93.4 | 81.5 |

Table D.7: **Testing neural network performance on the ovarian cancer data set with bias weighting on benign cases setting target to 0.6:** The setup for these tests were exactly the same as that detailed in Table D.4. The only difference between these tests is that the target values for benign cases were set to 0.6 instead of 0.999.

| Test No. | Percentage accuracy % | | |
|---|---|---|---|
| | Overall | Benign | Malignant |
| 1 | 75.3 | 64.6 | 88.1 |
| 2 | 88.4 | 93.7 | 82.1 |
| 3 | 74.7 | 62.0 | 89.6 |
| 4 | 88.4 | 92.4 | 83.6 |
| 5 | 84.9 | 93.7 | 74.6 |
| Mean | 82.3 | 81.3 | 83.6 |

Table D.8: **Testing neural network performance on the ovarian cancer data set with bias weighting on benign cases setting target to 0.5:** The setup for these tests were exactly the same as that detailed in Table D.4. The only difference between these tests is that the target values for benign cases were set to 0.5 instead of 0.999.

| Results | Percentage accuracy % | | |
|---|---|---|---|
| | Separate (139,146) | Leave one out (285) | Leave one out (394) |
| Overall | 78.8 | 81.1 | 79.2 |
| Benign | 75.9 | 82.5 | 84.1 |
| Malignant | 82.1 | 78.7 | 69.9 |

Table D.11: **Initial KNN test accuracy results using ovarian cancer data sets of different sizes:** These KNN tests were carried out using the standard settings, analysing to 1 nearest neighbour and using the standard Euclidean distance.

| K value | Percentage accuracy % | | |
|---|---|---|---|
| | Overall | Benign | Malignant |
| 1 | 78.8 | 75.9 | 82.1 |
| 10 | 79.4 | 86.1 | 71.6 |
| 20 | 82.2 | 92.4 | 70.1 |
| 30 | 81.5 | 94.9 | 65.7 |
| 40 | 79.5 | 97.5 | 58.2 |
| 50 | 76.0 | 97.5 | 50.7 |

Table D.12: **Results of KNN testing the ovarian cancer data set with different numbers of nearest neighbours:** These tests were carried out with the separate training BARTS139, testing BARTS146 data sets described earlier. The KNN algorithm was performed using the standard Euclidean distance metric.

| Results | Percentage accuracy % | | |
|---|---|---|---|
| | Separate (139,146) | Leave one out (285) | Leave one out (394) |
| Overall | 78.8 | 81.1 | 79.2 |
| Benign | 75.9 | 82.5 | 84.1 |
| Malignant | 82.1 | 78.7 | 69.9 |

Table D.13: **Initial DWKNN test results using ovarian cancer data sets of different sizes:** These DWKNN tests were carried out using the standard settings, analysing to 1 nearest neighbour and using the standard Euclidean distance.

| K value | Percentage accuracy % | | |
|---|---|---|---|
| | Overall | Benign | Malignant |
| 1 | 78.8 | 75.9 | 82.1 |
| 10 | 78.8 | 78.5 | 79.1 |
| 20 | 78.8 | 78.5 | 79.1 |
| 30 | 79.5 | 79.7 | 79.1 |
| 40 | 79.5 | 79.7 | 79.1 |
| 50 | 79.5 | 79.7 | 79.1 |

Table D.14: **Results of DWKNN testing the ovarian cancer data set with different numbers of nearest neighbours:** These tests were carried out with the separate training BARTS139, testing BARTS146 data sets described earlier. The DWKNN algorithm was performed using the standard Euclidean distance metric.

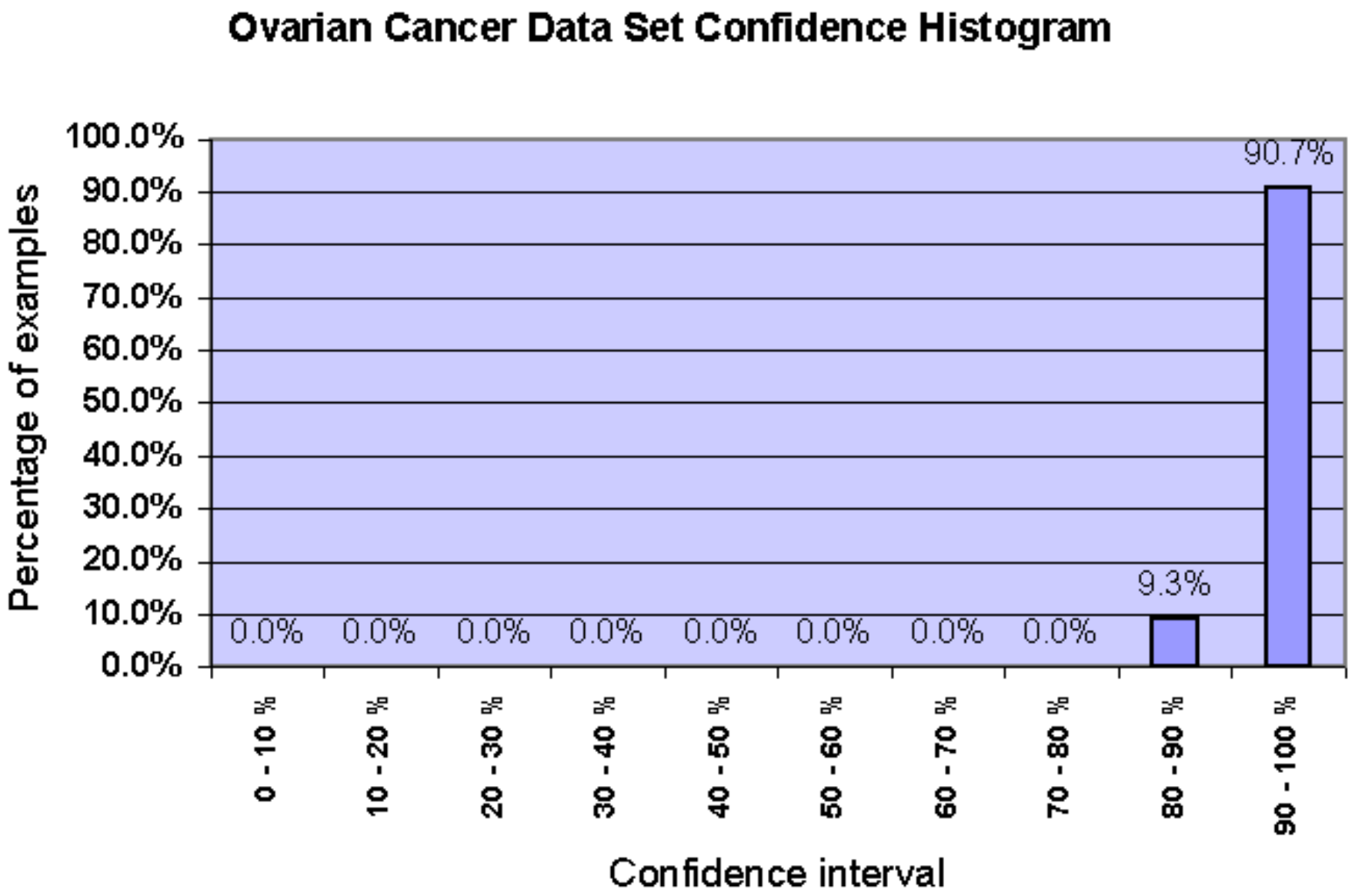

Figure D.1: This is a histogram of the percentage of examples that lie between fixed confidence level intervals. This is the results we would hope for in that most examples clearly have high confidence levels, indicating there is a clear distinction between most examples

| Class | Test 1 | Test 2 | Test 3 | Test 4 | Test 5 |
|---|---|---|---|---|---|
| Overall | 67.5 | 66.1 | 67.4 | 66.6 | 66.6 |
| Appendicitis | 74.1 | 67.2 | 62.5 | 59.1 | 44.8 |
| Diverticulitis | 0.0 | 0.0 | 0.0 | 0.0 | 0.0 |
| Perf peptic ulcer | 0.0 | 0.0 | 0.0 | 0.0 | 0.0 |
| Non-spec abdom pain | 78.1 | 83.3 | 78.3 | 87.0 | 90.9 |
| Cholisistitis | 71.0 | 79.0 | 75.5 | 64.5 | 57.5 |
| Intestinal obstr | 0.0 | 44.8 | 0.0 | 0.0 | 0.0 |
| Pancreatitis | 0.0 | 0.0 | 0.0 | 0.0 | 0.0 |
| Renal colic | 74.1 | 0.0 | 76.2 | 62.6 | 57.1 |
| Dyspepsia | 78.4 | 71.3 | 83.8 | 67.5 | 77.4 |

Table D.15: **Repeated tests using Neural networks on the abdominal pain data set:** The results above were obtained using a 135-10-10-9 feed forward network. The network was trained for 100,000 weight updates using the 4387 example data set, and tested using the 2000 example data set.

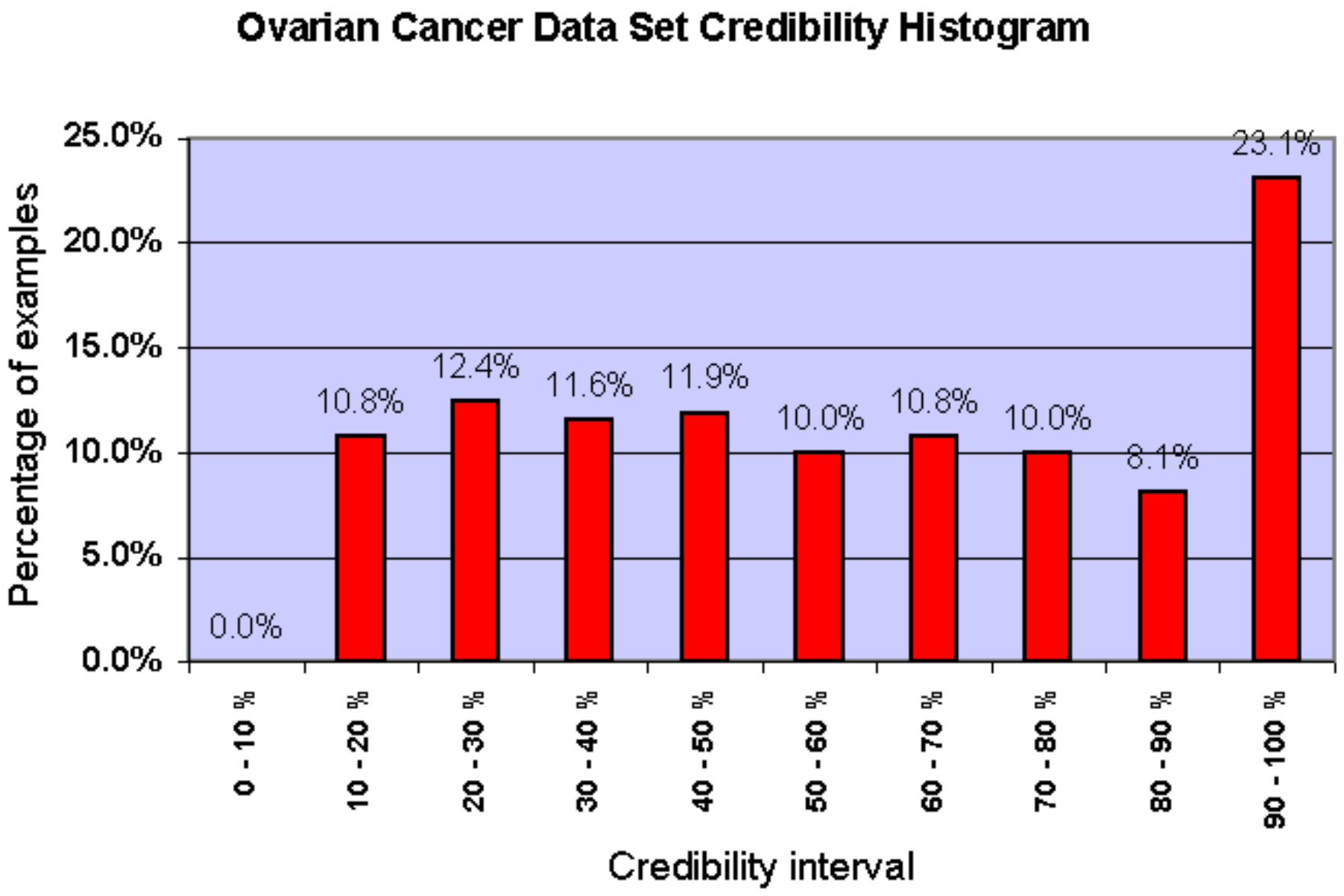

Figure D.2: This is a histogram of the percentage of examples that lie between fixed credibility level intervals. Credibility for this data set is very poor.

| Test No. | Percentage accuracy % | | |
|---|---|---|---|
| | Overall | Benign | Malignant |
| 1 | 98.0 | 97.8 | 98.6 |
| 2 | 97.6 | 97.2 | 98.6 |
| 3 | 97.2 | 96.6 | 98.6 |
| 4 | 97.2 | 96.6 | 98.6 |
| 5 | 98.0 | 97.8 | 98.6 |
| Mean | 97.6 | 97.2 | 98.6 |

Table D.17: **Testing neural network performance with equal weighting on benign and malignant cases in the Wisconsin breast cancer data set:** These tests were run using a 9-10-10-2 fully interconnected network. The output used a 1 of N output encoding, and equal bias was put on malignant and benign cases. The learning rate was set to 0.1 and the weights were initialised randomly between ±0.05. The network was trained for 100,000 weight updates on the WBC433 data set and tested on the WBC250 data set. Several tests were made because the random nature of the weight intialisation yields slightly different results each time.

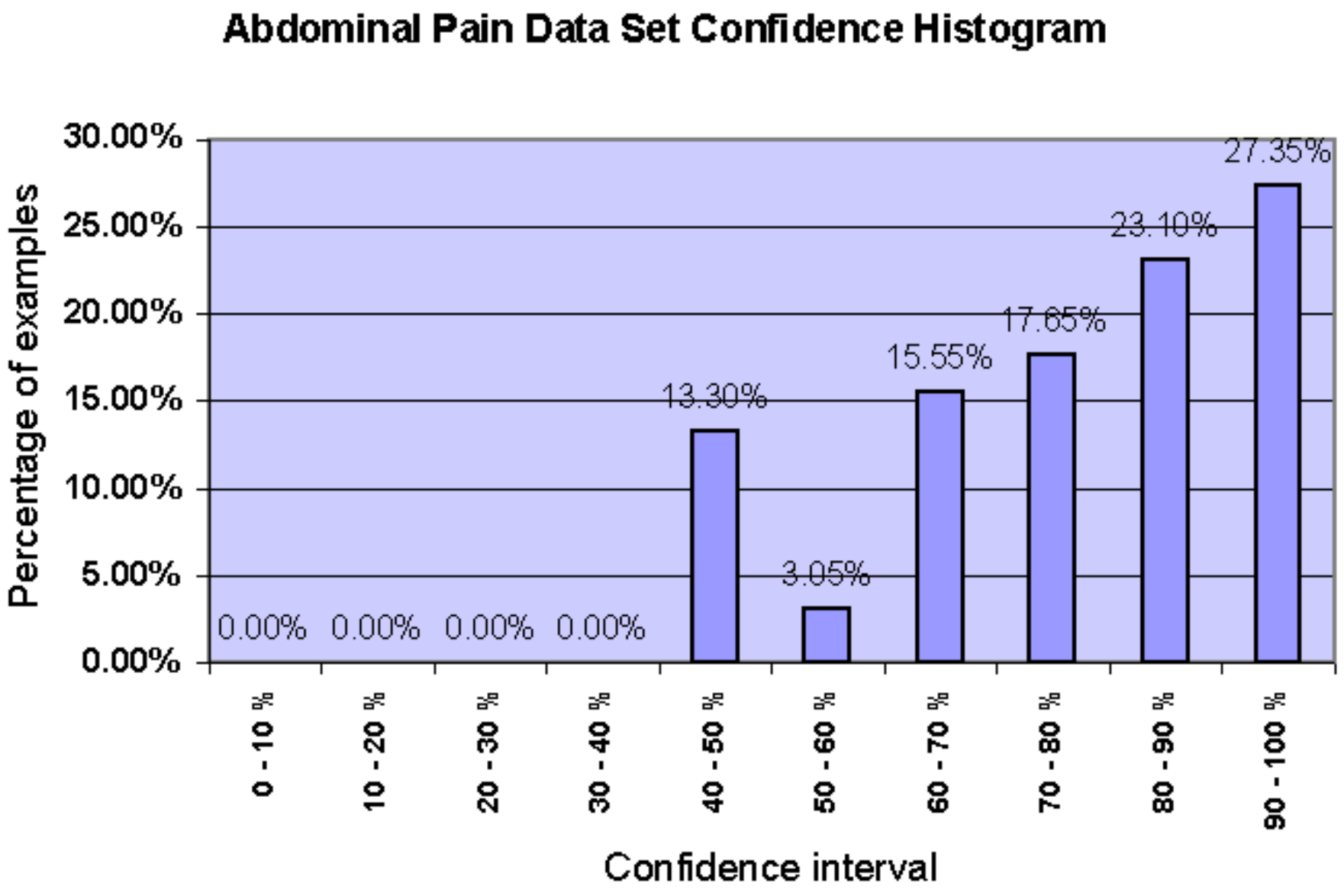

Figure D.3: This is a histogram of the percentage of examples that lie between fixed confidence level intervals. Ideally we would hope that the majority of examples would have high confidence readings.

| Significance level | Percentage accuracy % | | | |
|---|---|---|---|---|
| | Overall | Benign | Malignant | Not Classified |
| 5% | 96.3 | 97.5 | 93.9 | 1.2 |
| 10% | 98.6 | 98.6 | 98.4 | 8.4 |
| 15% | 99.2 | 99.3 | 98.8 | 13.5 |
| 20% | 99.5 | 99.8 | 98.5 | 18.8 |
| 25% | 99.8 | 99.8 | 100 | 24.2 |
| 30% | 100.0 | 100.0 | 100.0 | 28.9 |

Table D.18: **Results of TCMNN leave one out testing the Wisconsin breast cancer data set using different significance levels:** These tests were carried out with a leave one out test on the WBC683 data set described earlier. The TCMNN algorithm was performed analysing to 1 nearest neighbour, using the standard Euclidean distance.

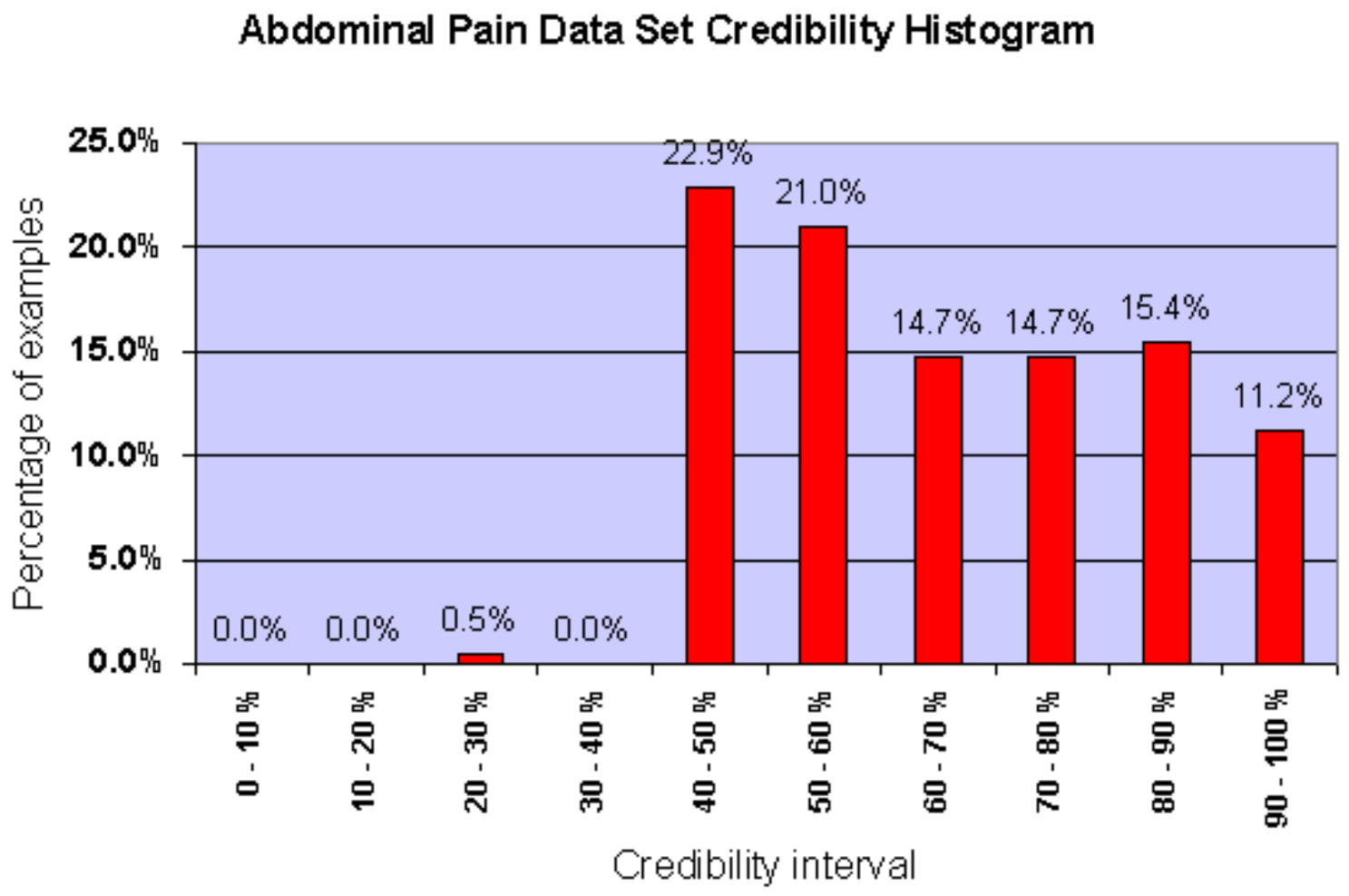

Figure D.4: This is a histogram of the percentage of examples that lie between fixed credibility level intervals. The ideal outcome would be for the majority of examples to have high credibility

# Appendix E

# Source Code

The source code for all the applications developed, in addition to the data files and results of the tests mentioned in the experiments chapters of this project, can be found at my web site:
`http://www.rhul.ac.uk/users/zjac048/3rdyearproject/`

The source code for the core component classes (ie TCMNN, Neural Networks, Image processing as shown in Figure 3.1 on page 38) can be found in the `classes` subdirectory.

The following data files that were used in the experiments can be found at the base of the `3rdyearproject` directory:

- Abdominal pain - `abdo2000.data`, `abdo4386.data`
- USPS - `usps2007.data`, `usps7291.data`
- Wisconsin breast cancer - `WBC_683.data`, `WBC_250.data`, `WBC_433.data`
- Ovarian cancer - `BARTS_394.data`, `BARTS_285.data`, `BARTS_139.data`, `BARTS_146.data`
- David/Sian artificial image data set - `DAVE_SIAN_IMAGES.data`

The test harness programs were used to run the tests shown in the experiments chapters. The test harness programs can be found for each data set (where `XXXX` is the name of the data set) in the base of the `3rdyearproject` directory:

- Neural network tests - `NNET_XXXX_test.java`
- Neural network augmented data set tests - `NNet_hidden_XXXX_test.java`
- TCMNN tests - `TCMNN_XXXX_test.java`
- (DW)KNN tests - `(DW)KNN_XXXX_test.java`

The results of the experiments are organised in the subdirectory to each data sets name.

The main applications developed for the project can be found in the base of the `3rdyearproject` directory:

- TCMNN testing applications - `TCMNN_LeaveOneOut_Application.java`, `TCMNN_Separate _Application.java`, `TCMNN_Manual_Application.java`, `TCMNN_Image_LeaveOneOut_Application.java`
- Data file creator applications - `DF_Creator.java`, `Image_Data_File_Creator.java`
- 2d TCMNN pattern classification application - `TCMNN_2d_Example.java`
- 2d KNN regression application - `NNRegressionExample.java`

An electronic version of this document can be found at:
`http://www.rhul.ac.uk/users/zjac048/3rdyearproject/dissertation.pdf`

# Appendix F

# Example outputs from programs

This appendix provides examples of some of the output of the TCMNN testing applications developed for my project.

The HTML output files (shown on the next couple of pages) are accessible via my web site:
`http://www.rhul.ac.uk/users/zjac048/3rdyearproject/`
and the file names are as follows:

- `BARTS_394_lres.html`
- `BARTS_394_lstat.html`
- `DAVE_SIAN_IMAGES_lres.html`
- `DAVE_SIAN_IMAGES_lstat.html`

# Appendix G

# User Guide

This appendix will give a basic overview of how to use the systems developed in this project.

## G.1 Overview of the TCMNN testing system

The TCMNN testing system is a learning system that predicts the classification of unseen, new testing data presented to it. The system is able to make its predictions based on training data provided to it, where the training data contains examples of data whose classifications are already known. In addition to providing classification predictions, the TCMNN testing system can also provide probability measures of how confident and credible the predications are.

Before any data can be analysed using the TCMNN testing system, it must first go through a series of preparative steps via the Data File Creator (DFC). The DFC is an application which can convert the data to be analysed from a tab delimited text file into a .data format which is recognisable by the TCMNN testing system. The .data file is created by the DFC by querying the user for specific details about the data, such as class names, number of classes, attribute names and so on.

The TCMNN testing system is then able to use the data in the .data format and perform four different types of tests on it. Four applications comprise these tests, and they are as follows:

1. Leave One Out Test (LOOT)
2. Separate Test (ST)
3. Manual Test (MT)
4. Batch Test (BT)

The LOOT requires one data file for use. This application will run a test on one selected data example in the data file and then cross-validate this example with the remaining examples in the data file. This test is therefore able to maximise the number of training examples which can be used for classification.

The ST is a simple application which requires two data files for use. One data file is used for training; the other is used for testing. It must be noted that the classes of the testing examples for the LOOT and ST applications must be known in order for the performances of the tests to be properly assessed.

The MT requires one data file. This data file is used for training and then the user can manually enter new examples for testing. The MT application is used in situations where the classes of the testing examples are not known. The MT application may be used in screening situations. For example, if a new data example from a patient is required to be classified, then the user can simply enter details about it into the MT application. The data example can then be classified accordingly.

The BT application is similar to the MT and the ST applications in that the classes of the testing examples are not known, yet two data files are required (one for testing and one for training).

## G.2 How to use the Data File Creator application

### G.2.1 How to create a data set txt file

Prior to using the Data File Creator application, the data set must be structured in a simple tab delimited text file. This can be done manually, however a far easier method is to structure the data set using a spreadsheet application such a Microsoft Excel as shown in Figure G.2. Once the data is structured correctly, click on menu bar **File** → **Save As** and select 'Text (Tab delimited)' in the **Save as type:** option menu.

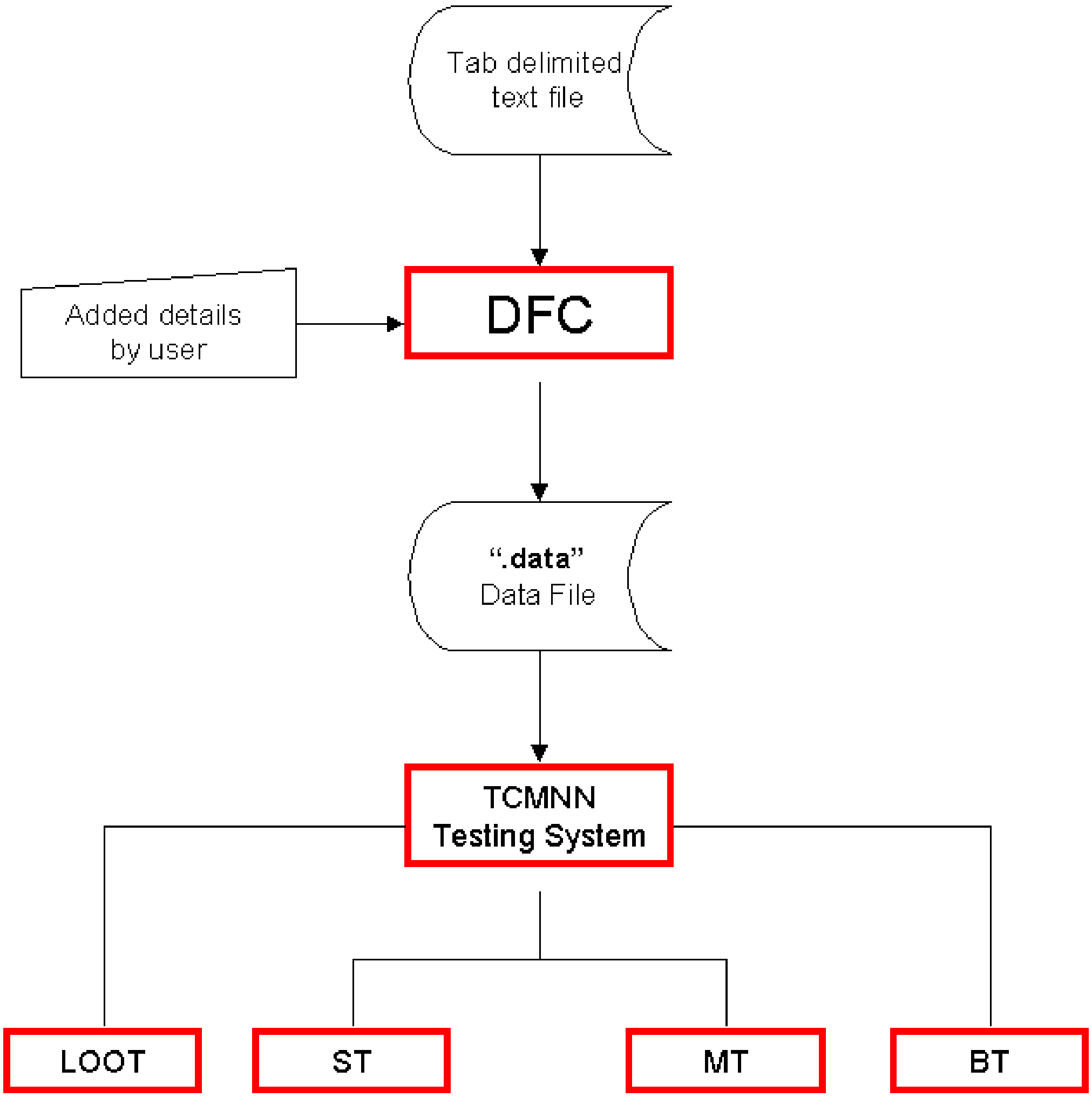

Abbreviations: DFC= Data File Creator; LOOT = Leave One Out Test; ST = Separate Test; MT = Manual Test; BT = Batch Test

Figure G.1: **An overview of the components involved in the TCMNN system.**

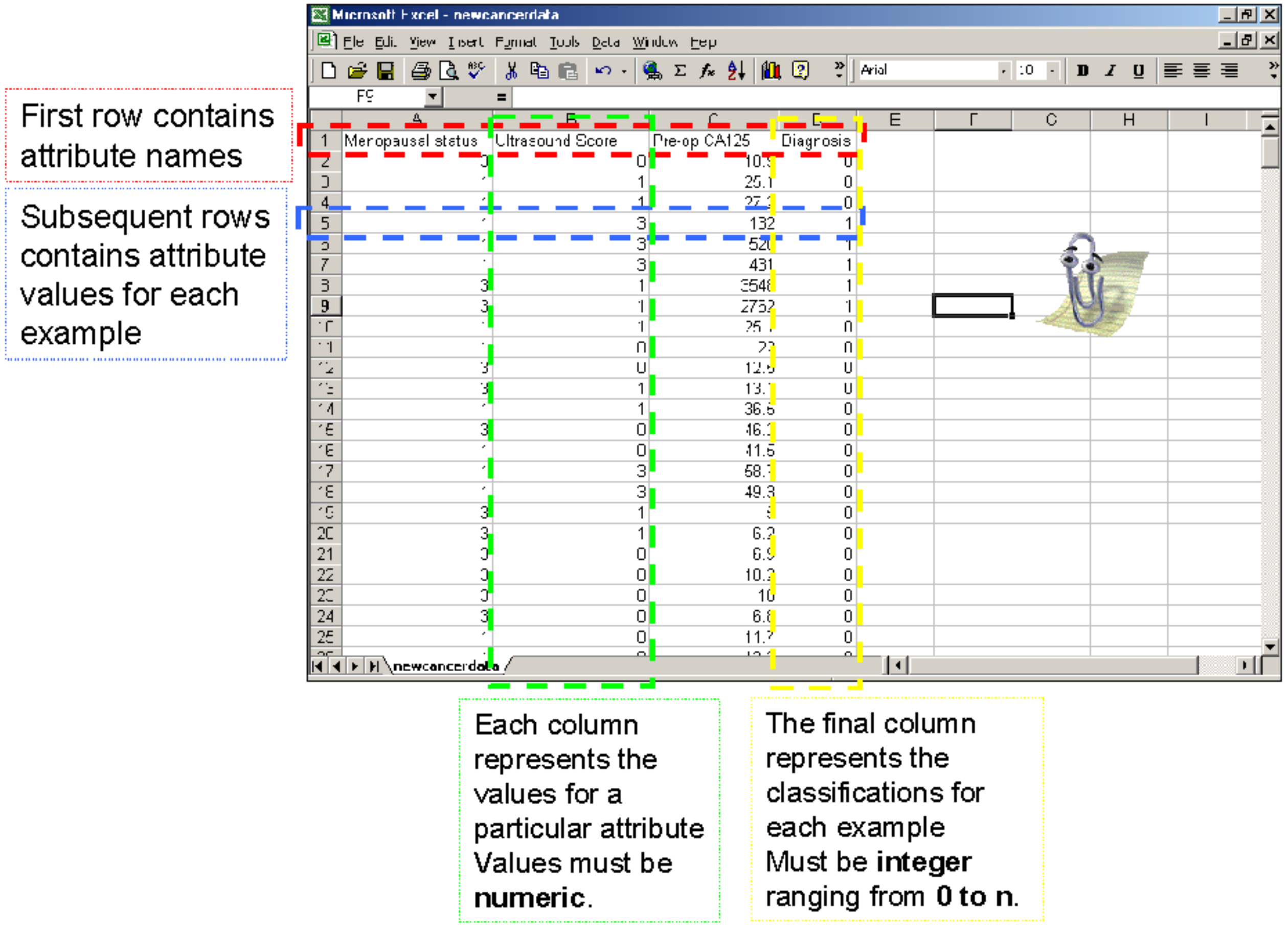

Figure G.2: **Example of how to use Microsoft Excel to structure and create a data set text file:** A simple spreadsheet application can be used to structure data sets so that they can be easily saved as tab delimited text files. These text files can then be converted into data files using the Data File Creator application.

### G.2.2 Running the Data File Creator application

At the dos prompt type:
`java DF_Creator`
The Data File Application will then start up, see Figure G.3.

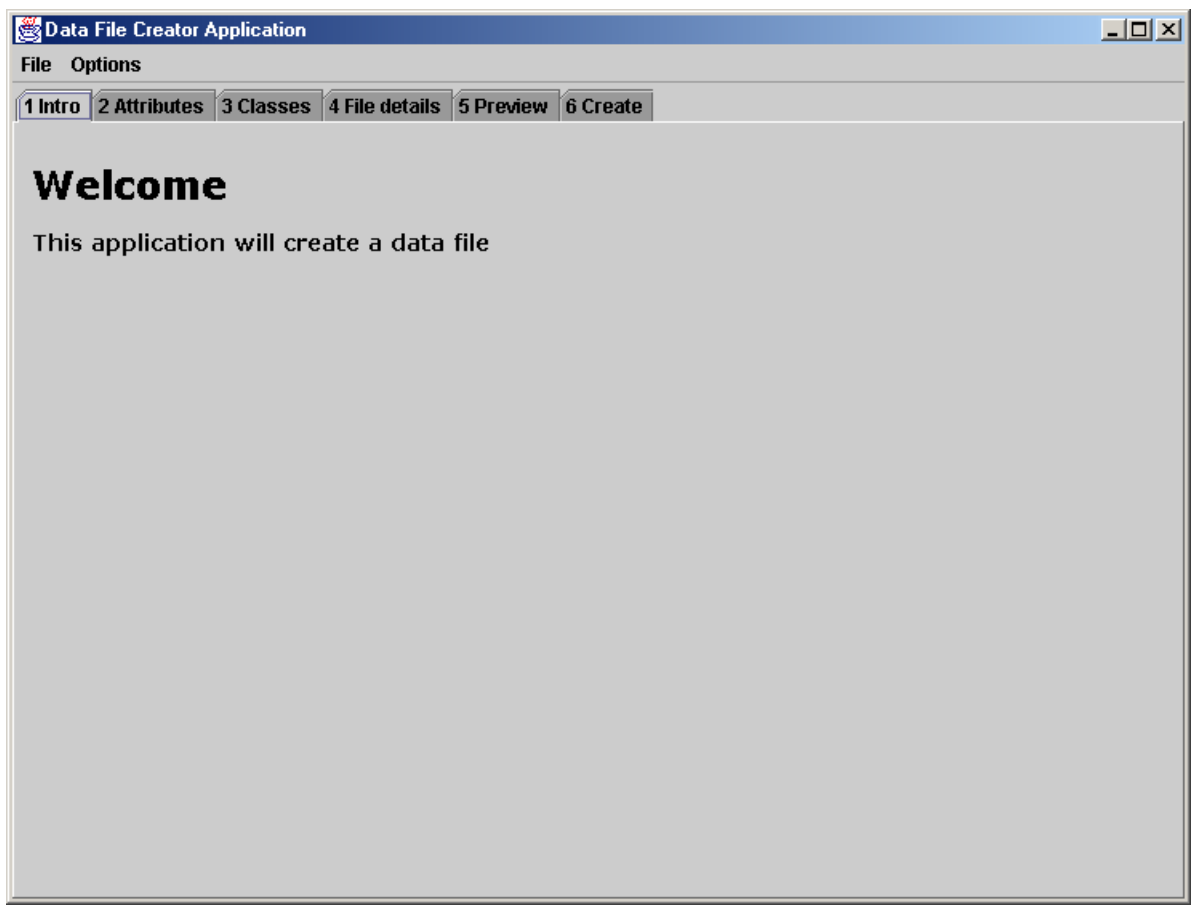

Figure G.3: **Data File Creator application screen shot:** The initial startup window.

Other applications that make up the TCMNN system are similar in layout to the application seen in Figure G.3. Each application uses drop down menus to provide more advanced functionality, and simple tabular panes to provide the main core input to the programs.

### G.2.3 How to create the data file

After running the Data File Creator application it is necessary to choose appropriate tabbed panes to enter specific details about the tab delimited text file created previously. These details will then be encapsulated in the resulting end data file. During data file creation, you will be provided with on screen help which should aid you through the process.

The next sections will explain the functions of each of the tabbed panes that make up the Data File Creator application and how they can be used.

#### Attributes Tab

This tabbed pane queries for details about the attributes that describe the data set. The input dialogue box shown in Figure G.4 shows the questions that are asked during creation of the data file.

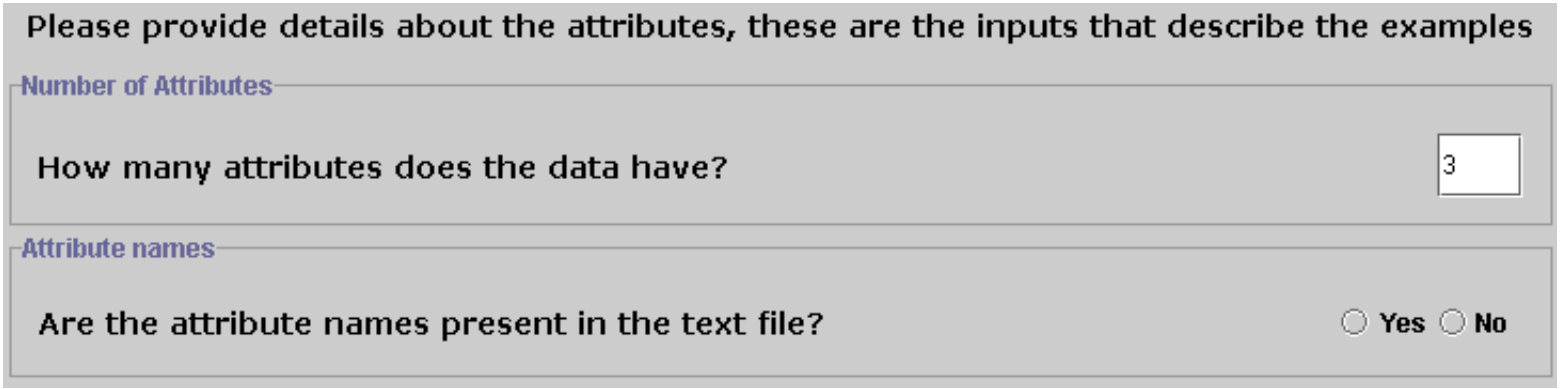

Figure G.4: **Entry dialogue for the attributes tab of the Data File Creator application :** Click the left button inside the text box to change the number of attributes. The number must be an integer value. Click either the 'yes' or 'no' radio buttons to specify whether the text file version of the data set contains the attribute names.

### Classes Tab

This tabbed pane queries for details about the classifications used in the in the data set. The input dialogue to this pane is shown in Figure G.5. The program dynamically creates text boxes to allow you to enter each classes name. If no names are provided the program uses by default the names 'CLASS NAME X', where X is a number from 0 to 1−the number of classes.

### Text file details Tab

This tab is the final required input that specifies the details about the data set text file. The input dialogue for this tabbed pane can be seen in Figure G.6. It is necessary to initially enter the number of examples that are present in the text file.

Finally details must be provided about the location of the text file on the local system. This can be done either by typing in the file name and its directory into the text box, or clicking on the 'browse' button to bring up the file chooser dialogue box shown in Figure G.7.

### Preview Tab

Now that the details about the text file have been entered, the program will provide a preview of the data set before you decide to create the data file as shown in Figure G.8. The scroll bars at the side can be used to view the complete data set.

This allows you to check that the details provided are correct. The data set is formatted using the terminology (attribute names , class names) provided on the previous tabbed panes of the application.

If the file is very large or if you do not wish to preview the data then you can disable this option by selecting from the menu bar **Options → Preview Disabled**.

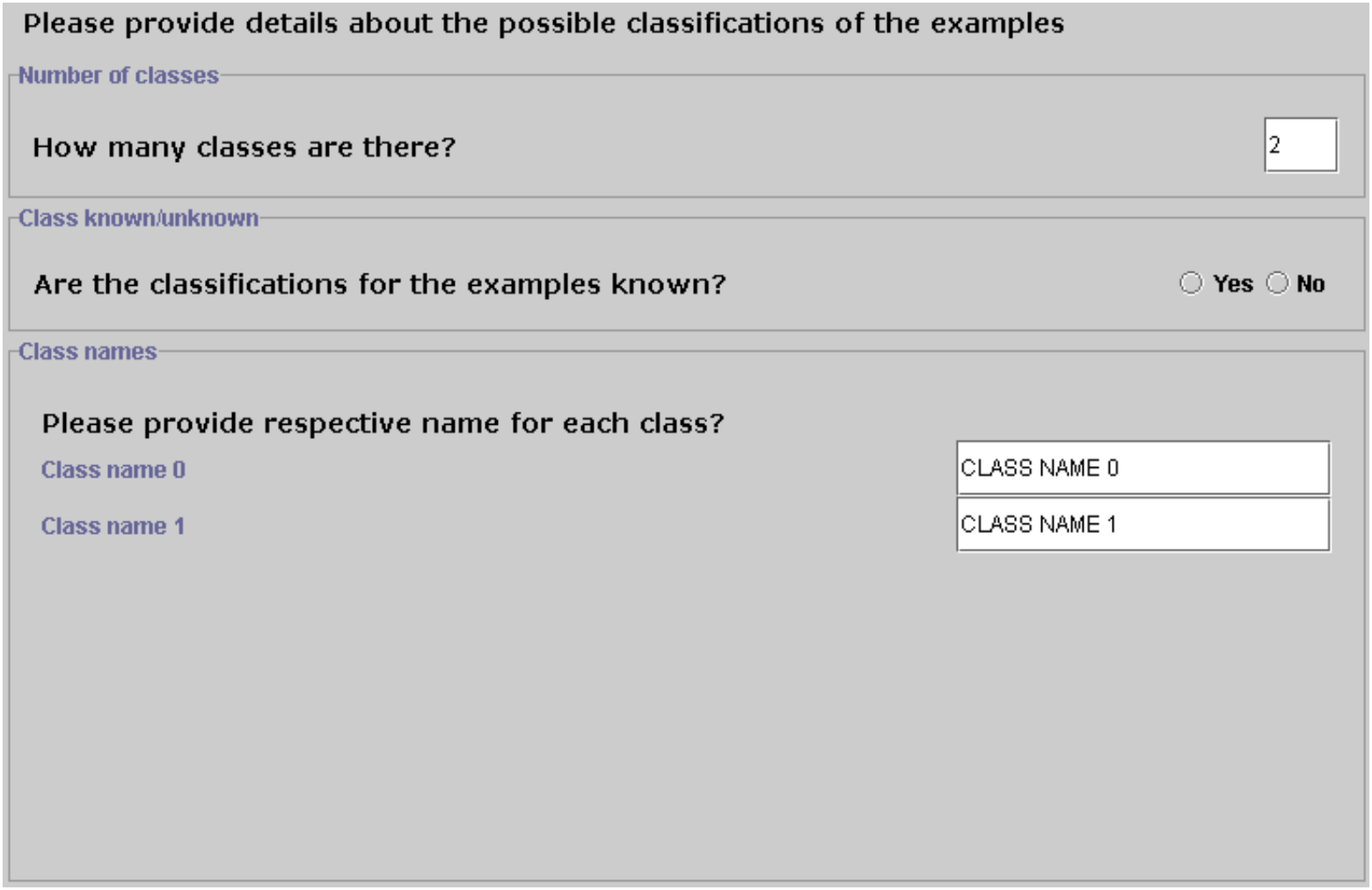

Figure G.5: **Entry dialogue for the classes tab of the Data File Creator application :** Click the left button inside the text box to change the number of classes. The number must be an integer value. Once you have specified the number of classes, the program will then create the appropriate number of text boxes so that you may enter the name of each class.

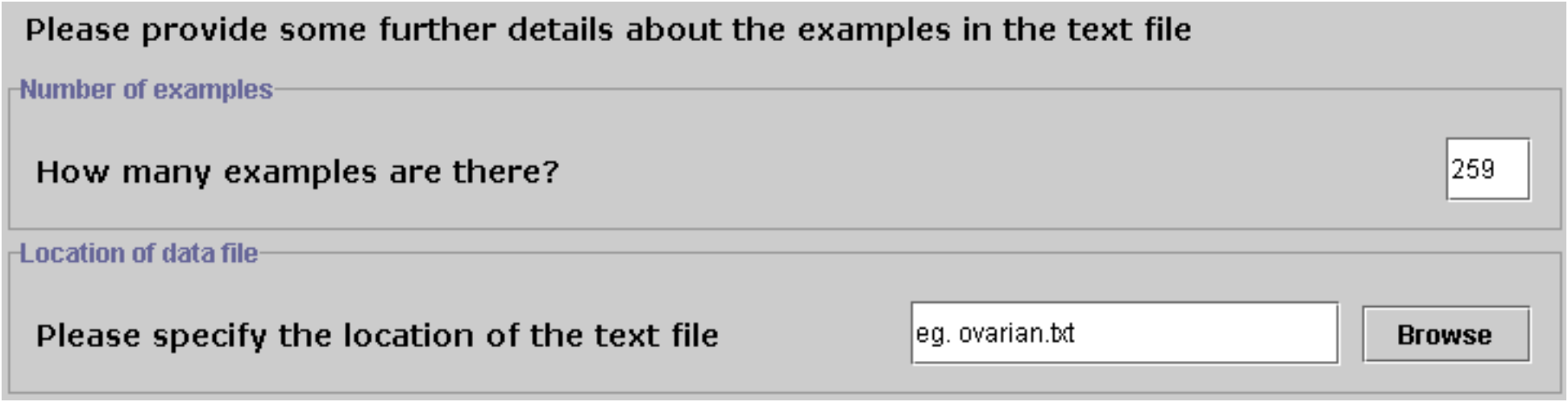

Figure G.6: **Entry dialogue for the Text file details tab of the Data File Creator application :** This is the final required input dialogue box of the system.

### Create data file Tab

This is the final tabbed pane of the application. Once you are satisfied with previous inputs select the 'Create' button as shown in Figure G.9.

If the creation of the data file has been successful the program will create a data file named 'filename.data', where 'filename' is the name of the original text file used to create it. The program will also signify the files successful creation by a pop up window as shown in Figure G.10. The data file will be automatically created in the working directory that the application is running in.

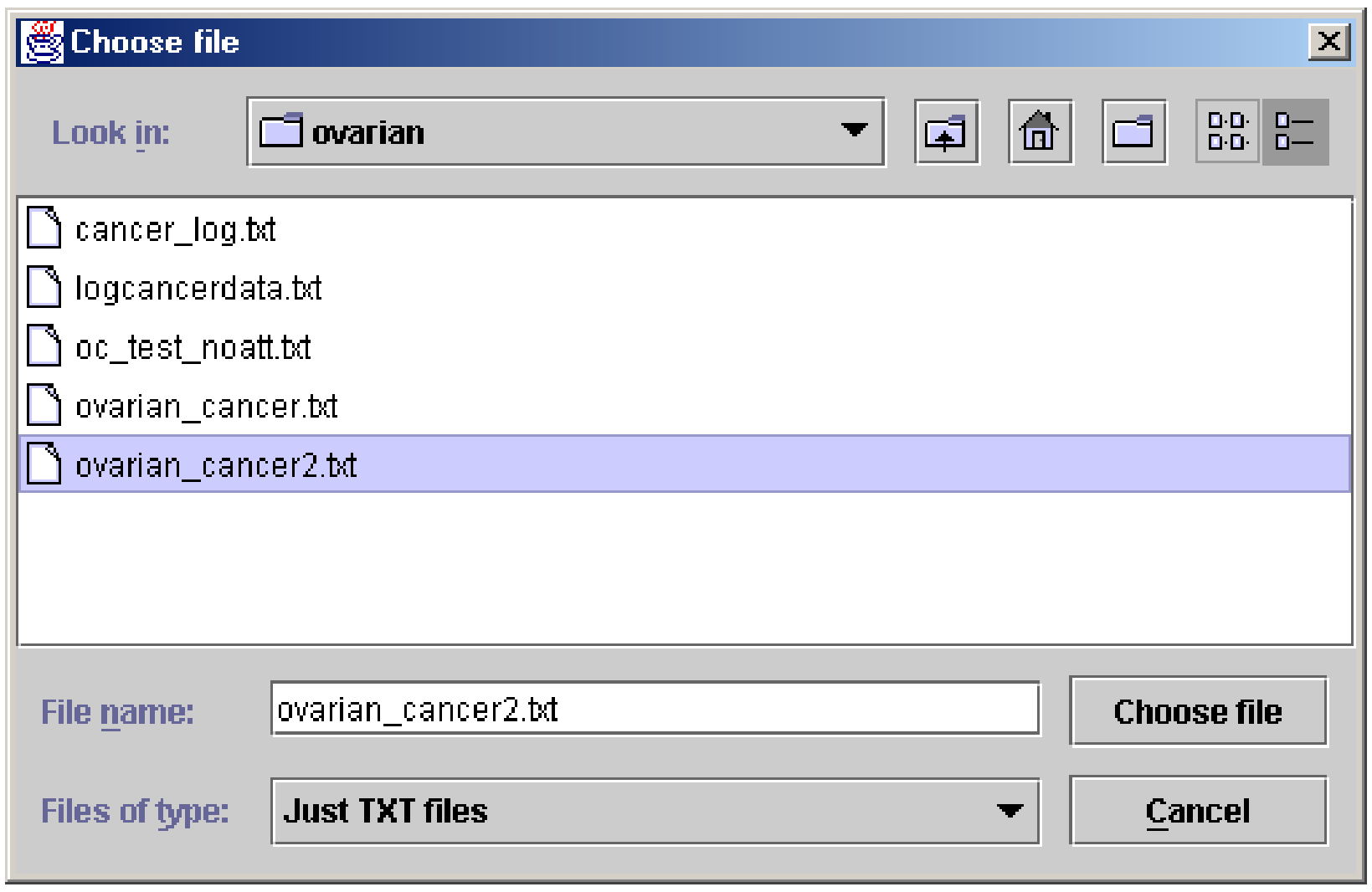

Figure G.7: **File chooser dialogue for Data File Creator application:** This file chooser searches and displays only text files on the local system. Once you have located the text file, select the text file, and then select the 'Choose File' button.

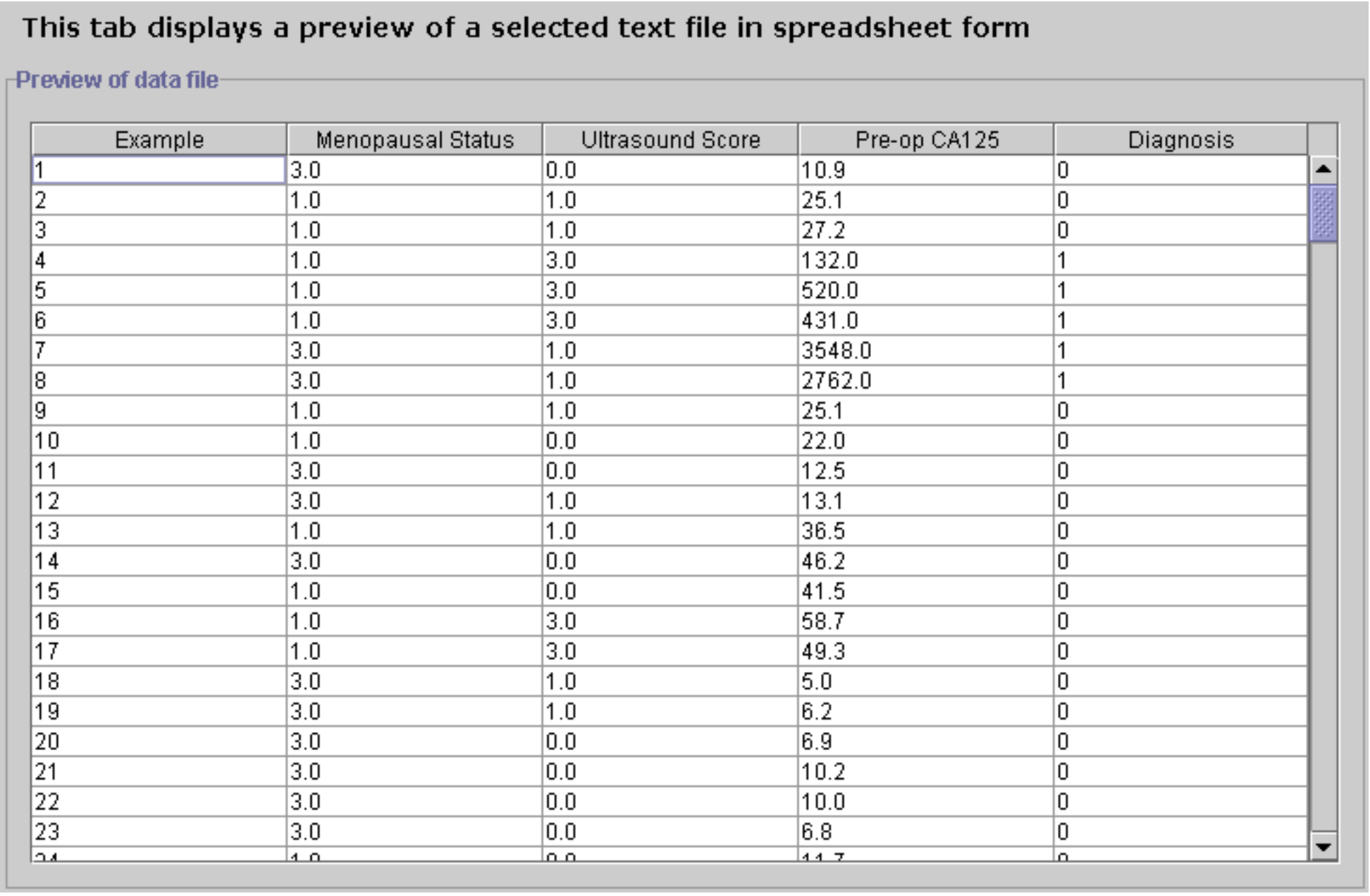

| Example | Menopausal Status | Ultrasound Score | Pre-op CA125 | Diagnosis |
|---|---|---|---|---|
| 1 | 3.0 | 0.0 | 10.9 | 0 |
| 2 | 1.0 | 1.0 | 25.1 | 0 |
| 3 | 1.0 | 1.0 | 27.2 | 0 |
| 4 | 1.0 | 3.0 | 132.0 | 1 |
| 5 | 1.0 | 3.0 | 520.0 | 1 |
| 6 | 1.0 | 3.0 | 431.0 | 1 |
| 7 | 3.0 | 1.0 | 3548.0 | 1 |
| 8 | 3.0 | 1.0 | 2762.0 | 1 |
| 9 | 1.0 | 1.0 | 25.1 | 0 |
| 10 | 1.0 | 0.0 | 22.0 | 0 |
| 11 | 3.0 | 0.0 | 12.5 | 0 |
| 12 | 3.0 | 1.0 | 13.1 | 0 |
| 13 | 1.0 | 1.0 | 36.5 | 0 |
| 14 | 3.0 | 0.0 | 46.2 | 0 |
| 15 | 1.0 | 0.0 | 41.5 | 0 |
| 16 | 1.0 | 3.0 | 58.7 | 0 |
| 17 | 1.0 | 3.0 | 49.3 | 0 |
| 18 | 3.0 | 1.0 | 5.0 | 0 |
| 19 | 3.0 | 1.0 | 6.2 | 0 |
| 20 | 3.0 | 0.0 | 6.9 | 0 |
| 21 | 3.0 | 0.0 | 10.2 | 0 |
| 22 | 3.0 | 0.0 | 10.0 | 0 |
| 23 | 3.0 | 0.0 | 6.8 | 0 |

Figure G.8: **Entry dialogue for the Preview tab of the Data File Creator application :** This tab provides a preview of the data set provided in the text file.

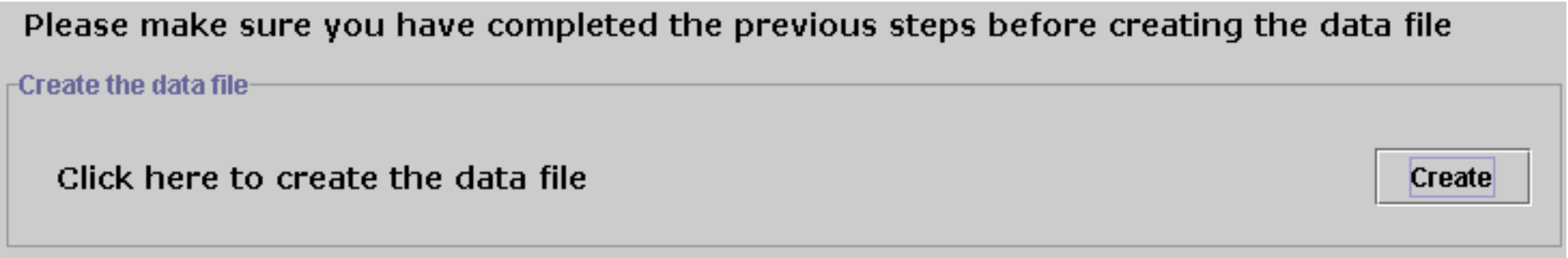

Figure G.9: **Entry dialogue for the Create data file tab of the Data File Creator application :** This is the final required input dialogue of the system.

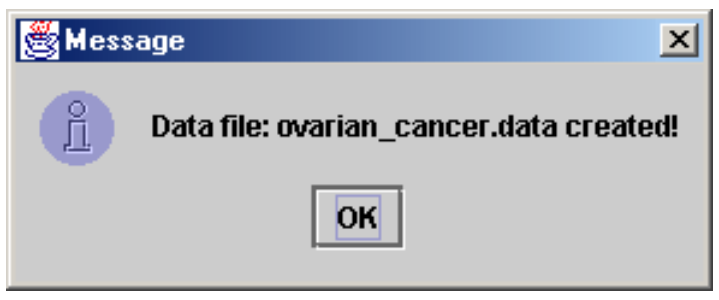

Figure G.10: **Pop up window for the Create data file tab of the Data File Creator application :**

## G.3 How to use the Leave One Out / Separate Test Applications

Both Leave One Out and Separate Test applications use similar interface layouts to that of the Data File Creator application. This section will give a general description of how to use these systems. These application(s) will allow you to run the TCMNN algorithm on your data set(s), receive hard copy outputs of predictions and mark the performance of the TCMNN's predictions.

### G.3.1 Requirements for usage

Both applications require data files where the classes are known to be able to output a statistical performance of the TCMNN process. Before using the system you must have created the data sets in data file format as described earlier. The Leave One Out test application requires one data file to cross validate with itself, whereas the Separate Test application requires two files, one for training and another for testing.

### G.3.2 How to run the application

Both systems are simply run from a dos prompt (or equivalent environment) by typing the following:
`java TCMNN_LeaveOneOut_Application`
or,
`java TCMNN_Separate_Application`
This will then bring up the main application screen as shown in Figure G.11.

These applications have a very similar interface to that of the Data File Creator application mentioned earlier. The drop down menus provide more advanced features for configuring the TCMNN algorithm. The main tabbed panes provide the core functionality of the system which must be completed to run the test.

#### Data file details Tab

This tab is used to specify which data file(s) will be tested using the TCMNN algorithm. The Separate test application has two of these tabs to specify both the training and testing data files individually. A screen shot of the input dialogue for this tab before a data file has been selected can be seen in Figure G.12.

The location of the data file to be used in the test must be specified to the program, this can be done by either typing the directory and the filename into the text box, or by clicking on the 'browse' button to bring up a file chooser dialogue as shown earlier in Figure G.7. This file chooser is automatically set to filter only data files on the local system to be shown.

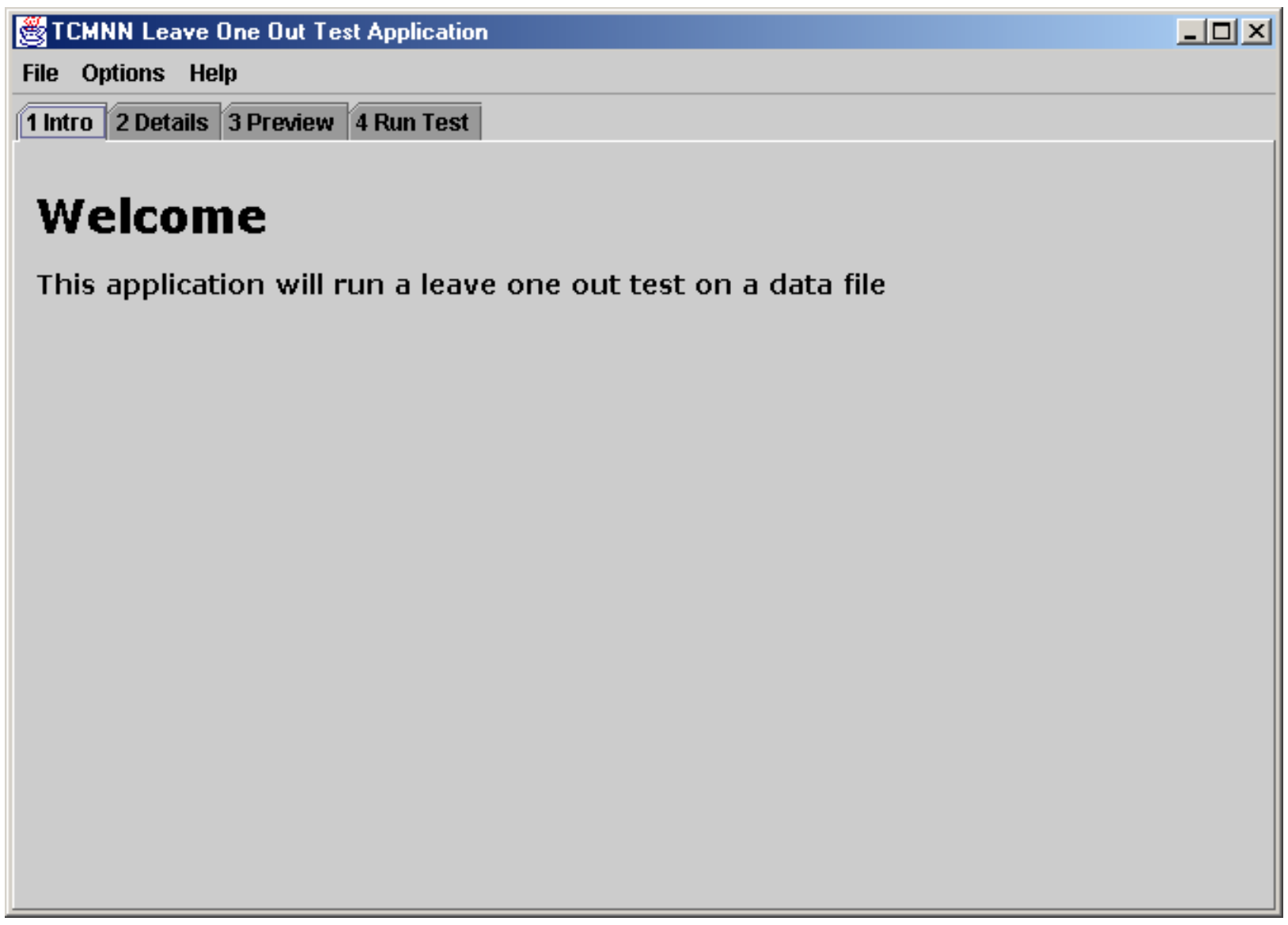

Figure G.11: **Leave one out test application main screen :**

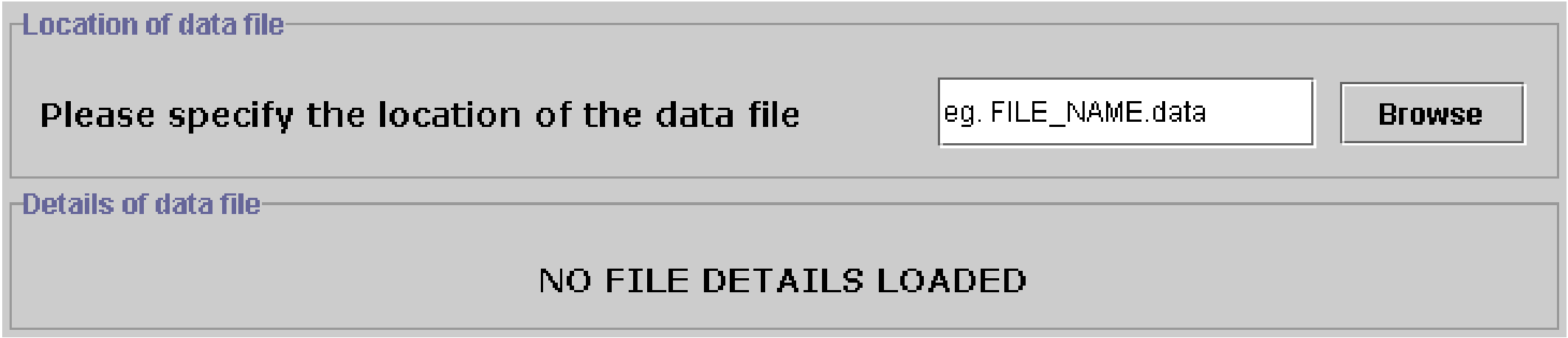

Figure G.12: **Leave one out test application Data file details tab screen shot before data file selected :** Once a data file has been selected this area is updated as seen in Figure G.13

Once the data file has been selected the Data file details tab shown in Figure G.12 is updated with an overview of all the details of the data file such as number of attributes. An example of this data file overview update is seen in Figure G.13. **IMPORTANT:** with separate tests, the training and testing data files must be of the same dimensions (same number of classes, attributes and so forth).

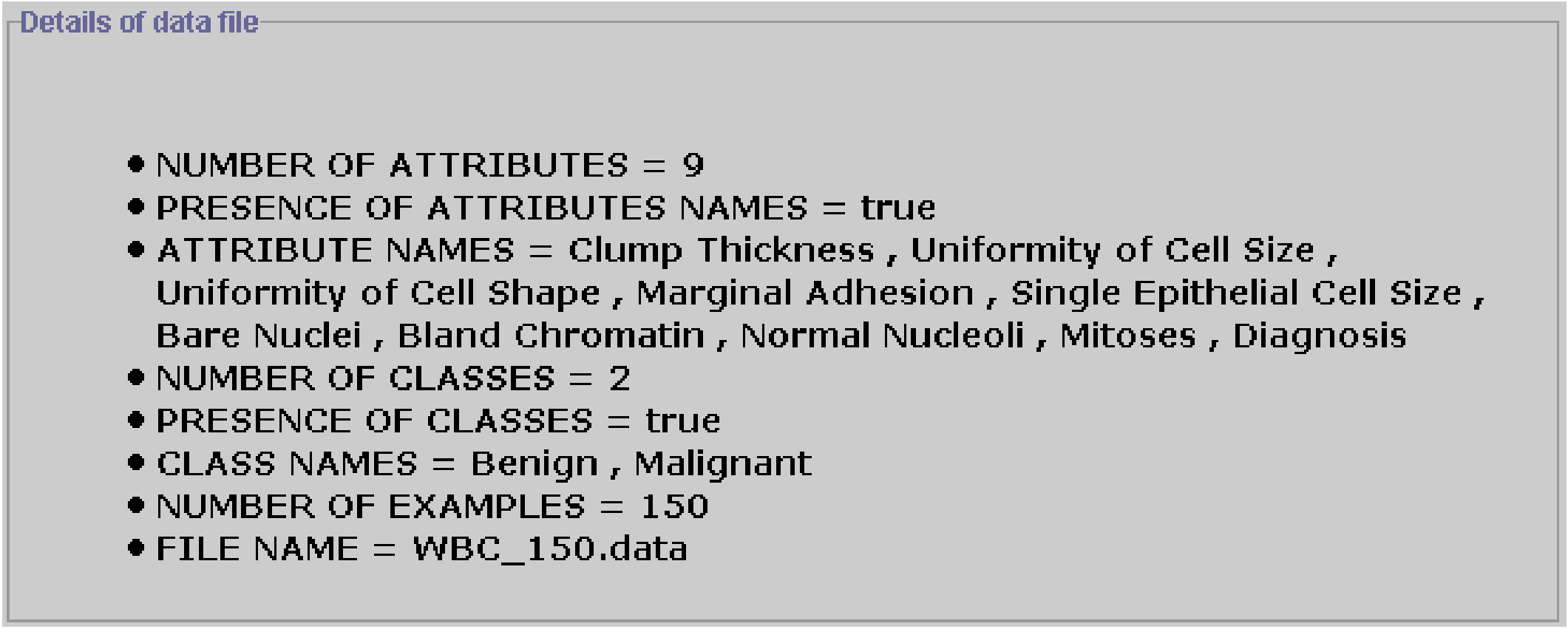

Figure G.13: **Leave one out test application Data file details tab screen shot after data file selected :** This uses the details stored in the data file to give an overview of the data set providing information such as names of class and attributes, number of examples etc.

### Preview Tab

This tab offers the same functionality as that of the Preview tab in the Data File Creator application (see Figure G.8). It enables you to preview the data file before running the TCMNN test. This can be disabled in the same way as mentioned earlier of clicking on the menu bar **Options** → **Preview Disabled**.

### Run test Tab

This is the final tab in both applications. Once the data files have been provided for the test, this tabbed pane can be used to initiate the test, and control where the outputs are stored. The Run test tab (Figure G.14) also displays the currently selected options for the test. The default settings are 1 nearest neighbour, using the standard Euclidean distance metric.

The results and statistics file name dialogue boxes are automatically filled with a default filename based on the original names of the data files provided for the test. These default names can be changed to alternative names by clicking inside in the text box and keying in a new name. These filenames must have the '.html' extension so that they can be read by web browsers to be viewed.

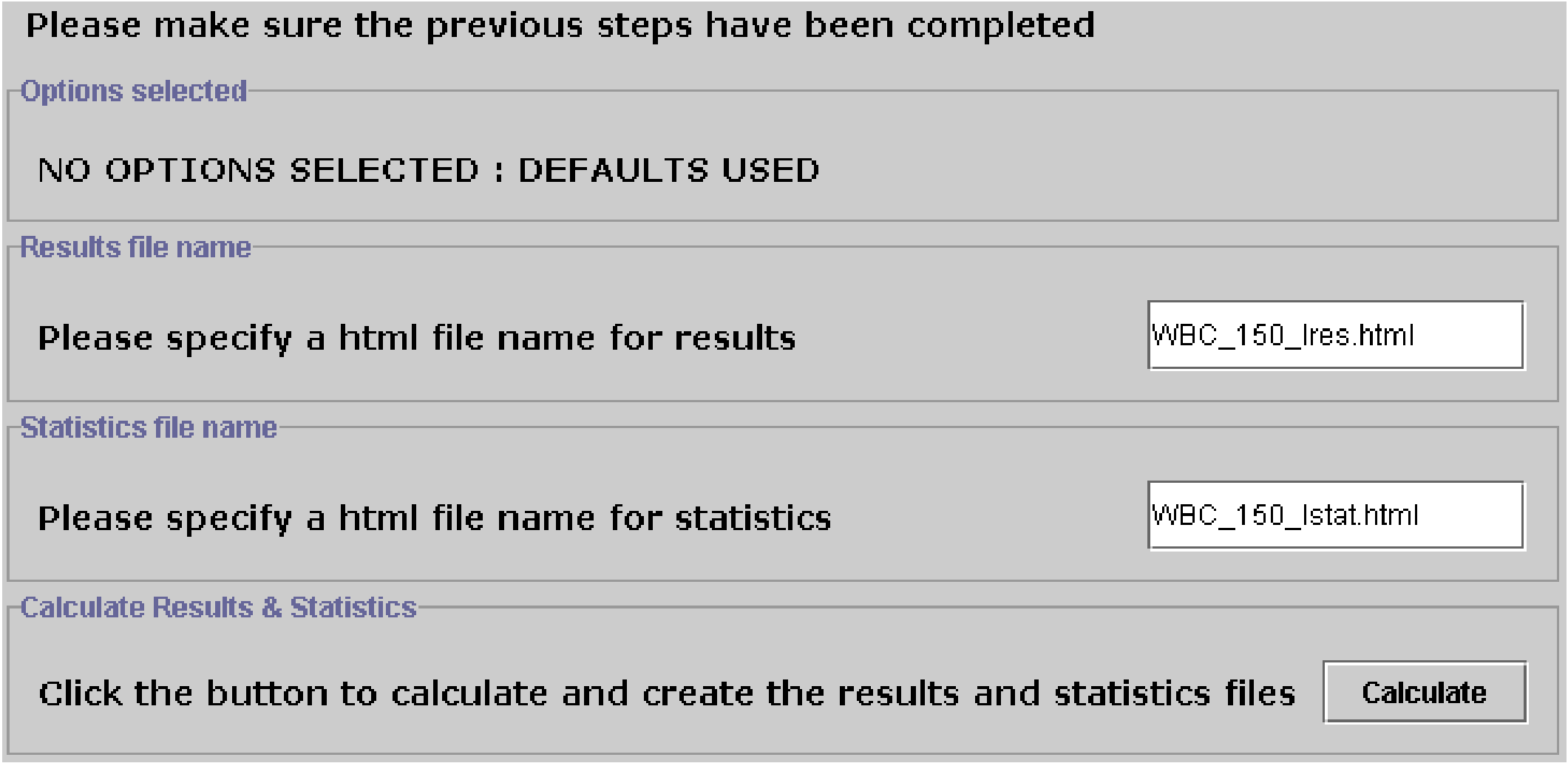

Figure G.14: **Leave one out test application Run test tab screen shot :** Currently selected settings for the TCMNN are displayed in the 'Options selected' screen area. The names of the results and statistics files can be changed from the default names by typing in the text boxes shown. Once satisfied with the setup of the TCMNN test, the test can be run by clicking on the 'Calculate' button.

### G.3.3 Options menu

All the TCMNN applications have an options menu to configure parameters of the TCMNN as well as to configure the results and statistics files output. This section will describe all the functionality available through these menus.

#### Set K value

This option allows you to configure the number of K nearest neighbours that the TCMNN algorithm analyses to on the data set. To set the K value, click on the menu bar, **Options** → **Set K value**. This will then bring up a pop up option pane to enter the number into shown in Figure G.15. This value must be a positive integer. The standard setting is K=1.

#### Select distance metric

This option menu allows you to select which distance metric is used with the TCMNN algorithm when testing on the data sets. There are three options available:

1. **Euclidean distance** - this is the standardly used distance metric. To select

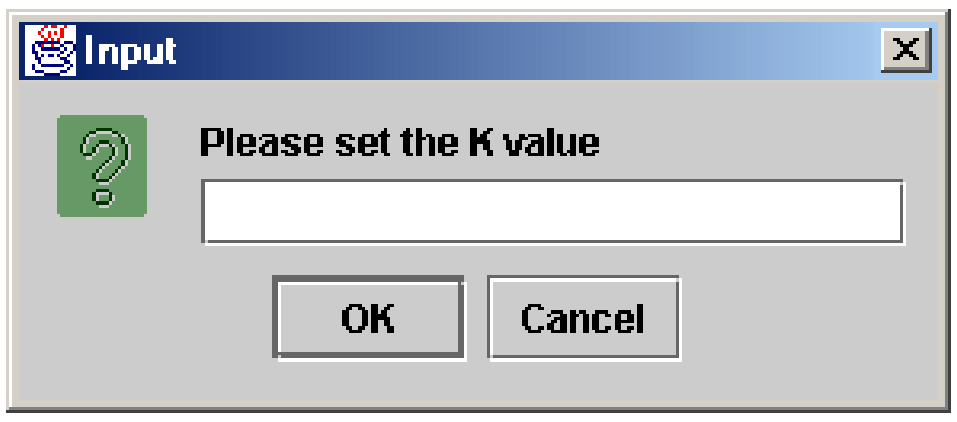

Figure G.15: **Leave one out test application K value option pane**

it, click on the menu bar, **Options** → **Select distance metric** → **Euclidean distance**.

2. **Polynomial kernel** - this computes the Euclidean distance between a polynomial kernel induced feature space. To select it click on the menu bar, **Options** → **Select distance metric** → **Polynomial kernels**. This will then bring up the polynomial kernel option panes as shown in Figure G.16.

3. **Minkowski metric** - this setting allows you to use Minkowski metrics of various degrees with the TCMNN algorithm. To select it click on the menu bar **Options** → **Select distance metric** → **Minkowski metric**. This will then bring up the polynomial kernel option panes as shown in Figure G.17.

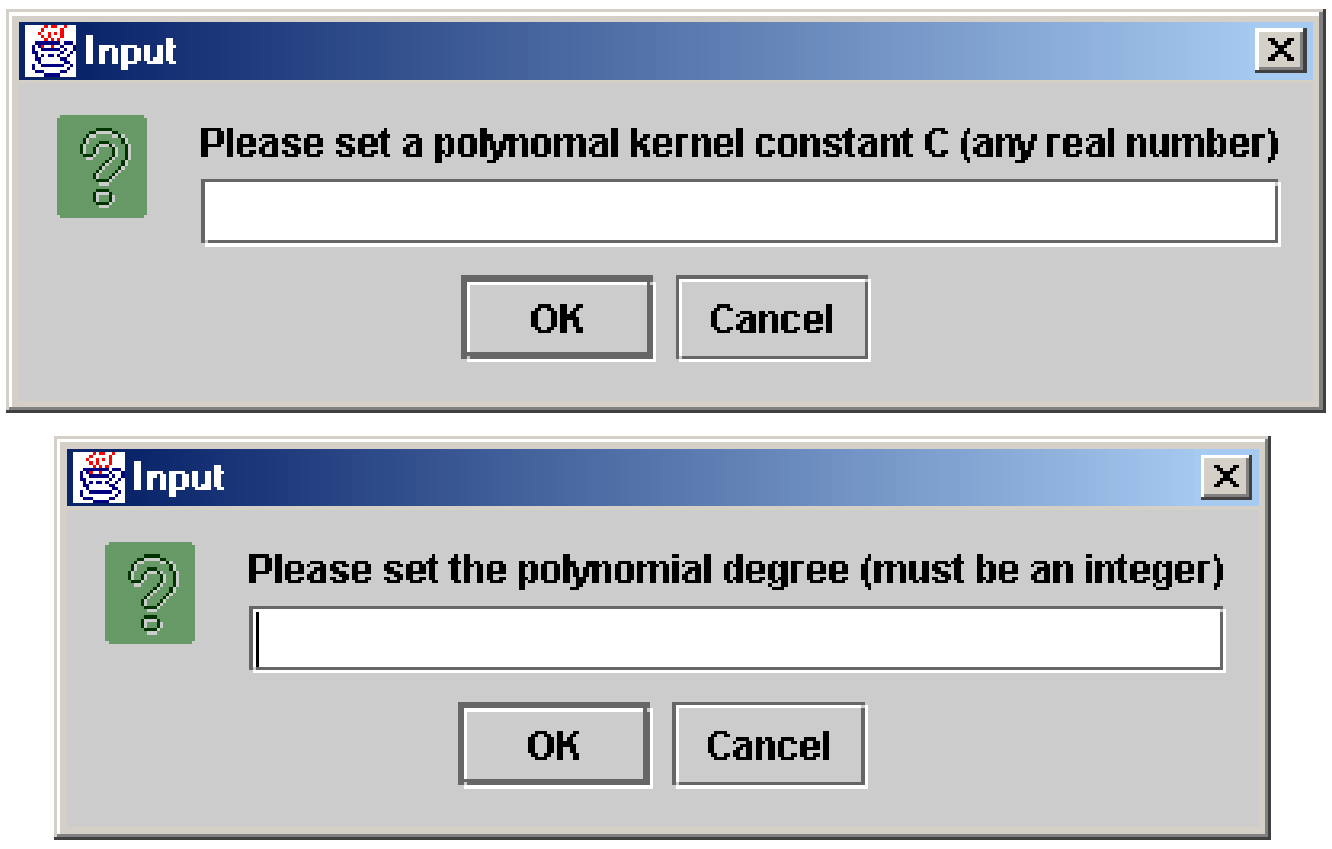

Figure G.16: **Leave one out test application Polynomial kernel option panes:** The panes allow you to set the polynomial kernel degrees and constants. The degree of the polynomial must be a positive integer, and the constant must be a positive real number.

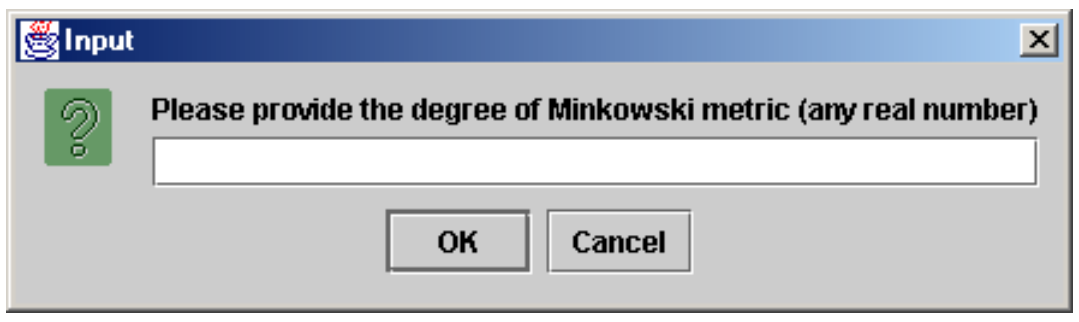

Figure G.17: **Leave one out test application Minkowski metric option panes:** This allows you to set the degree of the Minkowski metric. The value must be a positive real number.

#### Statistics

This tab allows you to configure the way in which the statistics of the TCMNN test performances are laid out. This menu has 2 available options.

1. **Basic stats** - This will set the output to only perform a basic statistical analysis, where the details of the individual class performances are not calculated. Confidence and credibility histograms will not be presented in this form of output. To select click on the menu bar, **Options** → **Statistics** → **Basic stats**.

2. **Set histogram interval** - This allows you to set the size of the confidence and credibility intervals used in the histograms output in the statistical analysis. Histograms will be discussed in more detail later on. To select click on the menu bar, **Options** → **Statistics** → **Set histogram interval**.

#### Results

This option allows you to configure how the results of the TCMNN test are displayed in the output HTML files. This option specifies whether you wish to view the attributes of each example in the results file. This can be useful if the problem has many attributes. To select click on the menu bar, **Options** → **Results** → **View attributes**.

### G.3.4 Running the TCMNN test

Once the configuration for the TCMNN test has been selected using the options menu, the test can finally be run. The options that have been selected for the TCMNN can be checked by looking at the 'Options selected dialogue' of the Run test tab (as seen in Figure G.18).

Once you are satisfied with the options that are selected, click on the 'Calculate' button. This will in turn bring up a progress bar shown in Figure G.19. The progress bar keeps track of how long the TCMNN process will take to complete. Once finished the application will create HTML file outputs of results and statistics (depending on particular application used). The process completion is signified by the pop up windows displayed in G.20.

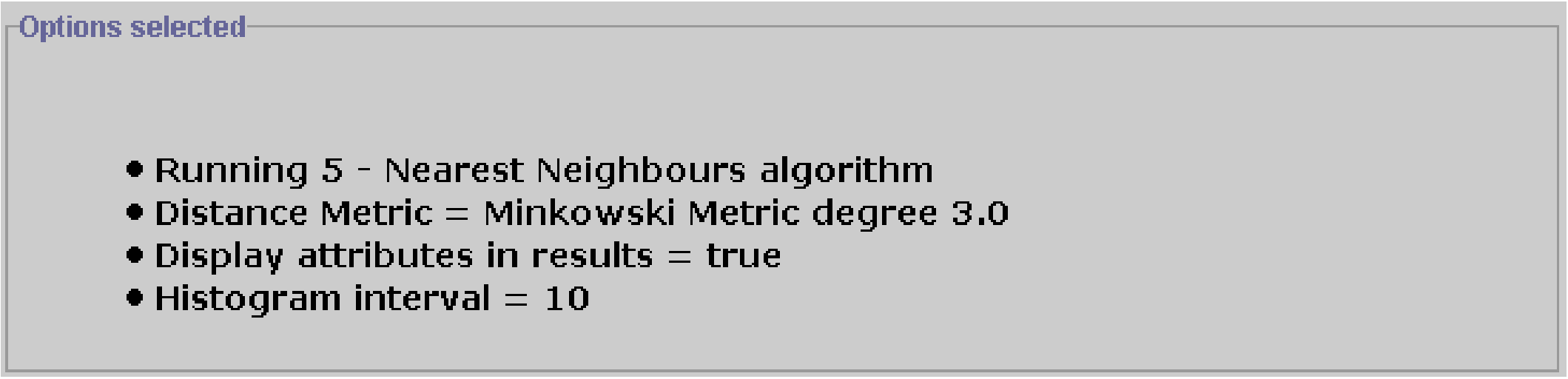

Figure G.18: **Leave one out test application Run test tab options selected dialogue:** This dialogue allows you to see which settings have been selected for use with the TCMNN.

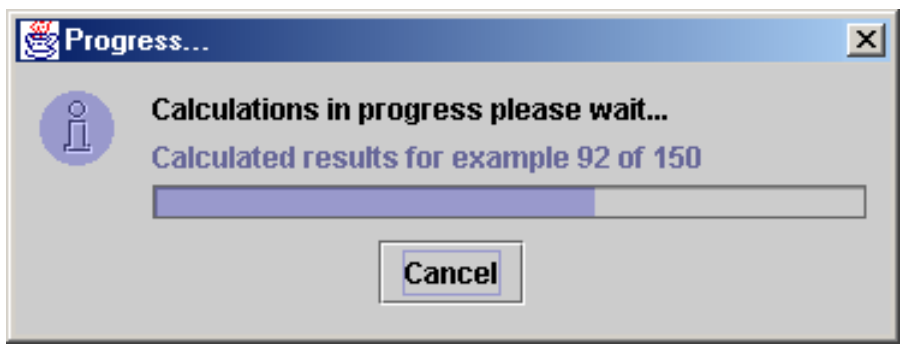

Figure G.19: **Leave one out test application Run test tab TCMNN progress bar:** This progress bar keeps track of the progress of the TCMNN process. The dialogue indicates what type of calculations are being performed.

### G.3.5 HTML output files

There are two main files output by the programs:

1. Results file - This provides details about the predictions made by the TCMNN algorithm. If the classes are known in the data files, then the prediction is compared with the real class.

2. Statistics file - This reports the performance of the TCMNN algorithm on the test data set. Note: this file is not generated with the manual and batch query testing systems as they do not have the classes known for the test set.

Both files start with a common header format that provides the following details:

1. Type of test performed eg. a testing mode such as Leave One Out, Separate etc.

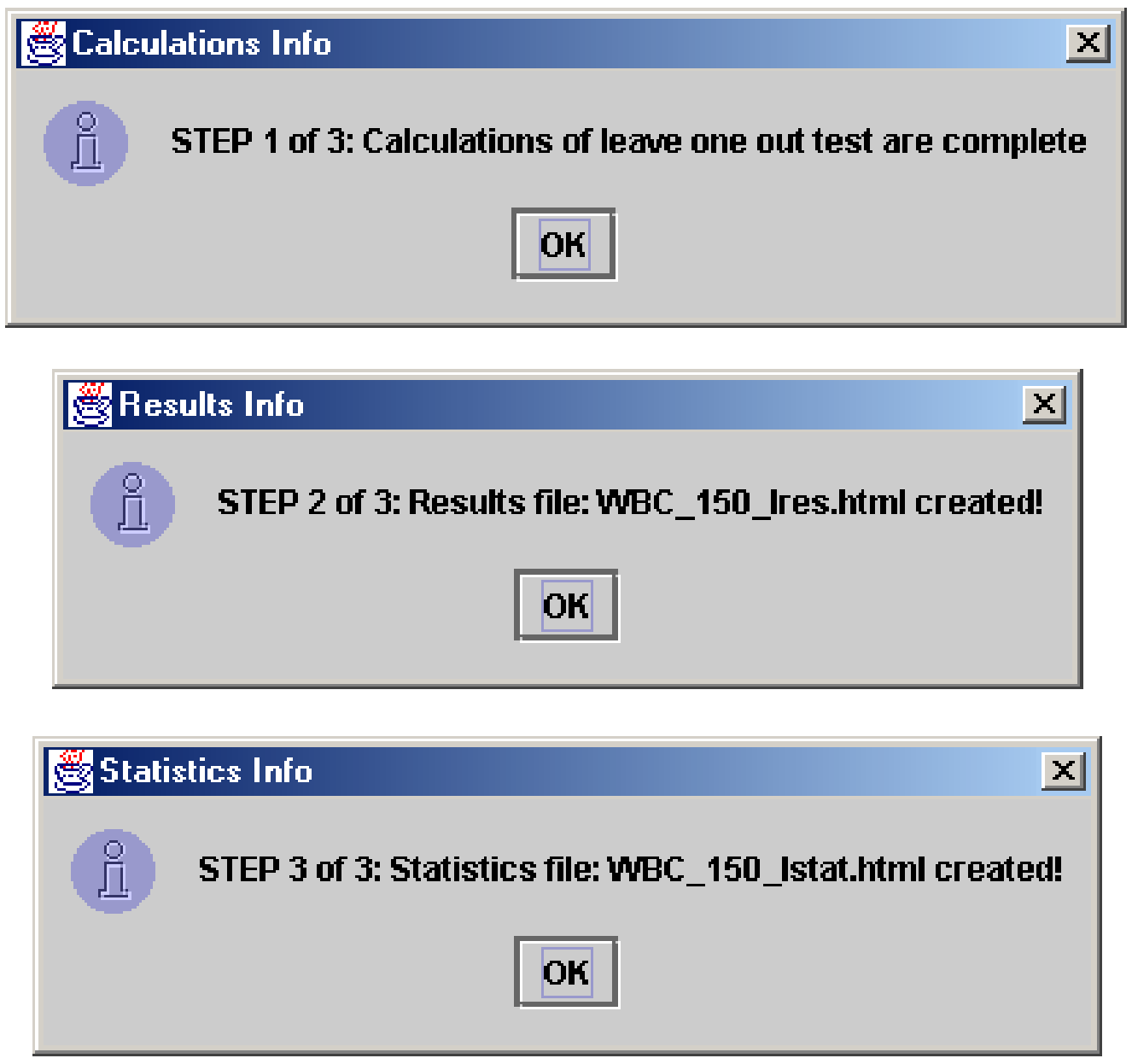

Figure G.20: **Leave one out test application Run test tab tests complete pop up dialogue:** This dialogue tells you when the TCMNN process is complete and that the output HTML files have been created successfully.

2. Number of nearest neighbours used.
3. Distance metric used such as Euclidean, polynomial kernel, Minkowski metric etc.
4. The data file names used in the test.

NOTE: Examples printouts of these HTML files can be seen in appendix F.

### Results HTML file

This file displays predictions from the TCMNN for each of the examples in the test set. The results are structured in simple HTML tables providing the following details:- (An example of these HTML results can be seen in Figure G.21)

- actual class (if classes known for test set)
- predicted class
- whether the example is correctly classified (if classes known for test set)

- confidence values
- credibility values
- p-values for each class (displayed graphically as bar chart)
- attribute values (can be disabled)

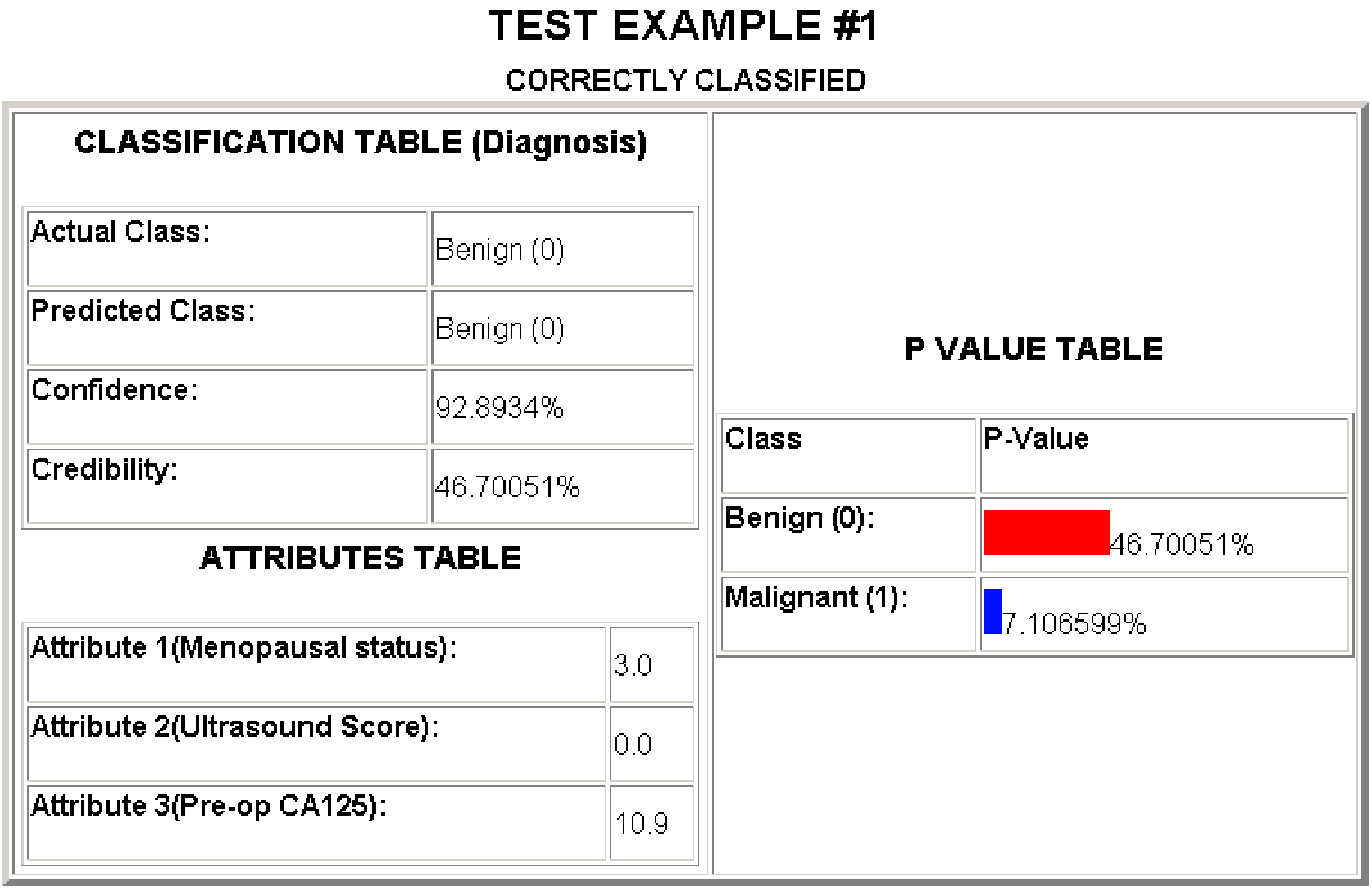

Figure G.21: **Example output of results HTML file:** The details of the result for each test example are structured using HTML tables. P-values are displayed graphically as coloured bar charts.

### Statistics HTML file

This file outputs a statistical performance of the TCMNN test. The analysis requires the real classes of the test examples, therefore this file is not output by the batch query and manual testing applications.

This file specifies the following details:

1. Overall and individual class performance assessment.
2. The number of training and testing examples provided in the data sets.
3. Average and confidence and credibility values.
4. Histograms of the confidence and credibility as seen Figure G.22.

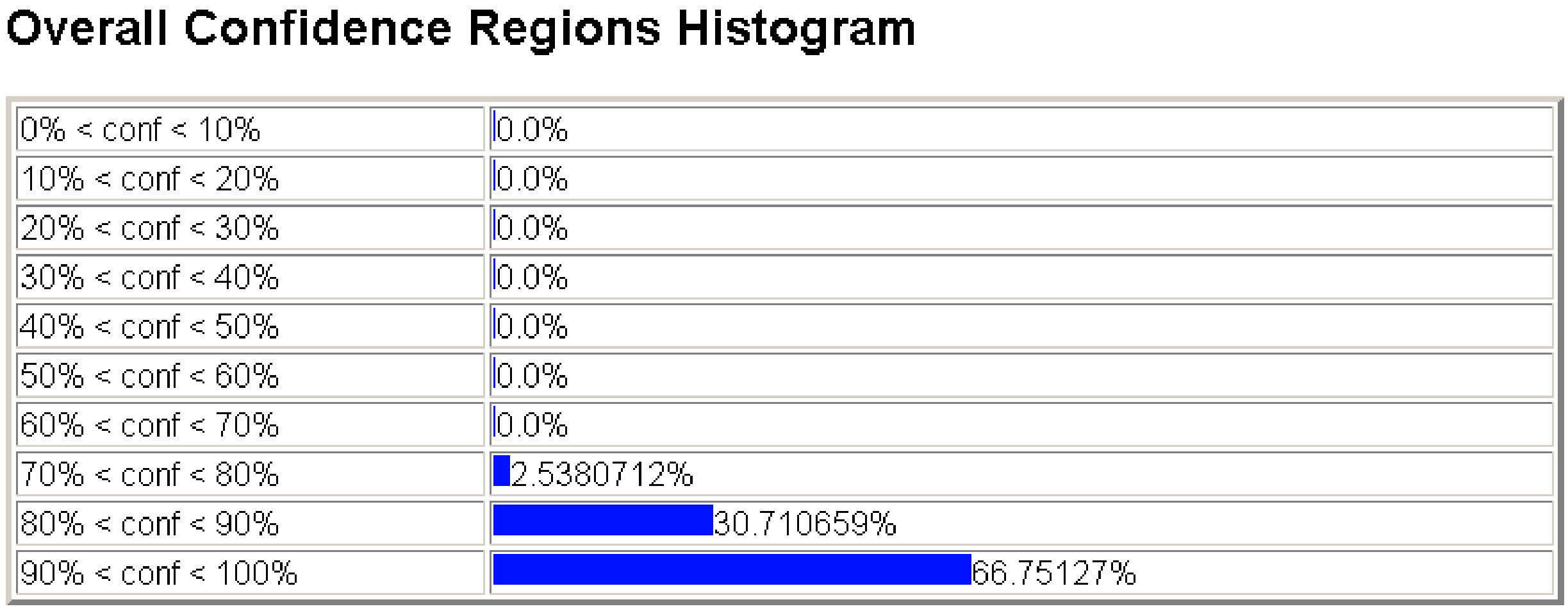

Overall Confidence Regions Histogram

| | |
|---|---|
| 0% < conf < 10% | 0.0% |
| 10% < conf < 20% | 0.0% |
| 20% < conf < 30% | 0.0% |
| 30% < conf < 40% | 0.0% |
| 40% < conf < 50% | 0.0% |
| 50% < conf < 60% | 0.0% |
| 60% < conf < 70% | 0.0% |
| 70% < conf < 80% | 2.5380712% |
| 80% < conf < 90% | 30.710659% |
| 90% < conf < 100% | 66.75127% |

Figure G.22: **Example confidence histogram of statistics HTML file:** Using the confidence and credibility values generated for each of the test examples, the application creates a histogram by counting the number of examples that have values between specific intervals. The size of these histogram intervals can be changed by the application as mentioned earlier.

## G.4 Manual and batch testing applications

These systems follow the same interface layout as the leave one out and separate test applications mentioned earlier. To run the manual and batch test applications type the following at a dos prompt (or equivalent environment):
`java TCMNN_Manual_Application`
or,
`java TCMNN_Batch_Application`

These applications only require one training data file (classes known) for use. The classes for the test examples do not need to be known for these tests. The main difference with the manual test application is that it allows you to manually enter the new examples to be tested. This is achieved by filling in a spreadsheet table as shown in Figure G.23. These applications output a results HTML file and not a statistics HTML file.

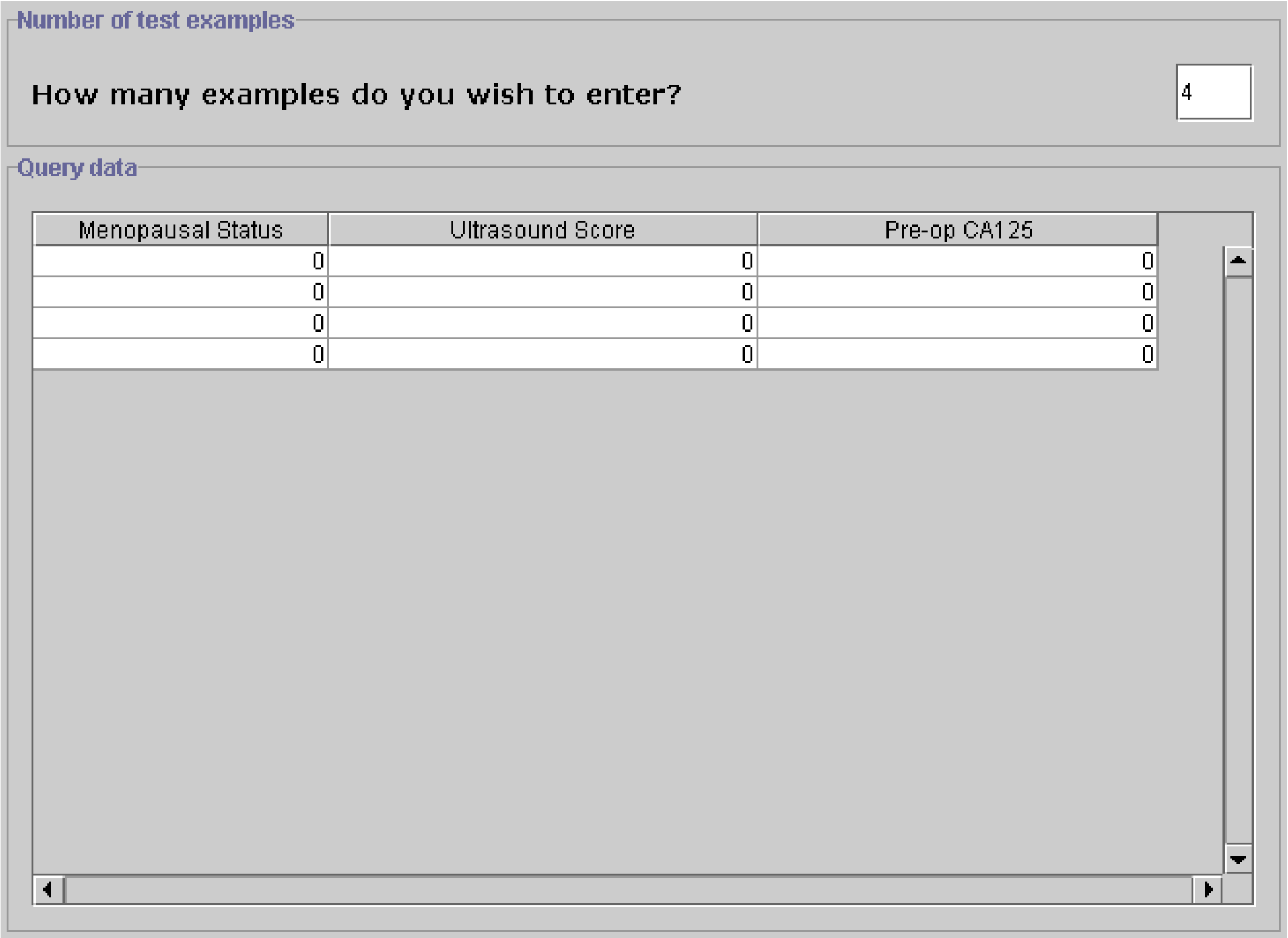

Figure G.23: **Query tab of the manual test application:** Entering the number of examples that you wish to test will create the appropriate amount of rows in the table for you to enter each examples details. Each column of the table corresponds to a particular attribute. The attribute names are read from the data file and used as the column headings.

## G.5 Image Data File Creator application

This application has similar functionality to the Data File Creator application mentioned earlier. This application converts directories of images into a data file that can be used by TCMNN image testing systems.

To run the Image Data File Creator application type the following at a dos prompt (or equivalent environment):
`java Image_Data_File_Creator`
This will then bring up the main screen shown in Figure G.24.

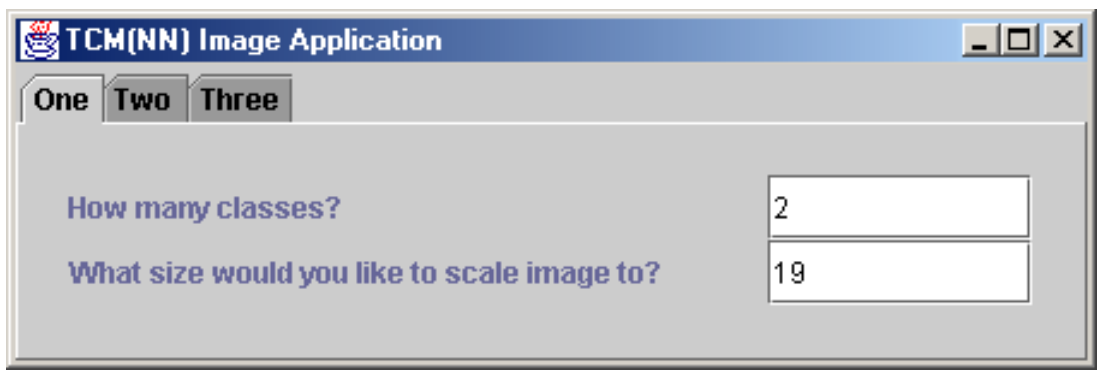

Figure G.24: **Main screen of Image Data File Creator application:** The application is made up of tabular panes and drop down menus.

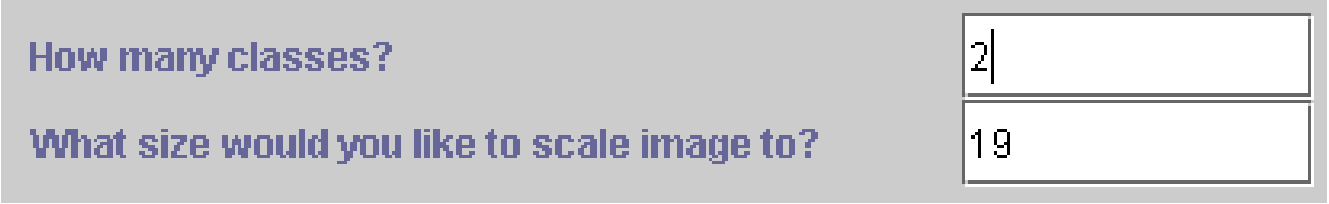

Figure G.25: **Classes details tab of Image Data File Creator application:** This is the first tab of the application. The first text box specifies how many different classes are going to be used in the problem. This number will set the appropriate number of text boxes needed on the second tab to enter the locations of the directories containing the images for each class.

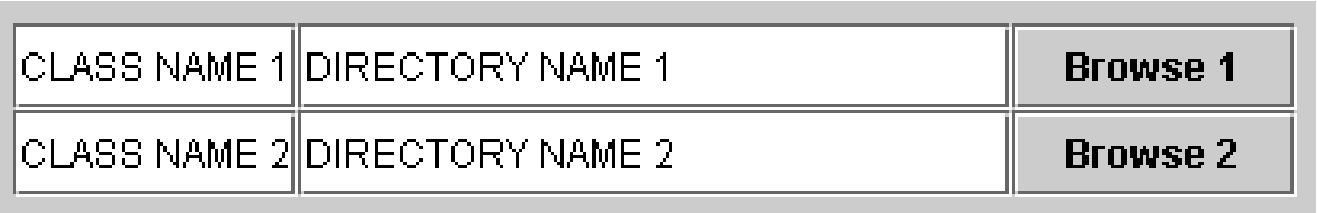

Figure G.26: **Class image directory location tab of Image Data File Creator application:** This tab allows you to specify the name of each class, and the location of the directory of images for that class.

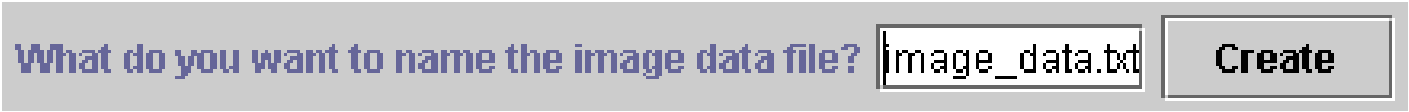

Figure G.27: **Image data filename tab of Image Data File Creator application:** This is the final tabbed pane in the application. This tab is used to specify the name of the data file.

## G.6 Image Leave One Out test application

This application uses an identical interface to that of the normal TCMNN testing system mentioned earlier. The main difference is the output of the results file created by this application. Example printouts of the results and statistics files generated by this application can be seen in appendix F.

To run the Image Leave One Out test application type the following at a dos prompt (or equivalent environment):

```
java TCMNN_Image_LeaveOneOut_Application
```

# TEST EXAMPLE #21

### CORRECTLY CLASSIFIED

**CLASSIFICATION TABLE**

| | |
|---|---|
| **Image:** | 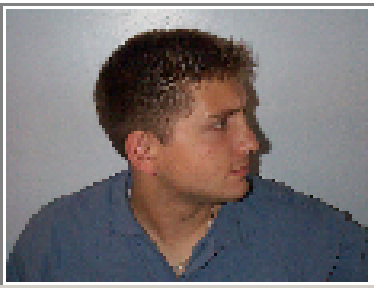 |
| **Actual Class:** | Dave (0) |
| **Predicted Class:** | Dave (0) |
| **Confidence:** | 98.181816% |
| **Credibility:** | 57.272728% |

**P VALUE TABLE**

| Class | P-Value |
|---|---|
| **Dave (0):** | 57.272728% |
| **Sian (1):** | 1.8181819% |

Figure G.28: **Example from results file create by Image leave one out application:** The results of the TCMNN test are structured in an HTML file. A thumbnail of the original image is included in the results for each example.